\documentclass[sigconf,screen]{acmart}%review,anonymous%,nonacm prologue,dvipsnames,table,xcdraw
\AtBeginDocument{%
  \providecommand\BibTeX{{%
    \normalfont B\kern-0.5em{\scshape i\kern-0.25em b}\kern-0.8em\TeX}}}
    
\usepackage[ruled,vlined]{algorithm2e}
\SetCommentSty{mycommfont}

\usepackage{makecell}
\usepackage{subcaption}
\usepackage{multirow}
\usepackage{color, soul}
\usepackage{multibib}
\usepackage{xr}
\makeatletter
\newcommand*{\addFileDependency}[1]{% argument=file name and extension
  \typeout{(#1)}
  \@addtofilelist{#1}
  \IfFileExists{#1}{}{\typeout{No file #1.}}
}
\makeatother

\usepackage{xcolor,colortbl}
\definecolor{green}{rgb}{0.1,0.1,0.1}
\definecolor{mycolor}{rgb}{0.9,0.7,0.1}
\copyrightyear{2022}
\acmYear{2022}
\setcopyright{rightsretained}
\acmConference[GECCO '22]{Genetic and Evolutionary Computation Conference}{July 9--13, 2022}{Boston, MA, USA}
\acmBooktitle{Genetic and Evolutionary Computation Conference (GECCO '22), July 9--13, 2022, Boston, MA, USA}
\acmDOI{10.1145/3512290.3528749}
\acmISBN{978-1-4503-9237-2/22/07}

\begin{document}

% \title{ENCAS: Evolutionary Neural Cascade Search across Supernetworks}
\title{Evolutionary Neural Cascade Search across Supernetworks}

\author{Alexander Chebykin}
\affiliation{%
  \institution{Centrum Wiskunde \& Informatica}
  \city{Amsterdam}
  \country{The Netherlands}
}
\email{a.chebykin@cwi.nl}

\author{Tanja Alderliesten}
\affiliation{%
  \institution{Leiden University Medical Center}
  \city{Leiden} 
  \country{the Netherlands} 
}
\email{t.alderliesten@lumc.nl}

\author{Peter A. N. Bosman}
\affiliation{%
  \institution{Centrum Wiskunde \& Informatica}
  \city{Amsterdam} 
  \country{the Netherlands} 
}
\affiliation{%
  \institution{TU Delft}
  \city{Delft} 
  \country{The Netherlands} 
}
\email{peter.bosman@cwi.nl}


\begin{abstract}

To achieve excellent performance with modern neural networks, having the right network architecture is important. Neural Architecture Search (NAS) concerns the automatic discovery of task-specific network architectures. Modern NAS approaches leverage supernetworks whose subnetworks encode candidate neural network architectures. These subnetworks can be trained simultaneously, removing the need to train each network from scratch, thereby increasing the efficiency of NAS.

A recent method called Neural Architecture Transfer (NAT) further improves the efficiency of NAS for computer vision tasks by using a multi-objective evolutionary algorithm to find high-quality subnetworks of a supernetwork pretrained on ImageNet. Building upon NAT, we introduce ENCAS~---~Evolutionary Neural Cascade Search. ENCAS can be used to search over multiple pretrained supernetworks to achieve a trade-off front of cascades of different neural network architectures, maximizing accuracy while minimizing FLOPs count.

We test ENCAS on common computer vision benchmarks (CIFAR-10, CIFAR-100, ImageNet) and achieve Pareto dominance over previous state-of-the-art NAS models up to 1.5 GFLOPs. Additionally, applying ENCAS to a pool of 518 publicly available ImageNet classifiers leads to Pareto dominance in all computation regimes and to increasing the maximum accuracy from 88.6\% to 89.0\%, accompanied by an 18\% decrease in computation effort from 362 to 296 GFLOPs. 
Our code is available at \url{https://github.com/AwesomeLemon/ENCAS}

\end{abstract}




\maketitle

\section{Introdution}

In recent years, deep neural networks have been successfully applied in domains ranging from text summarization~\cite{brown2020language} to medical image segmentation~\cite{isensee2018nnu}. Much of this success has been enabled by task-specific neural network architectures that are designed manually while making use of expert knowledge. The research direction of Neural Architecture Search (NAS)~\cite{zoph2016neural} has the goal of making architecture design automatic and data-driven. Tremendous progress has been made since the first approaches: performance of the found models improved~\cite{liu2018darts, sharma2020alphanet}, their size decreased~\cite{pham2018efficient, dong2019one} (smaller models usually work faster and require less storage space), and the search process itself became much more efficient (with required GPU-hours decreasing from tens of thousands~\cite{real2019regularized} to single-digit numbers~\cite{stamoulis2019single}).

\begin{figure}[h]
  \centering
  \includegraphics[width=0.8\linewidth]{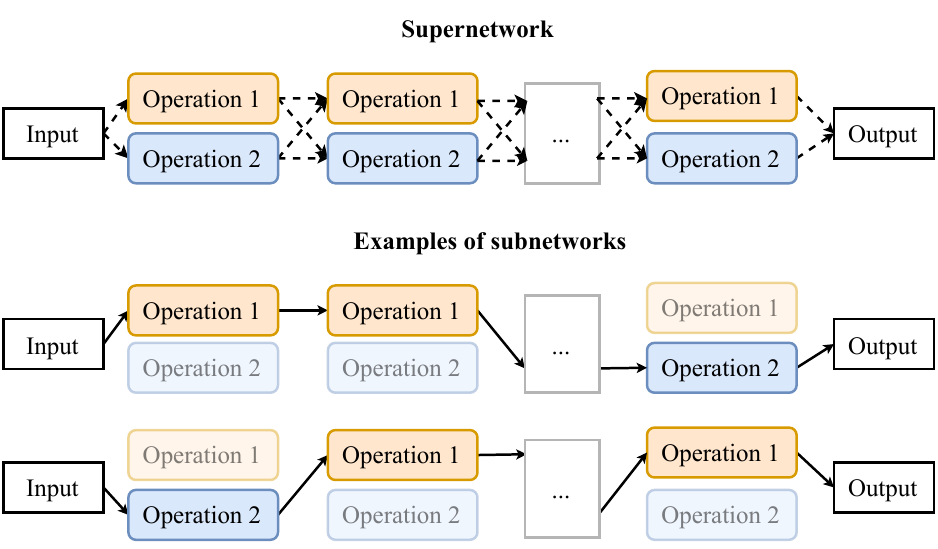}
  \vspace{-5pt}
  \caption{A schematic of a generic supernetwork. 
%   A supernetwork stores weights for all possible operations, each subnetwork reuses a part of these weights that corresponds to its operations.
}
\vspace{-2pt}
  \Description{}
  \label{fig:intro:supernet}
\end{figure}

These search efficiency gains can mostly be attributed to the idea of weight sharing via a supernetwork~\cite{pham2018efficient} (with performance prediction~\cite{baker2017accelerating, white2021powerful} also playing a role). Instead of training the weights of each candidate architecture from scratch, a supernetwork is constructed such that each architecture in the search space is a subset of the supernetwork (see Fig.~\ref{fig:intro:supernet}). To evaluate the quality of an architecture, the weights of the relevant part of the supernetwork are copied. With various architectures potentially sharing the same operations (e.g. convolution~\cite{lecun1995convolutional}, attention~\cite{bahdanau2014neural}), the amount of training needed decreases drastically.

% While initially weights of the final models needed to be reset and trained from scratch, modern methods[][] do not require this step, potentially allowing reusage of supernetworks pretrained on ImageNet for transfer learning to smaller datasets

% \begin{figure}[t]
%   \centering
%   \includegraphics[width=\linewidth]{_pics/1_intro/imagenet_before_2000.png}
%   \caption{ImageNet performance of various NAS methods and EfficientNet baselines. ENCAS achieves dominance for FLOPS below 2000. Note that no additional weight training has been performed, the gains arise due to the search for the cascades in pretrained supernetworks. \textit{DEBUG: will add more NAS approaches - as far as I know, they are all worse than the line we already have. Note that the fronts in the plot extend beyond 2000, but we don't dominate there, and I think this plot should only show the part where we do}}
%   \Description{Performances of different models on the ImageNet dataset, x-axis is FLOPS, y-axis is top-1 accuracy. The ENCAS trade-off front dominates others up to a point.}
%   \label{fig:intro:imagenetbefore2000}
% \end{figure}

However, by requiring that each architecture is a path within a supernetwork, the supernetwork approach inherently limits the diversity of architectures that can be produced. Thus, the manual choice of which supernetwork to use for the automated NAS procedure plays a large role, as it restricts the search space before the search algorithm even starts. With the growing number of available supernetworks, the issue of choosing the supernetwork is becoming increasingly important, and yet, to the best of our knowledge, there exists no method taking it into account.

There are many ways to improve neural network performance that are different from NAS. Ensembling~\cite{rokach2010ensemble} is one such technique that involves passing the same input through several different models and combining their predictions to get a better final prediction. It has been shown to work well if the models' mistakes are independent~\cite{esposito2003monte}, which is helped by the models being different from each other~\cite{wichard2003building}. NAS has been used~\cite{zaidi2020neural, chen2021one} to produce models that together make a good ensemble. Modern supernetwork-based approaches seem very fitting to this purpose because they do not incur additional training costs for ensembles of arbitrarily large size (once a supernetwork is trained, weights for trained subnetworks can be extracted from it and used without additional retraining\footnote{This holds for modern state-of-the-art approaches~\cite{cai2019once, sharma2020alphanet, wang2021attentivenas} but does not hold for all, especially older, approaches~\cite{liu2018darts, stamoulis2019single}.}).

\vspace{-2pt}

Cascading~\cite{viola2001rapid} is a particular case of ensembling with the focus on efficiency: whereas an ensemble requires that every input is processed by every model, a cascade proceeds in a sequential manner, invoking a larger model only if the predictions of smaller models are not confident enough (judged by the confidence function, see Section~\ref{relwork:cascade} for details). Thus, easy inputs consume less computational resources, while harder inputs can still be predicted well by utilizing every model in the cascade. Despite the potential of cascades to produce efficient and effective models, they remain underexplored in the context of deep learning~\cite{wang2020multiple}, and in particular no work has yet been done on combining NAS with cascade search.

To perform any kind of NAS, efficient search algorithms are indispensable. Evolutionary Algorithms (EAs) are particularly fit to the task, as they do not require the search space to be continuous, are known to solve real-world problems efficiently~\cite{chiong2012variants}, and excel in solving multi-objective and dynamic problems~\cite{van1998multiobjective, singh2006comparison, branke2003designing}.

Driven by the observations above, in this paper we present an algorithm called Evolutionary Neural CAscade Search (ENCAS). ENCAS is supernetwork-based and designed to take advantage of various pretrained supernetworks. ENCAS can search over a user-specified set of arbitrary supernetworks that may have e.g. different operations or numbers of layers (the only restriction being that subnetworks extracted from a supernetwork require no retraining). Our algorithm is multi-objective with the goal of finding models on the optimal effectivity-efficiency trade-off front.

The main contributions of our paper are threefold:

\begin{itemize}
    \item This work is the first to research the combination of NAS and cascades.
    % They are found to work well together, demonstrating feasibility of further research and confirming recent findings~\cite{wang2020multiple} that cascades are very beneficial for deep learning approaches.
    \item This work is the first to investigate the feasibility of using several different supernetworks in NAS. We explore whether the additional diversity they provide is helpful for creating cascades.
    % and find that \textit{PRELIM: there is a slight but statistically significant benefit, which should only increase as more diverse supernetworks become available.}
    \item The ENCAS algorithm is introduced to search for efficient and effective cascades. 
    % Variations of ENCAS can search for ensembles (ENCAS-ensemble) or adapt cascade weights and architectures simultaneously (ENCAS-joint).
    % The ENCAS algorithm is introduced for simultaneous search of single models, cascades, and ensembles over different supernetworks.
    % ; it is able to efficiently find effective models on benchmark datasets, outperforming previous NAS SOTA in the limited-FLOPS scenario. Application of ENCAS to manually designed SOTA ImageNet networks leads to a Pareto dominant front in all computation regimes. \textit{DEBUG: will mention the concrete metric numbers once I have them}
\end{itemize}

% The rest of the paper has the following structure: we review related work in Section~\ref{relwork} and describe our approach in Section~\ref{methods}. Our main experimental results are presented in Section~\ref{exp}, with additional experiments shown in Section~\ref{abl}. We discuss our approach and its limitations in Section~\ref{disc}, and conclude the paper in Section~\ref{conclusion}.

\section{Related work} \label{relwork}

\subsection{Neural Architecture Search}

The first methods to learn neural network architectures trained each candidate architecture from scratch, taking tens of thousands of GPU hours~\cite{zoph2016neural}. ENAS~\cite{pham2018efficient} introduced the ideas of weight sharing and supernetworks, which drastically decreased search costs. Multiple algorithms followed, most famously DARTS~\cite{liu2018darts} that reformulated the problem of operation selection as a continuous one, achieving great efficiency. 

However, supernetwork weights were discarded after NAS, with the final model being retrained from scratch (because it led to better results). This becomes costly when more than one model is required, e.g. considering both server-based and smartphone-based deployment. OnceForAll (OFA)~\cite{cai2019once} introduced an algorithm for training supernetwork weights such that they could be used in extracted subnetworks without any further training
% (though the ImageNet results of OFA are reported with additional training because it boosted performance by 1 percentage point)
. AttentiveNAS~\cite{wang2021attentivenas} and AlphaNet~\cite{sharma2020alphanet} introduced training techniques that lead to even better performance (albeit using a different search space).
% (albeit they were applied in a different search space of narrower and deeper networks, so no direct comparison was made); in addition, they achieve SOTA ImageNet results without any additional fine-tuning.

Neural Architecture Transfer~\cite{lu2021neural} (NAT) is an approach for fine-tuning a pretrained supernetwork for smaller datasets. The key difference from the previous approaches is that the architectures are adapted together with the weights in a multi-objective search procedure, trading off size and accuracy. The main idea is training only subnetworks close to the currently known trade-off front. To this end, a population of networks is evolved by Non-dominated Sorting Genetic Algorithm III (NSGA-III)~\cite{deb2013evolutionary}, a prominent many-objective EA. In this work, we aim to reproduce and to build upon NAT. It should be noted that NAT uses two sets of supernetwork weights that share the same search space\footnote{This was not clear to us from~\cite{lu2021neural}, but it was explained to us in private communication with the authors.}; in contrast, in this work we are primarily interested in supernetworks that represent different search spaces.

% In NAS, a variety of search algorithms has been used for the search itself: reinforcement learning~\cite{zoph2016neural}, continuous relaxation together with gradient descent~\cite{liu2018darts}, EAs~\cite{lu2021neural} (see Section~\ref{relwork:ea}).

\subsection{Search for architectures of neural ensembles}

Neural Ensemble Search~\cite{zaidi2020neural} (NES) is the first approach to search architectures of neural ensembles. It trained multiple networks separately, without weight sharing. The follow-up work, Multi-headed Neural Ensemble Search~\cite{narayanan2021multi} (MH-NES), utilizes weight sharing by having different ensemble members share first layers of the network. MH-NES achieves gains in robustness, search efficiency and model efficiency.
% but no accuracy gains, potentially because the ensemble members are too alike. 
% Another approach that went into the direction of having ensemble members share the network trunk is Neural Ensemble Architecture Search (NEAS)~\cite{chen2021one}. 
Neural Ensemble Architecture Search (NEAS)~\cite{chen2021one} is a similar approach with the key idea of gradual removal of the least promising operations from the supernetwork. 
% The results are reported only on ImageNet, whereas for NES and MH-NES no results were reported on that dataset, which makes direct comparison problematic.

Neural Ensemble Search via Sampling (NESS)~\cite{shu2021going} is supernet-based but does not require the ensemble models to have the same first layers. For this, a supernetwork is first trained, then subnetworks are sampled via novel sampling algorithms.
% sampled via either Monte Carlo sampling or particle-based Stein variational gradient descent with regularized diversity.

Our algorithm, ENCAS, substantially differs from all the ones discussed above. Firstly, ENCAS searches for architectures of cascades, for which no prior work exists (to the best of our knowledge). Secondly, ENCAS is truly multi-objective, with a single run producing multiple networks on a trade-off front of model size and performance, whereas NES, MH-NES, and NESS do not directly optimize model efficiency; NEAS uses model size as a constraint in single-objective optimization, which thus needs to be run once for each target model size. Thirdly, none of the existing algorithms take advantage of pretrained supernetworks, while we purposefully design our algorithm to rely on them, bearing in mind that pretraining plays a huge role in the success of deep learning approaches~\cite{donahue2014decaf}. Finally, all existing algorithms require the user to specify the ensemble size in advance. Our algorithm has the \emph{maximum} cascade size as a hyperparameter, meaning that it can output cascades with fewer networks if adding more networks does not improve the performance.

\vspace{-2pt}
\subsection{Cascades of neural networks} \label{relwork:cascade}

As mentioned, cascading is a way of efficiently combining multiple available models. It requires the user to choose a confidence function and confidence thresholds. The confidence function estimates how confident a model is in its prediction for a specific input. An example of that could be the maximum predicted probability or the gap between the largest and the second-largest logit values~\cite{streeter2018approximation}. The confidence thresholds are used to decide when to stop evaluation and to return the current output. A cascade operates sequentially in the following way: the current model makes a prediction and a confidence value for it is determined by the confidence function; if this value is above the confidence threshold for the current model, the cascade is terminated, otherwise the process is repeated for the next model. Note that the output of a cascade can be either the output of the last used model~\cite{streeter2018approximation} (i.e. the outputs of unconfident models are discarded), or the averaged outputs of all the used models~\cite{wang2020multiple}. We follow~\cite{wang2020multiple} in using the averaged outputs.

Cascades are popular in machine learning~\cite{kukenys2008classifier, xu2014classifier}, but in deep learning there are only a few works utilizing them~\cite{cai2015learning, angelova2015real}. Recently, \cite{wang2020multiple} pointed out that cascades can dominate single models in terms of performance, efficiency, and training time. Cascades in~\cite{wang2020multiple} were constructed via an exhaustive search of a small search space of predefined networks. 

GreedyCascade~\cite{streeter2018approximation} achieved good results by designing an efficient greedy algorithm for cascade selection from a somewhat larger number of networks. GreedyCascade has a fundamental limitation: by construction, it cannot produce a cascade that would perform better than the best model in the model pool. Our algorithm does not have this limitation in its design. In addition, GreedyCascade scales quadratically in the number of networks.

Our algorithm searches in a search space that is substantially larger than ever used before for cascade search (it contains hundreds of networks instead of dozens). Also, ENCAS is multi-objective and requires only a single run to create the trade-off front, unlike the exhaustive search procedure of~\cite{wang2020multiple}.

\vspace{-2pt}
\subsection{Evolutionary algorithms} \label{relwork:ea}

An EA is a population-based optimization algorithm that relies on the ideas of \textit{(1)} fitness-based selection and \textit{(2)} variation (most often mutation and crossover, i.e. information transfer between solutions in the population). EAs achieve SOTA results on a variety of benchmark and real-world problems~\cite{chiong2012variants, petchrompo2022pruning, bouter2019gpu}. 

In this paper, we use NSGA-III~\cite{deb2013evolutionary} (as part of NAT) and the Multi-objective Gene-pool Optimal Mixing Evolutionary Algorithm (MO-GOMEA)~\cite{luong2014multi}. NSGA-III relies on non-dominated sorting and pre-supplied reference points to keep the population spread-out in the objective space. The two key ideas behind MO-GOMEA are linkage learning (which leads to dependent variables being exchanged between solutions as a single group) and Gene-pool Optimal Mixing (which ensures that crossover always leads to a fitness improvement).
% Linkage learning means that during the search, genes are dynamically clustered into groups; during crossover, groups cannot be broken up, they have to be transferred or not transferred as a whole~---~this ensures that if an appropriate building block of a good solution is found, it will not be lost due to only one part of it being changed. The GOM operator is used for crossover: for a specific individual and a specific building block, GOM iterates over the population and tries to change the values of the building block of the individual to the values from the current population member, accepting the change only if it improves the objective function. 
Since we use the algorithm without any modifications, we refer the interested reader to~\cite{luong2014multi, thierens2011optimal} for details. We choose to use this algorithm because it performs well in many problems~\cite{rodrigues2016multi, luong2015scalable} and because it does not require setting any hyperparameters (such as population size, crossover type, or mutation rate).

\section{Methods} \label{methods}

\subsection{Searching for cascades of dynamic size} \label{methods:ens}

Let us assume that a supernetwork has been trained via the NAT procedure. In addition to the supernetwork weights, the procedure generates a trade-off front of network architectures. The architectures from this front will be used in ENCAS.

%ENCAS is conceptually simple.
Creating a cascade of a specific size out of a predefined model pool comes down to selecting the appropriate sequence of models, and their confidence thresholds. Let us focus on the models first. Each out of $N$ models can be encoded as a categorical value $1..N$. To consider cascades with fewer models (or even a single model), we add the value $0$ to encode a "no operation" model~---~a model that does nothing. Then, a solution to the problem of searching for a cascade of maximum size $k$ can be encoded as a list of $k$ values, each ranging from $0$ to $N$. In the interest of robustness we do not search for the weights of the models in the cascade, and take a simple average of their outputs instead. 

As to the confidence threshold values, they are encoded as ${k-1}$ additional categorical variables. In our experiments we use 51 possible values from $0.0$ to $1.0$ with step size $0.02$. If all thresholds are equal to $1$, the cascade becomes an ensemble, since the confidence of any model will always be smaller than $1$, and thus all the models will be used.

\begin{figure}[h]
  \centering
  \vspace{-6pt}
  \includegraphics[width=0.7\linewidth]{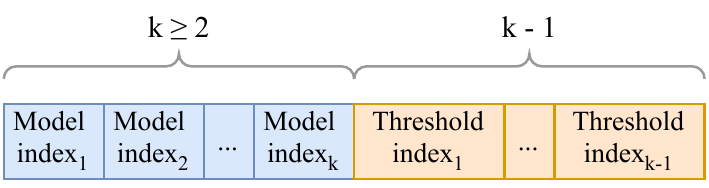}
  \vspace{-4pt}
%   \caption{Representation of a solution for ENCAS (full representation) and ENCAS-ensemble (the highlighted part). In our experiments $k=5$.}
  \caption{Representation of a solution for ENCAS.}
  \vspace{-5pt}
  \Description{}
  \label{fig:methods:genome}
\end{figure}

A visual representation of a solution can be seen in Figure~\ref{fig:methods:genome}.

To evaluate a solution, every subsequent model is used to only update probabilities of inputs for which previous models were not confident (once the confidence for an input is above the current threshold, cascading stops). The final probabilities are used as cascade predictions to evaluate its performance. The FLOPs of a cascade are computed as a weighted sum of the FLOPs of all the models in the cascade, where each weight is the fraction of the total number of inputs that this model was used on. Since the models in the model pool are known in advance, their outputs on the validation set can be precomputed, which leads to fast search times of under 1 GPU-hour even on a large dataset (e.g. ImageNet) and with hundreds of base models to choose from.

Note that this approach is trivial to extend to multiple supernetworks by adding the models from the trade-off fronts of all the supernetworks to the model pool. Since the algorithm relies on network outputs, the problem of different supernetworks having different operations is side-stepped.

Any multi-objective search algorithm can be run to maximize validation accuracy and minimize FLOPs. We use MO-GOMEA~\cite{luong2014multi}. Pseudocode of ENCAS is listed in Algorithm~\ref{algo:encas}.

\vspace{-1pt}
\begin{algorithm}[h]
\DontPrintSemicolon
\SetKwInOut{KwIn}{Input}
\SetKwInOut{KwOut}{Output}
%   \KwIn{Supernetworks \{S_i\}, possible first thresholds \{t_i\}, possible threshold differences \{dt_i\}, maximum cascade size k}
  \KwIn{Supernetworks $\{S_i\}$, possible thresholds $\{t_i\}$, maximum cascade size $k$}
  model\_pool = []
  
  \For{$S_i$ in $\{S_i\}$}{
    trade\_off\_front$_i$ = NAT($S_i$)
    
    model\_pool = model\_pool $\cup$ trade\_off\_front$_i$
  }
  
  fitness\_func = make\_fitness\_func(model\_pool, $\{t_i\}$, $k$)
  
  \tcc{make\_fitness\_func defines the procedure for decoding and encoding the values (see Fig.~\ref{fig:methods:genome}), and for evaluating a solution, i.e. a cascade (see Section~\ref{relwork:cascade}).}
  
  cascades = MO-GOMEA(fitness\_func) 
  
  \textbf{return} cascades \tcp{the trade-off front}
  
%   \;
  
%   \SetKwFunction{FMain}{make\_cascade\_func}
%   \SetKwProg{Fn}{Function}{:}{}
%   \Fn{\FMain{model\_pool, $\{t_i\}$, $\{dt_i\}$, $k$}}{
%         a\;
%         b\;
%         \KwRet\;
%   }
\caption{ENCAS}
\label{algo:encas}
\end{algorithm}

% ACHTUNG! The vspace below is useful in non-anon version, but breaks things in anon version
% \vspace{-4pt}

Empirically, we observed that ENCAS finds hundreds of cascades. To reduce that number and to combat overfitting, the trade-off front found by ENCAS is filtered: we traverse the non-dominated front from least accurate to most accurate cascades, and a cascade is kept if its accuracy on the validation set after rounding to the first significant digit is higher than the accuracy of the previous cascade.

% \vspace{-2pt}
\subsection{Joint training and cascade search} \label{methods:joint}

In ENCAS, only the models from the trade-off front of each supernetwork are considered. This means that ENCAS works with models that are very good on their own, but it also means that it cannot create cascades of models that might be subpar individually but extremely good together. Models with weights that complement each other in this way may not even exist in separately-trained supernetworks, so ideally the training of a supernetwork and the cascade search should happen simultaneously. To investigate whether this idea is feasible, we construct a version of ENCAS called ENCAS-joint (indicating that training and search are performed jointly).

\begin{figure}[h]
  \centering
  \vspace{-3pt}
  \includegraphics[width=\linewidth]{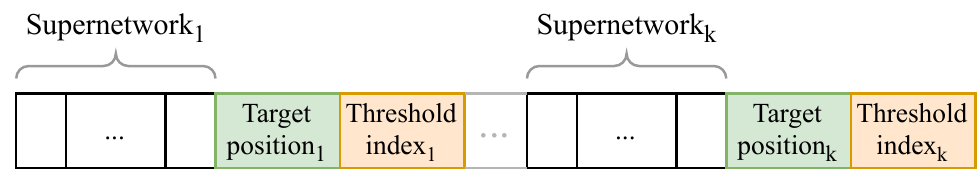}
  \vspace{-6pt}
  \caption{Representation of a solution in ENCAS-joint.}
  \vspace{-5pt}
  \Description{}
  \label{fig:methods:joint}
\end{figure}

ENCAS-joint extends NAT to training several different supernetworks simultaneously. Whereas in NAT a solution represents an architecture of a single model, in ENCAS-joint a solution represents architectures of all the models in a cascade, their target positions, and the threshold values. Confidence thresholds are restricted to 10 possible values from $0.1$ to $1.0$ with step size $0.1$ to decrease the search space size; the confidence threshold of the last network is not used. These per-supernetwork representations are concatenated to encode the whole cascade. Figure~\ref{fig:methods:joint} visualizes the encoding.

Note that our encoding of the order of the networks is generally inefficient (i.e. a permutation would be more efficient), but since we use a small number of supernetworks (5), the number of possible orderings is quite small, and our Cartesian encoding should suffice.

Before evaluating a cascade, the networks are ordered (with ties broken arbitrarily), after which the cascade is evaluated as usual. NAT requires defining a surrogate that predicts the fitness of an architecture. To extend this to multiple supernetworks in a simple way, we create such a surrogate for each supernetwork used, and an additional surrogate that combines outputs of supernetwork-wise surrogates for a prediction for the whole cascade. Yet another surrogate is used to estimate the FLOPs count for a cascade from FLOPs of individuals models, target positions, and thresholds (this is necessary because changing thresholds or the order of networks impacts not only its performance but also the FLOPs count). Each of the surrogates is a Radial Basis Function (RBF)~\cite{broomhead1988multivariable} ensemble, the same as in NAT.

Each supernetwork is trained separately based on which subnetworks from it are present in the population. In order not to disadvantage supernetworks that might require more training to achieve good accuracy, we train all the supernetworks for an equal number of steps. To avoid tuning hyperparameters of NSGA-III to the new scenario, we exchange it for MO-GOMEA, for which no hyperparameters are tuned. 

% If the $i$-th supernetwork search space has size $size_i$, for $k$ supernetworks the combined search space size becomes $\prod_{i=1}^k size_i$. For sufficiently similar supernetworks this can potentially be avoided by modifying the encoding to be more efficient, but this cannot be done for dissimilar supernetworks and search spaces, which is a scenario that is more interesting.

Note that this approach has a limitation of not allowing a cascade to contain several models from the same supernetwork. Therefore, it is possible to further improve results by running ENCAS on the supernetworks trained by ENCAS-joint (we refer to this combination as ENCAS-joint+). In the next section we compare all versions of our algorithm.

% \vspace{-2pt}
\section{Experiments} \label{exp}

We conduct experiments on established computer vision benchmark datasets: ImageNet (ILSVRC2012)~\cite{russakovsky2015imagenet}, CIFAR-10~\cite{krizhevsky2009learning}, and CIFAR-100~\cite{krizhevsky2009learning}. In our experiments, we consider the bi-objective problem of maximizing top-1 accuracy while minimizing the FLOPs count.
% \footnote{Counting the floating points operations allows for a better understanding of the model "size" than counting the parameters~\cite{}}
Note that hyperparameter selection and all search procedures were performed on the validation subsets, while the experimental results are reported on the test sets. This means that a trade-off front that is monotonous when evaluated on the validation set often becomes non-monotonous when evaluated on the test set. Since one should not use the test set to select models, we show all the models, even if they are dominated once the test accuracy is considered.

The validation sets for CIFAR-10 and CIFAR-100 consist of 10,000 images randomly split off from the training set (that contain 50,000 images; test sets contain 10,000 images). For ImageNet we rely on the pretrained networks that use the whole training set and report the results on the ILSVRC2012 validation set, as is established practice, since the true test set is not publicly available. Images seen during training cannot be used during the search because their activations have different statistics~\cite{streeter2018approximation}, and for comparability our results should be reported on the ILSVRC2012 validation set (which is treated as the test set). As the actual validation set, we therefore used 20,683 images from ImageNetV2~\cite{recht2019imagenet}, which is a dataset designed to match the ImageNet collection procedure as closely as possible\footnote{ImageNetV2 contains three sets of 10,000 images that were collected slightly differently, we use images from all three sets: removing duplicates gives 20,683 images.}. We would prefer to avoid using this additional data, but cannot; as the number of these images is only 1.6\% of the ImageNet training set size, we assume that the unfair advantage we gain by using it is negligible.
% and is offset by the distributional shift between the datasets~\cite{recht2019imagenet}.

We use the normalized hypervolume indicator~\cite{zitzler2003performance} as a metric of the quality of a trade-off front (see Appendix~\ref{appendix:hypervolume} for details). Every experiment is run 10 times, mean and standard deviation are reported. We plot the median run (in terms of hypervolume), along with a shaded area delimited by the worst and best fronts achieved over all the runs. Appendix~\ref{appendix:hyperparameters} contains our hyperparameters. Search time is measured on a single Nvidia 2080TI GPU. For statistical testing we use the Wilcoxon signed-rank test~\cite{wilcoxon1992individual} with Bonferroni correction~\cite{dunn1961multiple} (target $p$-value=0.01, 20 tests, corrected $p$=0.0005, mentions of statistical significance in the text imply smaller $p$, all $p$-values are reported in Appendix~\ref{appendix:stats}).

Our code is public\footnote{\url{https://github.com/AwesomeLemon/ENCAS}}. We have worked with our own implementation of NAT because we did not have access to the authors' code of NAT that was used for the NAT article.

% \vspace{-2pt}
\subsection{Baseline supernetworks} \label{exp:base}

We are interested in utilizing pretrained supernetworks, as training one from scratch takes on the order of thousands of GPU-hours~\cite{cai2019once}. Unfortunately, many papers do not release either code or pretrained weights. As more supernetworks become available in the future, the value of searching across supernetworks should only increase.

In our experiments we rely on five different supernetworks pretrained on ImageNet: AttentiveNAS~\cite{wang2021attentivenas}, AlphaNet~\cite{sharma2020alphanet}, ProxylessNAS~\cite{cai2018proxylessnas}, OFA-w1.0~\cite{cai2019once}, OFA-w1.2~\cite{cai2019once}. All of them are built from inverted residual blocks~\cite{sandler2018mobilenetv2}. Moreover, ProxylessNAS, OFA-w1.0, OFA-w1.2 have the same search space, with only width multipliers being different (to get the actual number of neurons in a layer, the base number of neurons is multiplied by the width multiplier). AttentiveNAS and AlphaNet are from the same search space, but the weights were trained via different approaches.

To adapt a supernetwork to CIFAR-10 and CIFAR-100, we apply the NAT procedure for each supernetwork separately. This produces trade-off fronts of models, which in the following sections will be used for cascade construction. For CIFAR-10 and CIFAR-100 we also reproduce NAT with its original hyperparameters using OFA-w1.0 and OFA-w1.2 (due to computational constraints, we do not reproduce NAT for ImageNet). For ImageNet, there is no need to further update the weights, however the trade-off front still needs to be found. For this reason, we run a version of NAT with no training and no reevaluation of the already evaluated networks.

% \begin{figure*}[h]
%     \centering
%     \begin{subfigure}[t]{0.25\textwidth}
%         \centering
%         \includegraphics[width=\textwidth]{_pics/4_experiments/baselines_cifar10.pdf}
%         \vspace{-14pt}
%         \caption{CIFAR-10}
%     \end{subfigure}%
%     ~ 
%     \begin{subfigure}[t]{0.239\textwidth}
%         \centering
%         \includegraphics[width=\textwidth]{_pics/4_experiments/baselines_cifar100.pdf}
%         \vspace{-14pt}
%         \caption{CIFAR-100}
%     \end{subfigure}
%     ~ 
%     \begin{subfigure}[t]{0.237\textwidth}
%         \centering
%         \includegraphics[width=\textwidth]{_pics/4_experiments/baselines_imagenet.pdf}
%         \vspace{-14pt}
%         \caption{ImageNet}
%     \end{subfigure}
%     % \subfloat[\centering label 1]{{\includegraphics[width=5cm]{_pics/4_experiments/baselines_c10.png} }}%
%     % \qquad
%     % \subfloat[\centering label 2]{{\includegraphics[width=5cm]{_pics/4_experiments/baselines_c100.png} }}%
% %   \includegraphics[width=\textwidth]{_pics/4_experiments/baselines_single}
%   \vspace{-7pt}
%   \caption{Results of running NAT~\cite{lu2021neural} with different supernetworks on CIFAR-10, CIFAR-100, ImageNet.}
%   \vspace{-7pt}
%   \Description{}
%   \label{fig:exp:baselines}
% \end{figure*}

\begin{figure}[h]
    \begin{subfigure}[t]{0.43\linewidth}
        \centering
        \includegraphics[width=\linewidth]{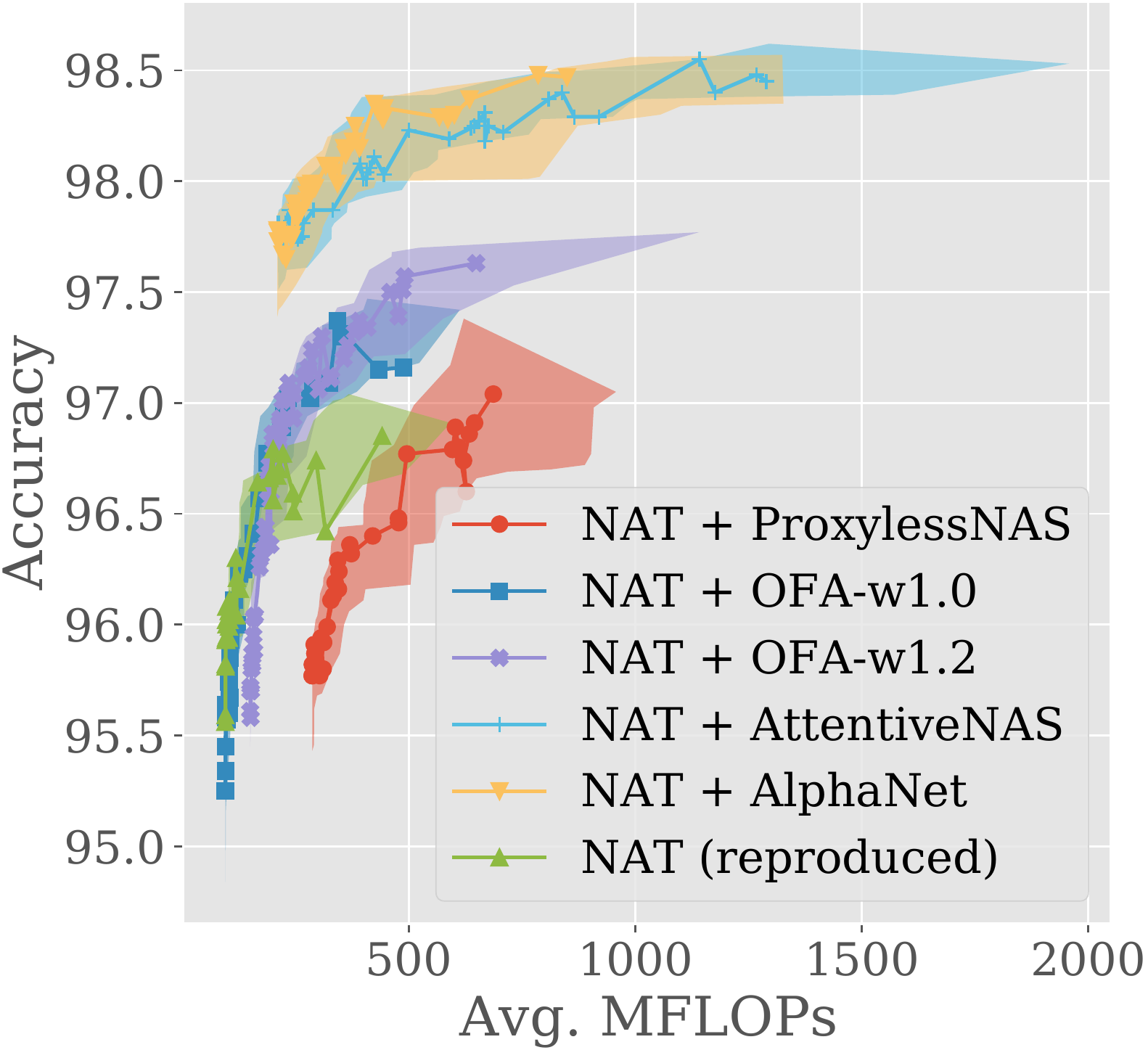}
        \vspace{-14pt}
        \caption{CIFAR-10}
        \vspace{4pt}
    \end{subfigure}%
    ~ 
    \begin{subfigure}[t]{0.41\linewidth}
        \centering
        \includegraphics[width=\linewidth]{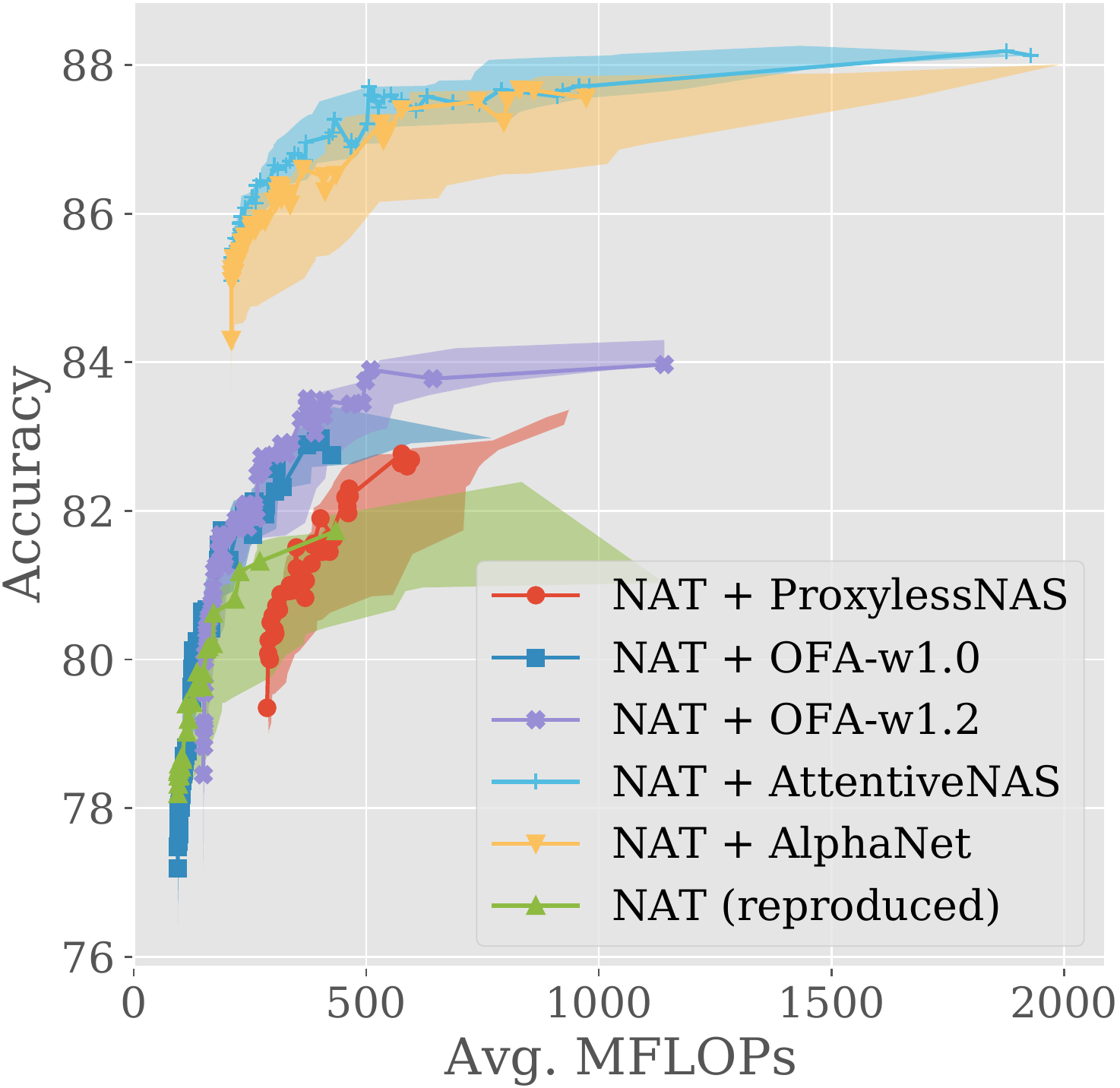}
        \vspace{-14pt}
        \caption{CIFAR-100}
        \vspace{4pt}
    \end{subfigure}
    
    \begin{subfigure}[t]{0.43\linewidth}
        \centering
        \includegraphics[width=\linewidth]{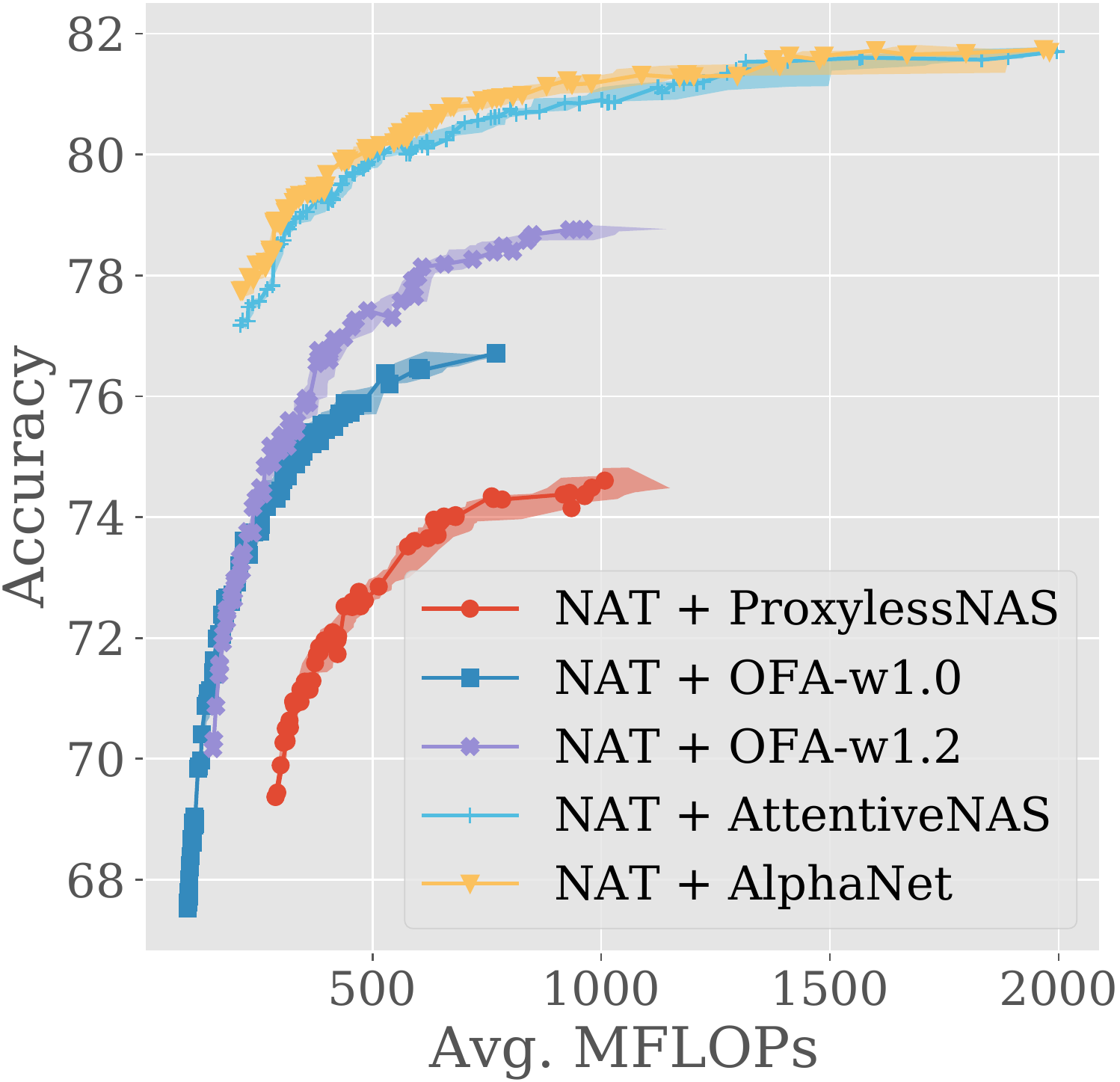}
        \vspace{-14pt}
        \caption{ImageNet}
    \end{subfigure}
  \vspace{-5pt}
  \caption{Results of running NAT~\cite{lu2021neural} with different supernetworks on CIFAR-10, CIFAR-100, ImageNet.}
  % ACHTUNG! The vspace below is useful in non-anon version, but breaks things in anon version
%   \vspace{-10pt}
  \Description{}
  \label{fig:exp:baselines}
\end{figure}

Results of using NAT with each supernetwork are presented in Fig.~\ref{fig:exp:baselines}. It can be seen that the choice of the supernetwork impacts the resulting trade-off front significantly, with supernetworks that perform better on ImageNet also performing better on CIFAR-10 and CIFAR-100, as expected~\cite{kornblith2019better}. Additionally, our reproduced NAT achieves results inferior to those reported in~\cite{lu2021neural}, even after we introduced changes that were not in the article but suggested by the authors in private communication (see Appendix~\ref{appendix:hyperparameters}). This prompted us to look for better hyperparameters, which are used in all our experiments (including those in Fig.~\ref{fig:exp:baselines}). With these hyperparameters, search time is 30 GPU-hours for OFA-w1.0, OFA-w1.2, or ProxylessNAS, and 45 GPU-hours for AlphaNet or AttentiveNAS.

% Our results are presented in Fig.~\ref{fig:exp:baselines}. It can be seen that adapting the supernetworks using NAT helps achieve competitive performance in comparison to other NAS algorithms. Nonetheless, we would like to note that the original NAT publication achieved results better than this while using only OFA-w1.0 and OFA-w1.2 supernetworks (which are outperformed by other supernetworks in our experiments). After consultation with the authors, we modified the algorithm accordingly (introducing changes not mentioned in the NAT paper), yet still were not able to reproduce the results completely. Since the performance of a cascade relies on the performance of the base models, we would be only glad if the authors released their full code or pointed out the mistakes in ours~---~it would both benefit our results and contribute to the culture of reproducibility. 

% ACHTUNG! The vspace below is useful in non-anon version, but breaks things in anon version
\vspace{-3pt}
\subsection{Cascading best NAT results} \label{exp:cas_single}

\begin{figure}[h]
  \begin{subfigure}[t]{0.43\linewidth}
        \centering
        \includegraphics[width=\linewidth]{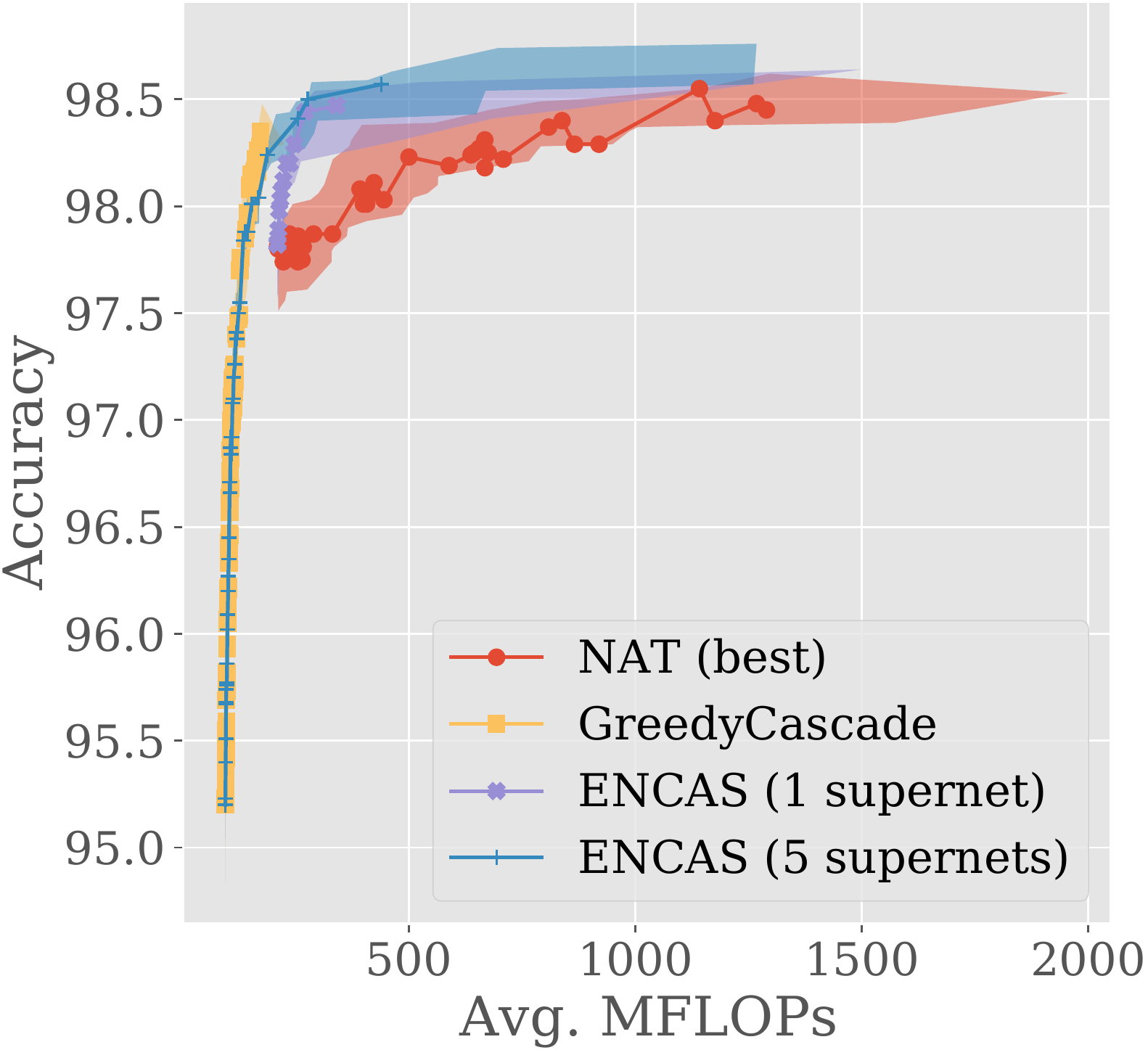}
        \vspace{-14pt}
        \caption{CIFAR-10}
        \vspace{4pt}
    \end{subfigure}%
    ~ 
    \begin{subfigure}[t]{0.42\linewidth}
        \centering
        \includegraphics[width=\linewidth]{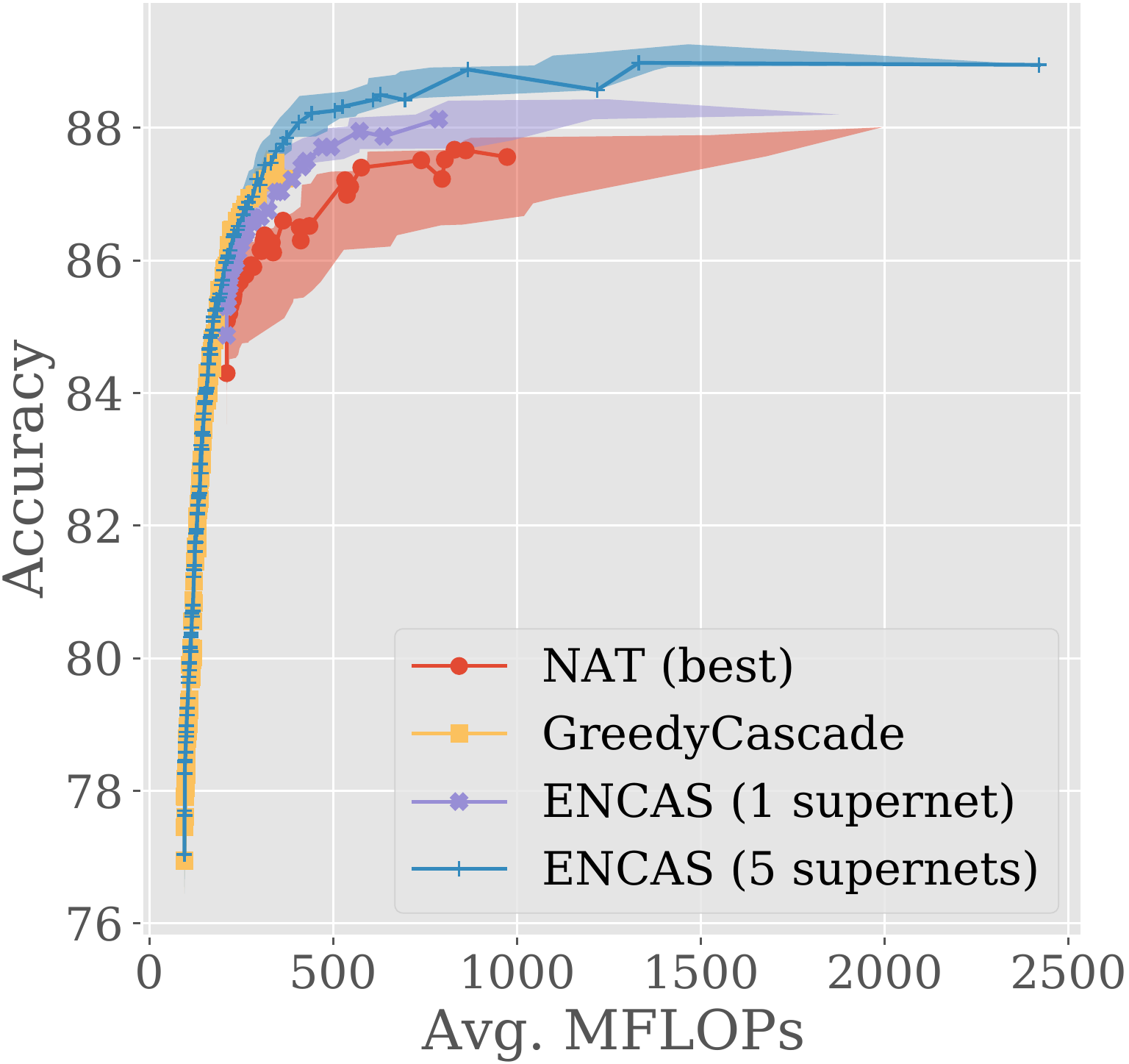}
        \vspace{-14pt}
        \caption{CIFAR-100}
        \vspace{4pt}
    \end{subfigure}
    
    \begin{subfigure}[t]{0.43\linewidth}
        \centering
        \includegraphics[width=\linewidth]{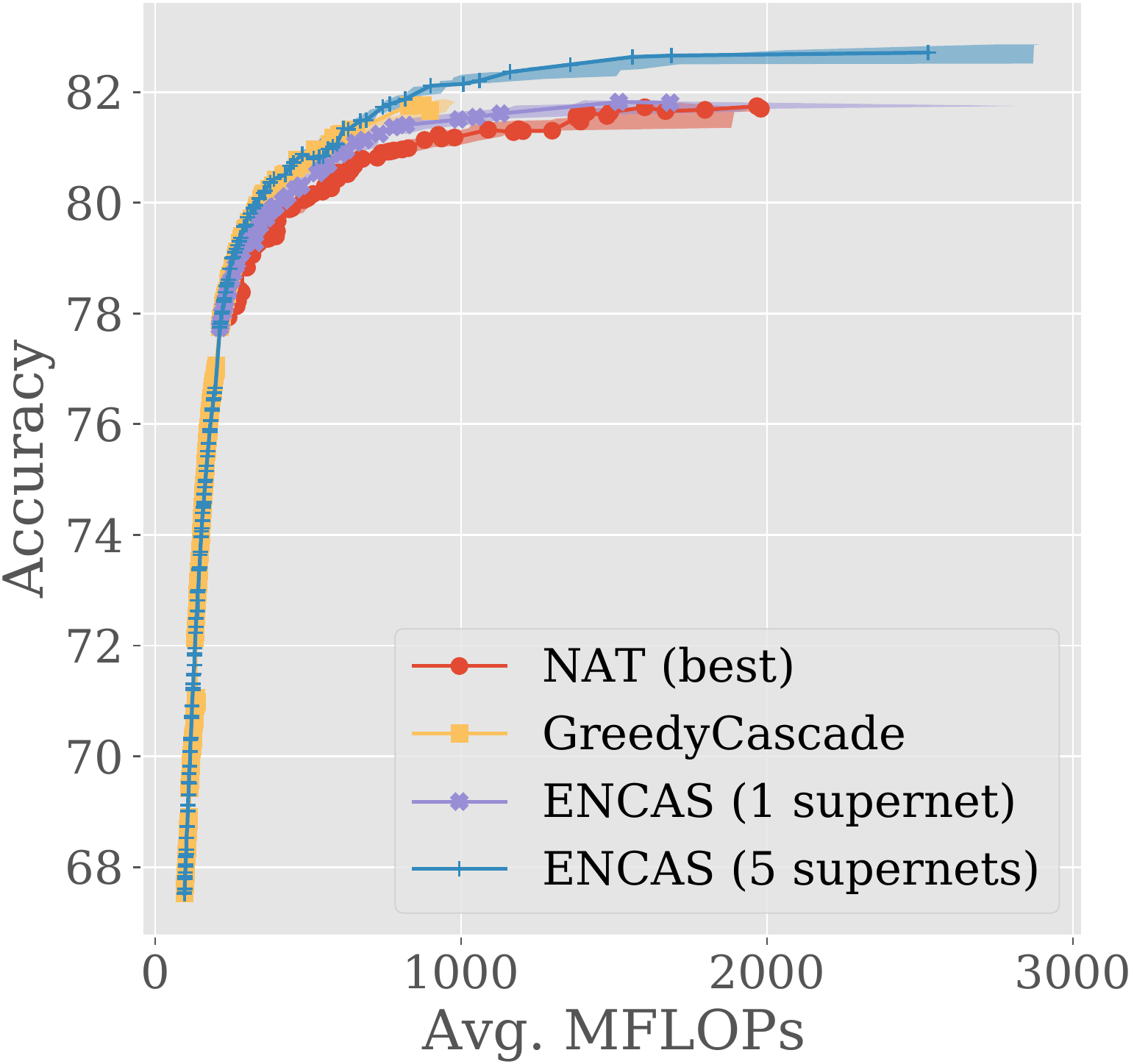}
        \vspace{-14pt}
        \caption{ImageNet}
    \end{subfigure}
\vspace{-5pt}
  \caption{Comparing ENCAS to the baselines.}
%   \vspace{-10pt}
  \Description{}
  \label{fig:exp:cascades}
\end{figure}

% Having established the baselines, we investigate the impact of cascade search. The first question is: can ENCAS improve the trade-off front of the best supernetwork by only using networks from this front? 
Figure~\ref{fig:exp:cascades} and Tables~\ref{tab:imagenet},~\ref{tab:cifar100},~\ref{tab:cifar10} show that using ENCAS on the NAT results with the single best supernetwork leads to the models on the fronts becoming more efficient across all datasets, with hypervolumes increasing (statistically significant; difference in maximum accuracy is not statistically significant). Visually, the effect is small on ImageNet, and noticeable on CIFAR-10 and CIFAR-100.

% \begin{figure*}[h]
%   \centering
%   \begin{subfigure}[t]{0.25\textwidth}
%         \centering
%         \includegraphics[width=\textwidth]{_pics/4_experiments/cascades_and_greedy_cifar10.pdf}
%         \caption{CIFAR-10}
%     \end{subfigure}%
%     ~ 
%     \begin{subfigure}[t]{0.25\textwidth}
%         \centering
%         \includegraphics[width=\textwidth]{_pics/4_experiments/cascades_and_greedy_cifar100.pdf}
%         \caption{CIFAR-100}
%     \end{subfigure}
%     ~ 
%     \begin{subfigure}[t]{0.25\textwidth}
%         \centering
%         \includegraphics[width=\textwidth]{_pics/4_experiments/cascades_and_greedy_imagenet.pdf}
%         \caption{ImageNet}
%     \end{subfigure}
% %   \includegraphics[width=\textwidth]{_pics/4_experiments/cascades_and_greedy}
%   \caption{Comparison of ENCAS performance when just the best supernetwork is used, or all of them. Baselines are the front from the best supernetwork and GreedyCascade. }
%   \Description{}
%   \label{fig:exp:cascades}
% \end{figure*}

% \vspace{-4pt}

\subsection{Cascading all NAT results} \label{exp:cas_many}

The next question is whether using supernetworks other than the best one will improve the results of ENCAS. As shown in Figure~\ref{fig:exp:cascades}, the results are strongly improved, with the differences in hypervolume and maximum accuracy to the best NAT baseline (and to ENCAS with 1 supernetwork) being statistically significant. We hypothesize that inclusion of better and more diverse supernetworks would make the gap even larger. Search time of ENCAS is 1 GPU-hour (with approximately 300 base models). Since we report all the cascades found by ENCAS (several dozen), we do not name them, but for the ease of reference we name a subset (see Appendix~\ref{appendix:named_models}).

\subsection{Comparison to SOTA} \label{exp:greedy}

We compare ENCAS with the SOTA cascade search algorithm GreedyCascade~\cite{streeter2018approximation} by applying it to the same model pool. As can be seen in Figure~\ref{fig:exp:cascades}, ENCAS matches the performance of GreedyCascade for smaller FLOPs on all datasets, and can find cascades with better accuracy than the best baseline model, which GreedyCascade cannot do. Note that the runtime of both algorithms (under 1 hour) is negligible in comparison to the supernetwork adaptation time (tens of hours). Differences in hypervolume and maximum accuracy between ENCAS and GreedyCascade are statistically significant.

\begin{table}
  \caption{ImageNet performance, "acc." is top-1 accuracy. The method producing the highest accuracy is in bold.
  }
  \vspace{-4pt}
  \label{tab:imagenet}
  \begin{tabular}{lccc}
    \toprule
    \multicolumn{1}{c}{Method} & \makecell{Hyper-\\volume} & \makecell{Max\\acc.} & \makecell{Max\\MFLOPs}\\
    \midrule
    EfficientNet B0-B3~\cite{tan2019efficientnet} &  0.464 & 81.6 & 1800\\
    EfficientNet C0-C3~\cite{wang2020multiple} &  0.482 & 82.2 & 1800\\
    MobileNetV3~\cite{howard2019searching} & 0.396 & 76.6 & 350\\
    OFA (\#75)~\cite{cai2019once} & 0.451 & 80.0 & 595\\
    \makecell[l]{NEAS~\cite{chen2021one}} & 0.458 & 80.0 & 574\\
    \makecell[l]{NSGANetV2~\cite{lu2020nsganetv2}} & 0.477 & 80.4 & 593\\
    \makecell[l]{AlphaNet~\cite{sharma2020alphanet}} & 0.490 & 80.8 & 709\\
    \makecell[l]{BigNAS~\cite{yu2020bignas}} & 0.480 & 80.9 & 1040\\
    NAT (best)~\cite{lu2021neural} & 0.506$_{\pm0.001}$ & 81.69$_{\pm0.02}$ & 1962$_{\pm33}$\\
    \makecell[l]{GreedyCascade~\cite{streeter2018approximation}} & 0.520$_{\pm0.002}$  & 81.72$_{\pm0.11}$ & 927$_{\pm36}$\\
    \midrule
    ENCAS (1 supernet) & 0.509$_{\pm0.001}$ & 81.78$_{\pm0.05}$ & 1929$_{\pm424}$\\
    \textbf{ENCAS (5 supernets)} & \textbf{0.537$_{\pm0.001}$} &\textbf{ 82.72$_{\pm0.10}$ }& 2616$_{\pm221}$\\
  \bottomrule
\end{tabular}
\end{table}

\begin{table}
  \caption{CIFAR-100 performance, "acc." is top-1 accuracy. The method producing the highest accuracy is in bold.
  }
  \vspace{-4pt}
  \label{tab:cifar100}
  \begin{tabular}{lcccc}
    \toprule
    \multicolumn{1}{c}{Method}& \makecell{Hyper-\\volume} & \makecell{Max\\acc.} & \makecell{Max\\MFLOPs}\\
    \midrule
    \makecell[l]{EfficientNet B0-B3~\cite{tan2019efficientnet}} & 0.664 & 89.9 & 1800\\
    \makecell[l]{NSGANetV2~\cite{lu2020nsganetv2}} & 0.658 & 88.3 & 796\\
    \makecell[l]{GDAS~\cite{dong2019searching}}  & --- & 81.87 & 519\\
    \makecell[l]{SETN~\cite{dong2019one}}  & --- & 82.75 & 722\\
    \makecell[l]{NAT (reproduced)~\cite{lu2021neural}} & 0.523$_{\pm0.012}$ & 81.53$_{\pm0.54}$ & 682$_{\pm298}$\\
    \makecell[l]{NAT (best)~\cite{lu2021neural}} & 0.659$_{\pm0.004}$  & 87.94$_{\pm0.23}$ & 1277$_{\pm287}$\\
    \makecell[l]{GreedyCascade~\cite{streeter2018approximation}} & 0.665$_{\pm0.005}$  & 87.25$_{\pm0.25}$ & 341$_{\pm30}$\\
    \midrule
    ENCAS (1 supernet) & 0.663$_{\pm0.005}$ & 88.09$_{\pm0.21}$ & 1051$_{\pm353}$\\
    ENCAS (5 supernets) & 0.699$_{\pm0.003}$ & 88.96$_{\pm0.17}$ & 1401$_{\pm420}$\\
    \midrule
    ENCAS-joint & 0.681$_{\pm0.005}$ & 88.55$_{\pm0.19}$ & 6678$_{\pm1735}$\\
    \textbf{ENCAS-joint+} & \textbf{0.701$_{\pm0.005}$} & \textbf{89.10$_{\pm0.22}$} & 1750$_{\pm418}$\\
  \bottomrule
\end{tabular}
\end{table}

% In addition, ENCAS is much faster. For one supernetwork (approx. 50 models) ENCAS can achieve good results in ~1 GPU-hour~\footnote{We resort to comparison in GPU-hours because GreedyCascade spends most of its compute not on evaluating the cascades, making comparison in terms of evaluations meaningless}, while GreedyCascade takes \textit{PRELIM: ~4 GPU-hours}. ENCAS also scales better with the increasing amount of networks: for all five supernetworks (approx. 250 models) it can achieve good results in the same ~1 GPU-hour, while GreedyCascade takes \textit{PRELIM: ~30 GPU-hours}.

% \textit{DEBUG: Also of note is that GreedyCascade is able to achieve better results for smaller FLOPS most likely due to the search space being different (model outputs are not averaged; logit gaps are used instead of max probability). I will try putting gomea in this same search space, it shouldn't be worse.}

\begin{figure}
  \vspace{10pt}
  \centering
  \begin{subfigure}[t]{0.43\linewidth}
        \centering
        \includegraphics[width=\linewidth]{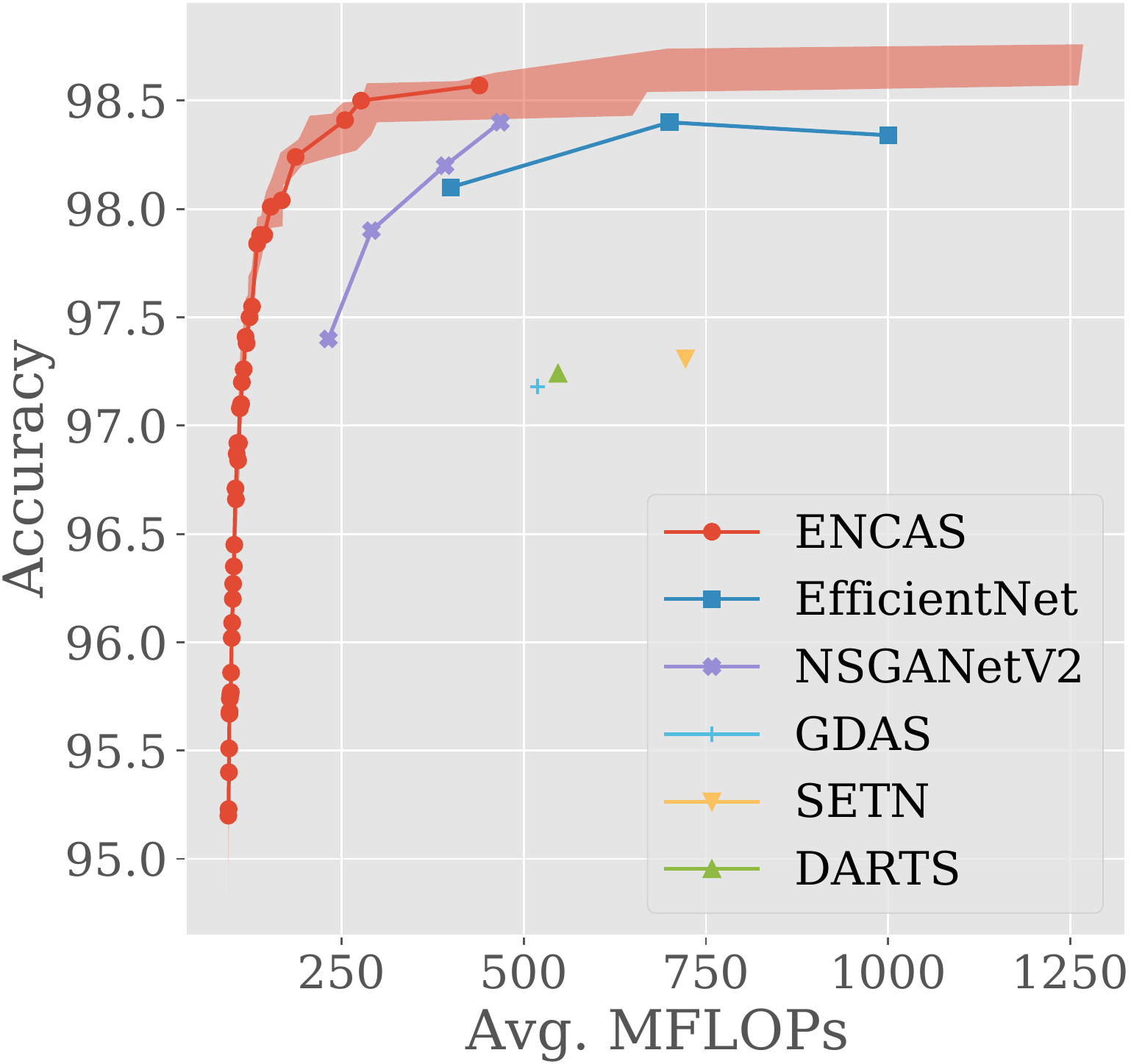}
        \vspace{-14pt}
        \caption{CIFAR-10}
        \vspace{4pt}
    \end{subfigure}%
    ~ 
    \begin{subfigure}[t]{0.43\linewidth}
        \centering
        \includegraphics[width=\linewidth]{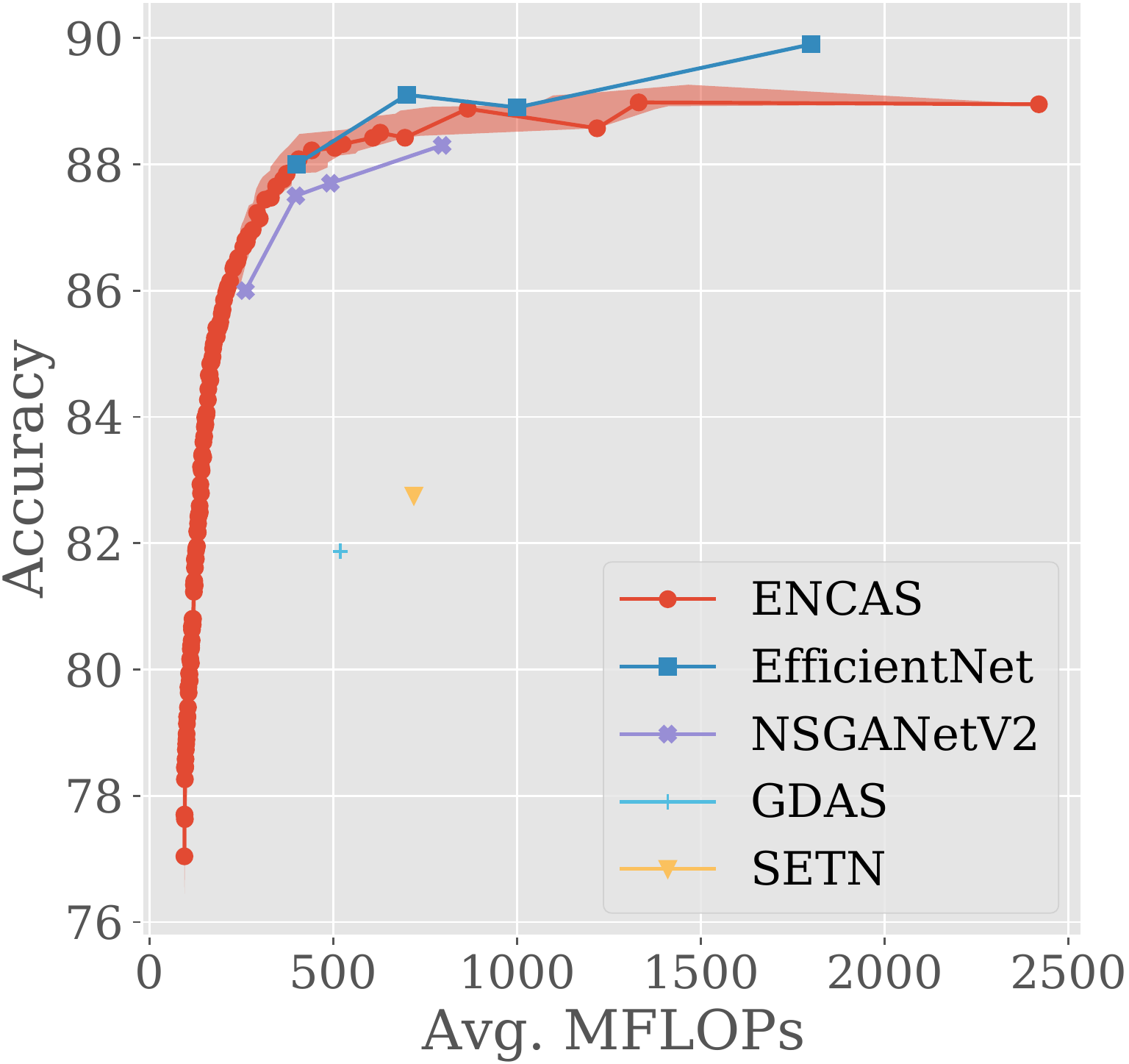}
        \vspace{-14pt}
        \caption{CIFAR-100}
        \vspace{4pt}
    \end{subfigure}
    
    \begin{subfigure}[t]{0.43\linewidth}
        \centering
        \includegraphics[width=\linewidth]{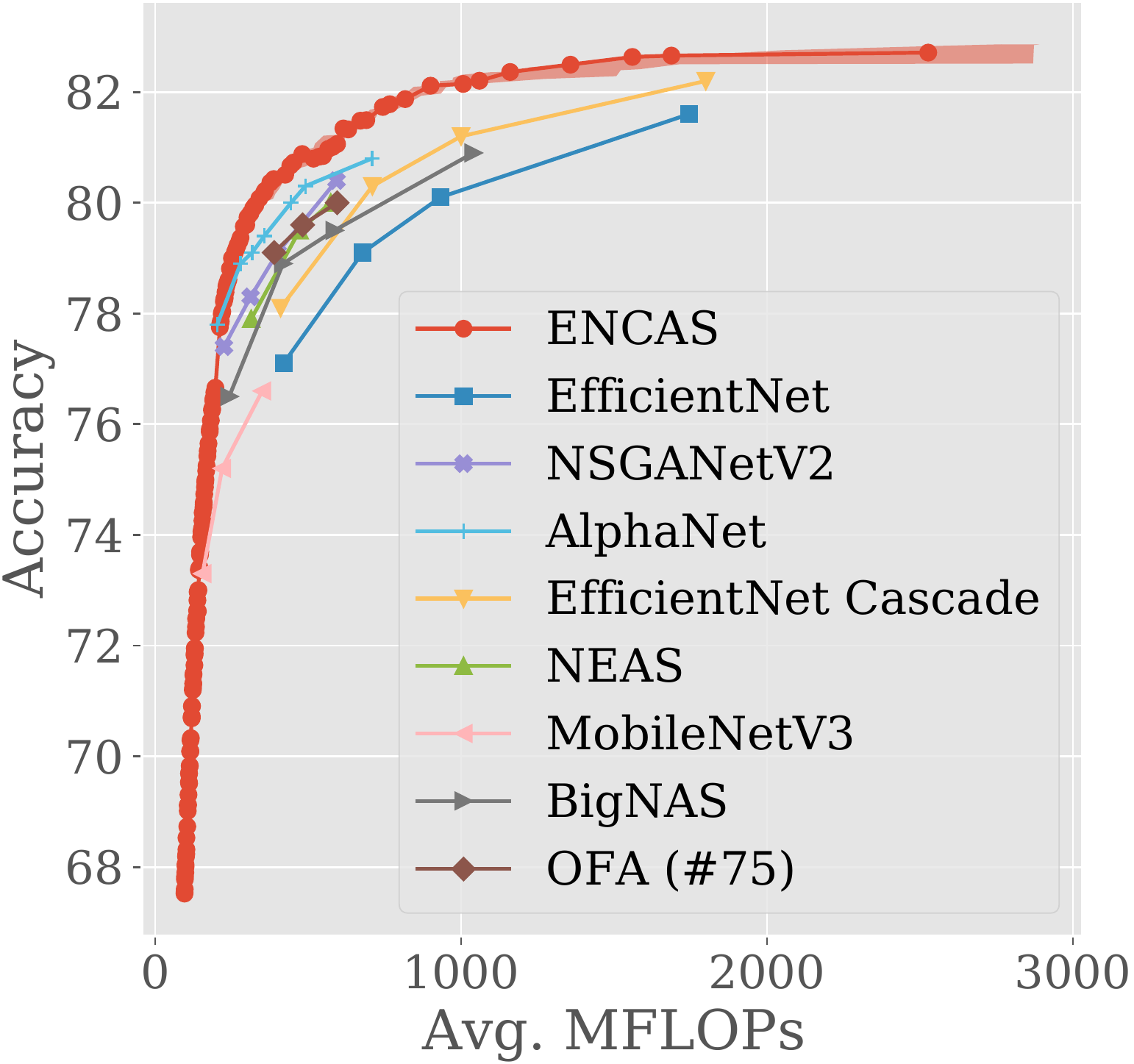}
        \vspace{-14pt}
        \caption{ImageNet}
    \end{subfigure}
\vspace{-5pt}
  \caption{Comparing ENCAS with SOTA NAS algorithms.}
%   \vspace{-10pt}
  \Description{}
  \label{fig:exp:cmp_sota}
\end{figure}

Our results are also compared with those of previous efficient NAS algorithms (see Tables~\ref{tab:imagenet},~\ref{tab:cifar100},~\ref{tab:cifar10}). Fig.~\ref{fig:exp:cmp_sota} shows that the trade-off fronts produced by ENCAS dominate other NAS approaches under 1.5 GFLOPs across the datasets. But it can also be seen that for CIFAR-100 while ENCAS is on par with EfficientNet-B0 to B2, it is outperformed by B3, even though the supernetworks we use outperform EfficientNet B0 to B3 on ImageNet. This may occur because training an individual network is much easier than training a supernetwork (in our experience, training a supernetwork is hard due to subnetworks having to share both weights and hyperparameters).

Note that we do not compare search times of different algorithms because the corresponding publications often report times that are not comparable due to e.g. using different hardware, not accounting for supernetwork training or final network retraining time. A fair comparison would require us to run all the algorithms, for which we lack compute. For future reference, the runtime associated with each part of our pipeline is mentioned in the section describing it.
% \begin{figure*}
%   \centering
%   \begin{subfigure}[t]{0.25\textwidth}
%         \centering
%         \includegraphics[width=\textwidth]{_pics/4_experiments/cmp_sota_cifar10.pdf}
%         \caption{CIFAR-10}
%     \end{subfigure}%
%     ~ 
%     \begin{subfigure}[t]{0.25\textwidth}
%         \centering
%         \includegraphics[width=\textwidth]{_pics/4_experiments/cmp_sota_cifar100.pdf}
%         \caption{CIFAR-100}
%     \end{subfigure}
%     ~ 
%     \begin{subfigure}[t]{0.25\textwidth}
%         \centering
%         \includegraphics[width=\textwidth]{_pics/4_experiments/cmp_sota_imagenet.pdf}
%         \caption{ImageNet}
%     \end{subfigure}
%   \caption{Comparison of ENCAS with SOTA efficient NAS algorithms}
%   \Description{}
%   \label{fig:exp:cmp_sota}
% \end{figure*}

% \begin{figure*}
%   \centering
%   \begin{subfigure}[t]{0.3\textwidth}
%         \centering
%         \includegraphics[width=\linewidth]{_pics/4_experiments/cmp_sota_cifar10.pdf}
%         \vspace{-14pt}
%         \caption{CIFAR-10}
%         \vspace{4pt}
%     \end{subfigure}%
%     ~ 
%     \begin{subfigure}[t]{0.3\textwidth}
%         \centering
%         \includegraphics[width=\linewidth]{_pics/4_experiments/cmp_sota_cifar100.pdf}
%         \vspace{-14pt}
%         \caption{CIFAR-100}
%         \vspace{4pt}
%     \end{subfigure}
%     ~
%     \begin{subfigure}[t]{0.303\textwidth}
%         \centering
%         \includegraphics[width=\linewidth]{_pics/4_experiments/cmp_sota_imagenet.pdf}
%         \vspace{-14pt}
%         \caption{ImageNet}
%     \end{subfigure}
% \vspace{-5pt}
%   \caption{Comparing ENCAS with SOTA NAS algorithms.}
% %   \vspace{-10pt}
%   \Description{}
%   \label{fig:exp:cmp_sota}
% \end{figure*}

\begin{table}
  \caption{CIFAR-10 performance, "acc." is top-1 accuracy. The method producing the highest accuracy is in bold.
  }
  \vspace{-4pt}
  \label{tab:cifar10}
  \begin{tabular}{lccc}
    \toprule
    \multicolumn{1}{c}{Method}& \makecell{Hyper-\\volume} & \makecell{Max\\acc.} & \makecell{Max\\MFLOPs}\\
    \midrule
    \makecell[l]{EfficientNet B0-B2~\cite{tan2019efficientnet}}  & 0.863 & 98.4 & 1000\\
    \makecell[l]{NSGANetV2~\cite{lu2021neural}}  & 0.904 & 98.4 & 468\\
    \makecell[l]{GDAS~\cite{dong2019searching}}  & --- & 97.18 & 519\\
    \makecell[l]{SETN~\cite{dong2019one}}  & --- & 97.31 & 722\\
    \makecell[l]{DARTS~\cite{liu2018darts}}  & --- & 97.24 & 547\\
    \makecell[l]{NAT (reproduced)~\cite{lu2021neural}}  & 0.899$_{\pm0.003}$ & 96.80$_{\pm0.13}$ & 361$_{\pm119}$\\
    \makecell[l]{NAT (best)~\cite{lu2021neural}}  & 0.911$_{\pm0.002}$ & 98.46$_{\pm0.08}$ & 1390$_{\pm274}$\\
    \makecell[l]{GreedyCascade~\cite{streeter2018approximation}}  & 0.935$_{\pm0.002}$  & 98.31$_{\pm0.05}$ & 191$_{\pm16}$\\
    \midrule
    ENCAS (1 supernet) & 0.911$_{\pm0.002}$ & 98.45$_{\pm0.09}$ & 698$_{\pm321}$\\
    ENCAS (5 supernets) & 0.941$_{\pm0.002}$ & 98.60$_{\pm0.09}$ & 749$_{\pm298}$\\
    \midrule
    ENCAS-joint & 0.935$_{\pm0.002}$ & 98.68$_{\pm0.04}$ & 4858$_{\pm1126}$\\
    \textbf{ENCAS-joint+} &\textbf{ 0.943$_{\pm0.002}$} & \textbf{98.68$_{\pm0.08}$} & 1060$_{\pm444}$\\
  \bottomrule
\end{tabular}
\end{table}

\subsection{Joint training and cascade search}

\begin{figure}[h]
  \vspace{7pt}
  \centering
    \begin{subfigure}[t]{0.5\linewidth}
        \centering
        \includegraphics[width=\textwidth]{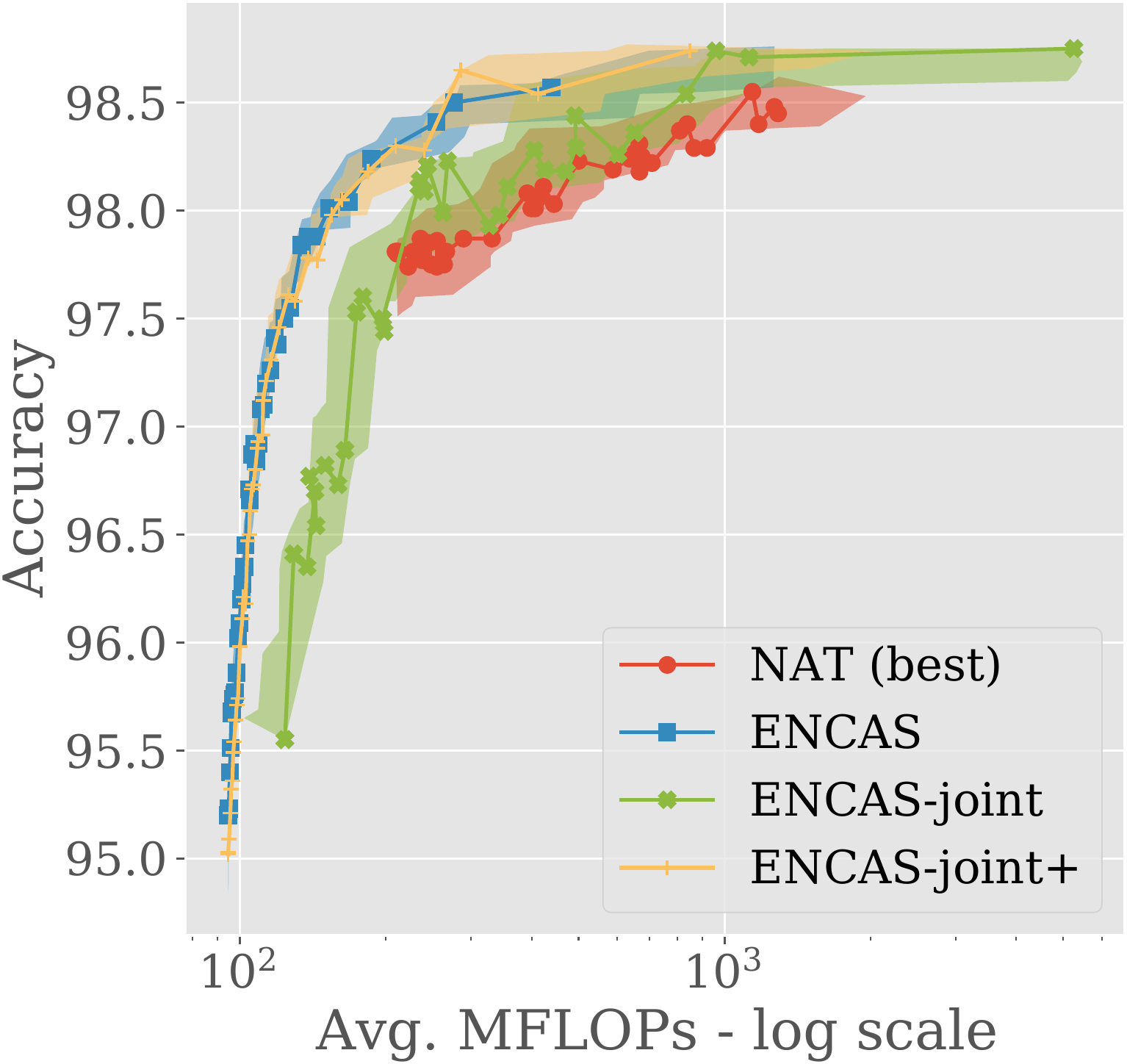}
        % \vspace{-14pt}
        \vspace{-11pt}
        \caption{CIFAR-10}
    \end{subfigure}%
    ~ 
    \begin{subfigure}[t]{0.48\linewidth}
        \centering
        \includegraphics[width=\linewidth]{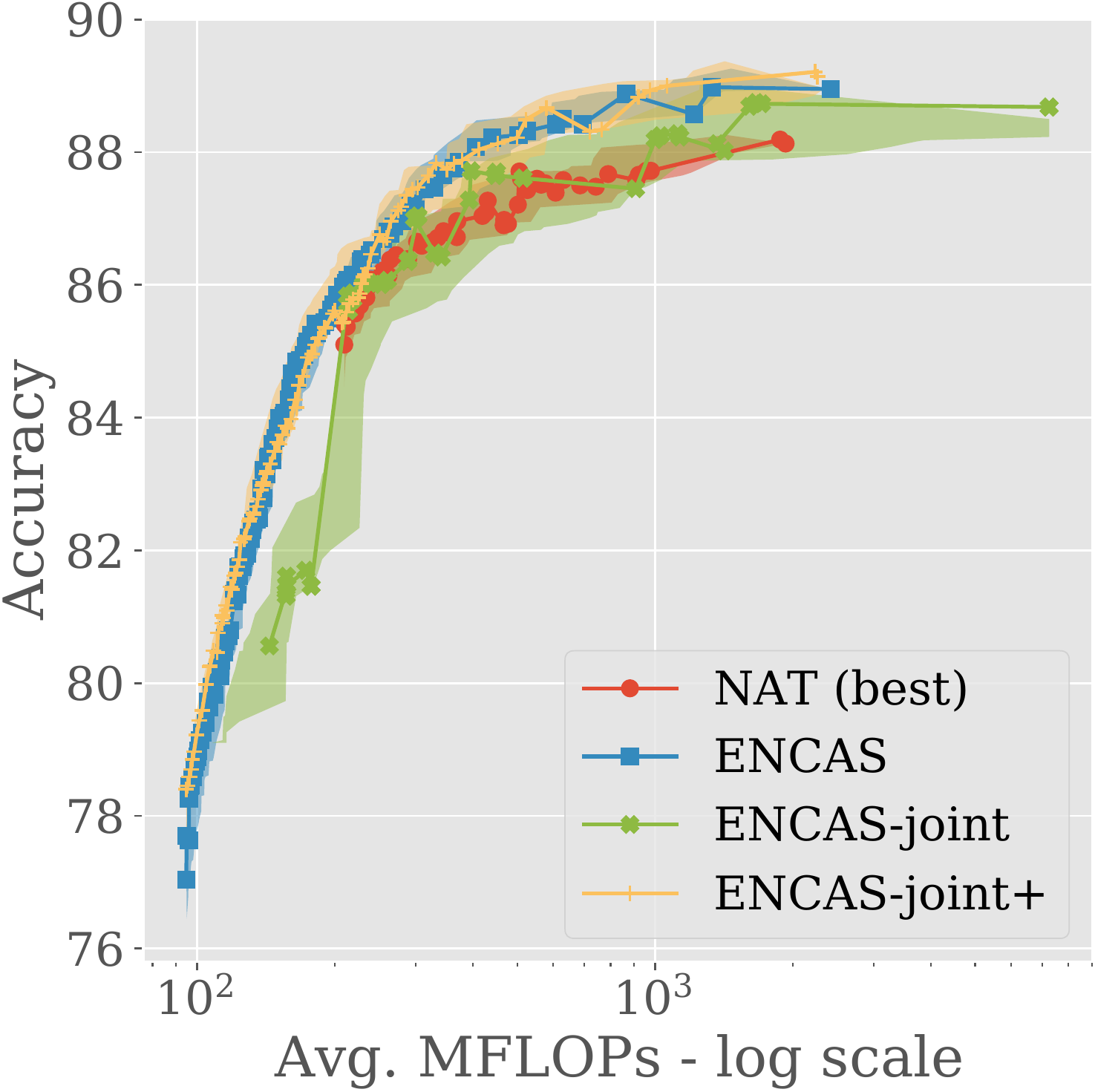}
        % \vspace{-14pt}
        \vspace{-11pt}
        \caption{CIFAR-100}
    \end{subfigure}
  \vspace{-5pt}
  \caption{Investigating benefits of joint training and search.}
  \vspace{-9pt}
  \Description{}
  \label{fig:exp:joint}
\end{figure}

Is joint weight training and search of cascade architectures beneficial? In Fig.~\ref{fig:exp:joint} we can see that ENCAS-joint finds a trade-off front that is worse than the one found by ENCAS. This likely happens due to the increased size of the search space. However, the trade-off front found by ENCAS-joint is better than the best NAT one.

We further see that ENCAS-joint+ (running ENCAS on the supernetworks trained by ENCAS-joint) improves upon ENCAS-joint on both CIFAR-10 and CIFAR-100. But is it better than running ENCAS on separately trained supernetworks? Although ENCAS-joint+ appears to outperform ENCAS in terms of hypervolume and maximum accuracy (see Tables~\ref{tab:cifar100},~\ref{tab:cifar10}), these results are not statistically significant.

Note that these experiments are not performed on ImageNet due to limitations in computational power. Training supernetworks jointly is computationally intensive in general (search time of ENCAS-joint is 240 GPU-hours) while also lacking flexibility, as adding or removing a network means restarting the whole process from scratch. Given inconclusiveness of improvements brought by ENCAS-joint+, we recommend using ENCAS and separate training of supernetworks (see Section~\ref{disc} for further discussion).

\subsection{Applying ENCAS to SOTA ImageNet models}

ENCAS relies on the architectures discovered via supernetwork-based NAS. However, using a supernetwork means that these architectures are necessarily not very large. Because of this, NAS results are typically evaluated in the context of a mobile phone setting, which is usually taken to mean $\leq 600$ MFLOPs.

However, nowadays there are hundreds of large well-performing ImageNet-pretrained models available online. Can our good results be extended from the mobile phone setting to dominating the complete trade-off front? To answer this question, we take 518 ImageNet models from the Pytorch Image Models (\texttt{timm})~\cite{rw2019timm} library, and run our search procedure on them. 

\begin{figure}[h]
  \centering
  \includegraphics[width=0.5\linewidth]{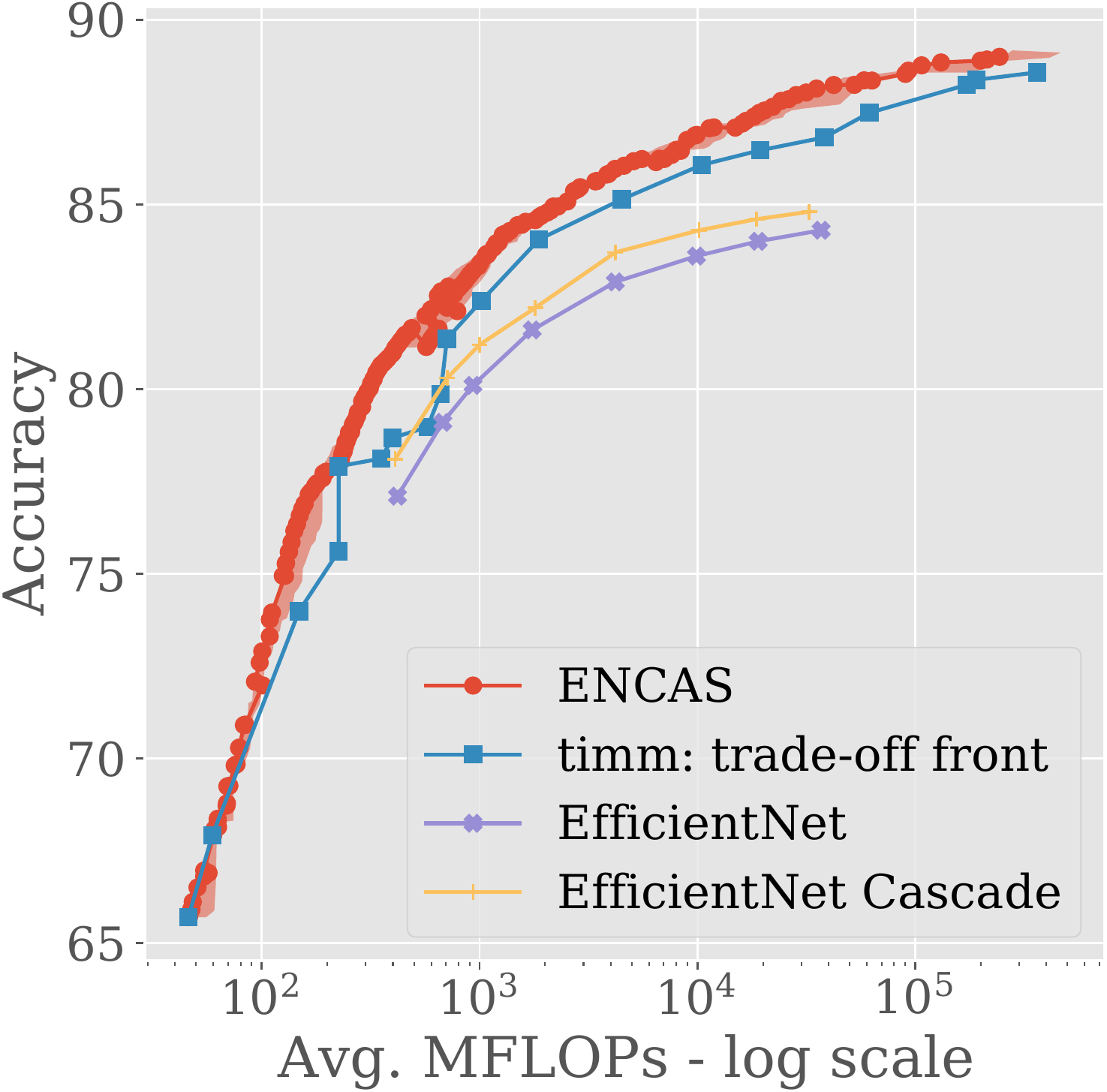}
  \caption{ENCAS discovers a dominating trade-off front on ImageNet by searching for cascades of 518 \texttt{timm} models (from which only the models on the trade-off front are shown).}
  \Description{}
  \label{fig:exp:imagenetsota}
\end{figure}

As shown in Figure~\ref{fig:exp:imagenetsota}, this indeed leads to a dominating trade-off front. The increase of 0.4 percentage points in the maximum ImageNet performance leads to our largest cascade achieving the highest ImageNet accuracy of publicly available models (89.01$_{\pm0.10}$), while simultaneously decreasing FLOPs by 18\% (from 362 GFLOPs to 296$_{\pm77}$ GFLOPs). Our cascades outperform those in~\cite{wang2020multiple}, in large part thanks to the ability of our algorithm to use a search space containing hundreds of models, which is not feasible for the exhaustive search approach used in~\cite{wang2020multiple}.

\section{Additional experiments} \label{abl}

In this section we further investigate the impact of using more than one supernetwork. Due to space constraints, a comparison to ensembles is provided in Appendix~\ref{appendix:ensembles} and a comparison to random search is provided in Appendix~\ref{abl:random}.

\subsection{Impact of increasing the number of supernetworks}

\begin{figure}[h]
  \centering
  \begin{subfigure}[t]{0.45\linewidth}
        \centering
        \includegraphics[width=\textwidth]{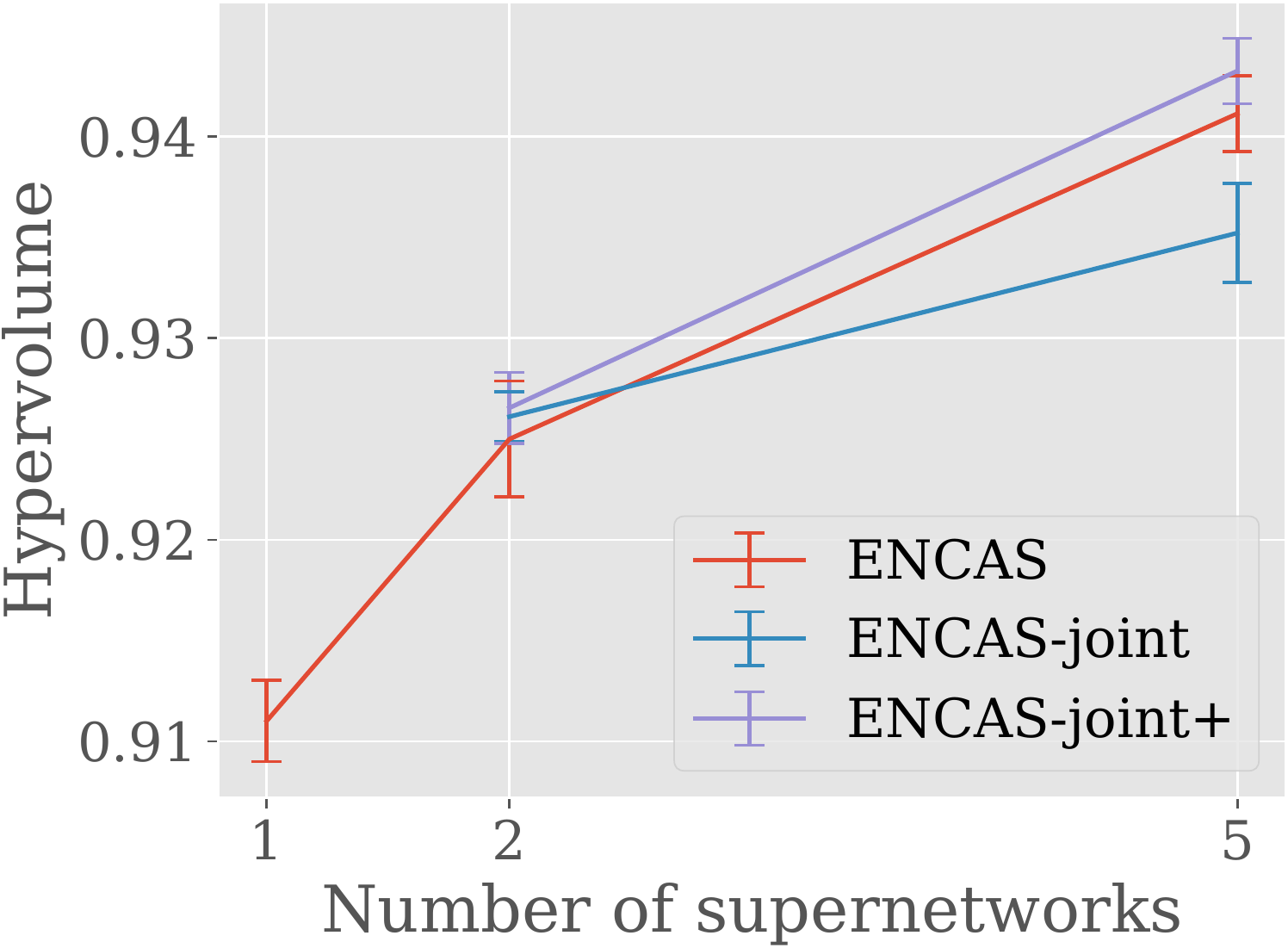}
        \caption{CIFAR-10}
    \end{subfigure}%
    ~ 
    \begin{subfigure}[t]{0.45\linewidth}
        \centering
        \includegraphics[width=\linewidth]{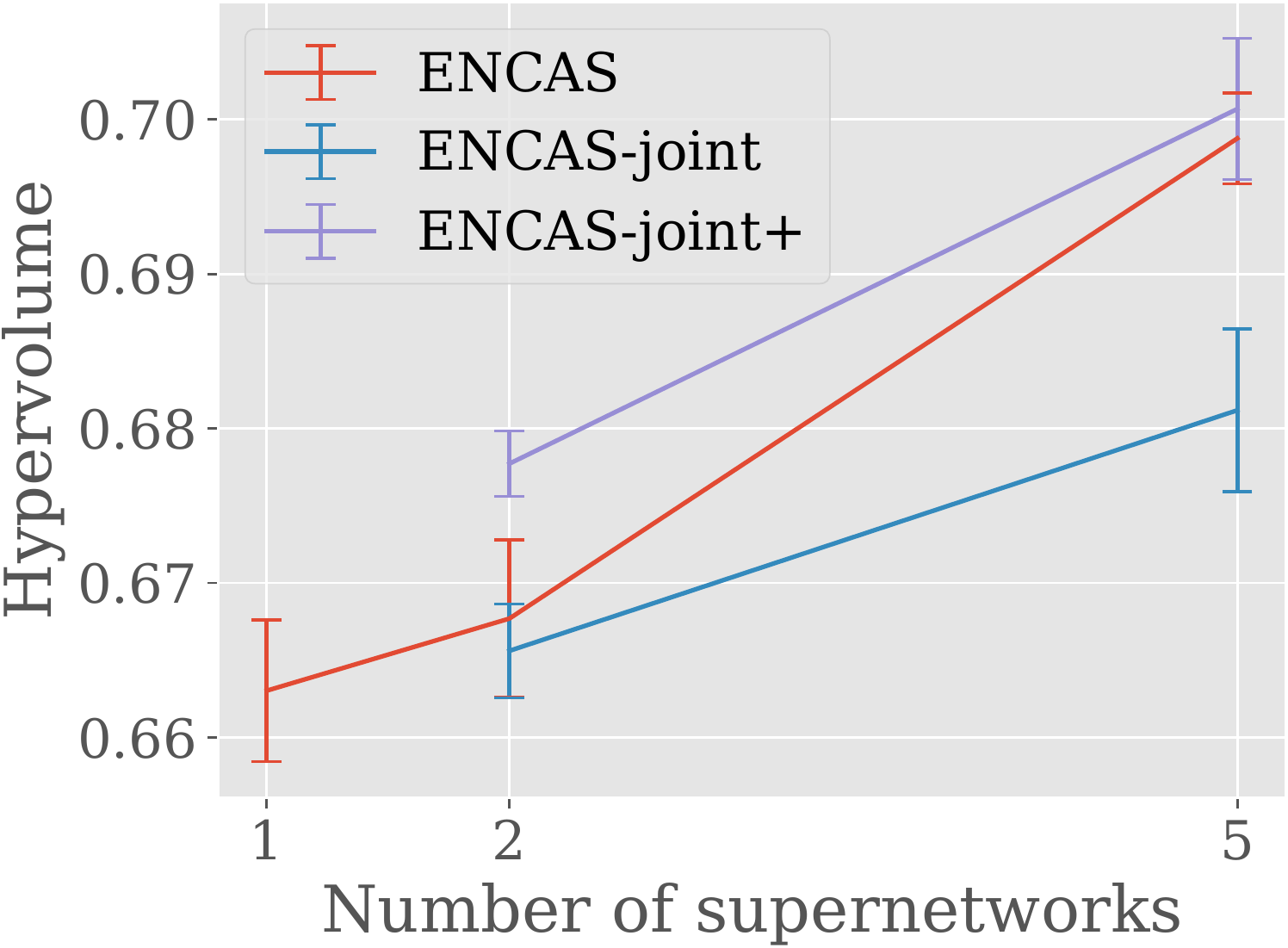}
        \caption{CIFAR-100}
    \end{subfigure}
  \caption{Impact of increasing the number of supernetworks in ENCAS, ENCAS-joint, ENCAS-joint+.}
  \Description{}
  \label{fig:abl:moresupernets}
\end{figure}

Figure~\ref{fig:abl:moresupernets} shows the impact of increasing the number of supernetworks used in ENCAS from 1 to 2 to 5. A trend of increasing hypervolume can be clearly observed.

For joint training (ENCAS-joint), the hypervolume also increases, but not as much. This can be explained by the increase in the search space that every additional supernetwork brings. Interestingly, the hypervolume obtained with ENCAS-joint+ grows about as fast as with ENCAS, which we interpret to mean that the joint training and search over an increasing number of supernetworks is beneficial for weights while harmful for the simultaneous search (given the same search budget). The larger search space may require a larger search budget to achieve better results, and therefore ENCAS-joint may have higher potential to benefit from more compute. Running ENCAS using the weights trained by ENCAS-joint (i.e. ENCAS-joint+) realizes the benefit of better weights and ameliorates the downside of a larger search space.

\subsection{Is using different supernetworks better than using the best one trained several times?}

Experiments in section~\ref{exp:cas_many} demonstrated that using several supernetworks is better than using just the best one. But is the source of the effect the diversities of architectures found, or just the increased quantity of networks with different weights? To answer this question, we train (via NAT) the best supernetwork 5 times with different seeds on CIFAR-10 and CIFAR-100 and apply ENCAS to the resulting trade-off fronts.

\begin{figure}[h]
  \centering
  \mbox{}\hfill 
  \begin{subfigure}[c]{0.6\linewidth}
        \centering
        \includegraphics[width=0.9\linewidth]{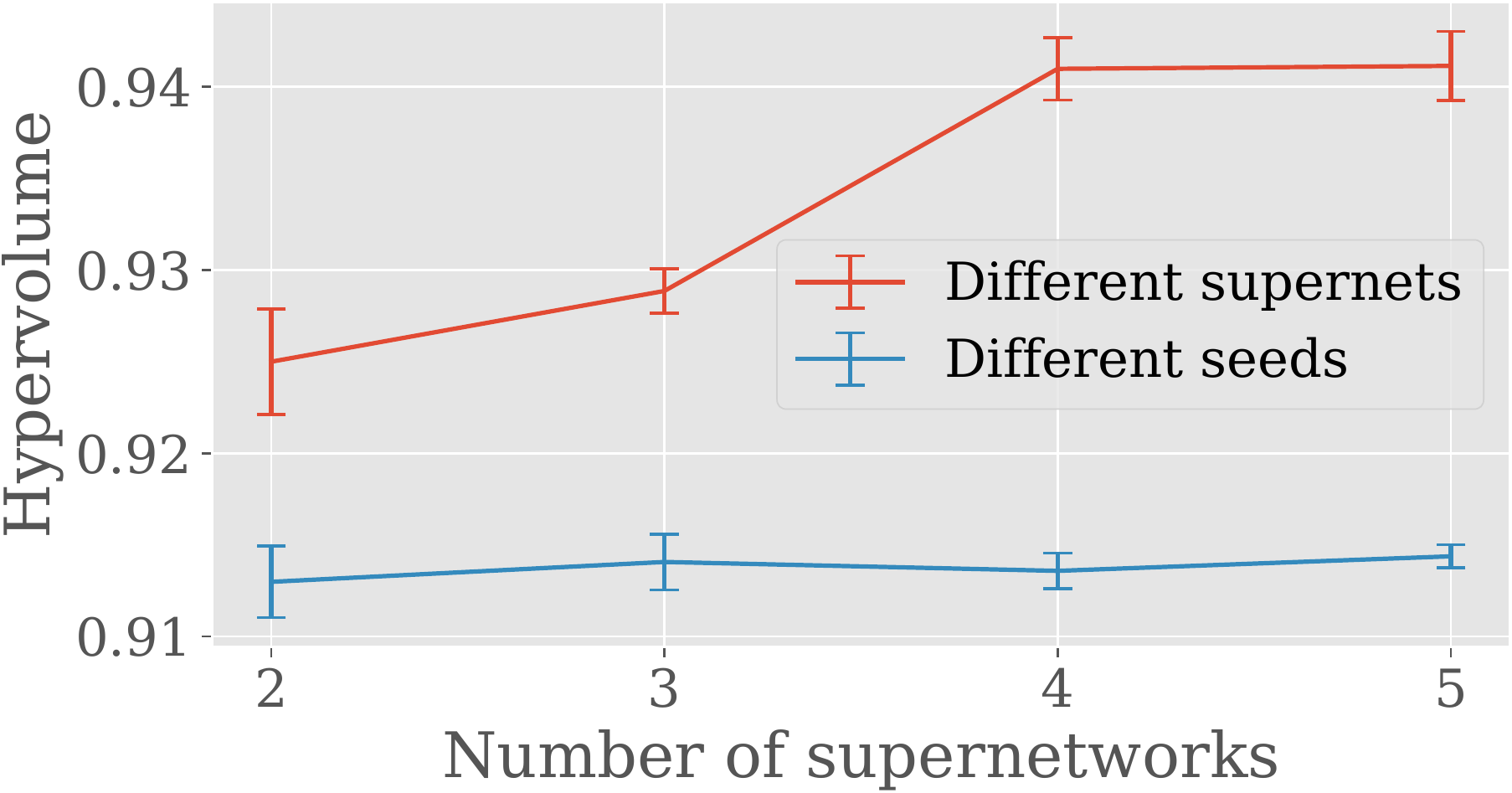}
        \\
        \includegraphics[width=0.9\linewidth]{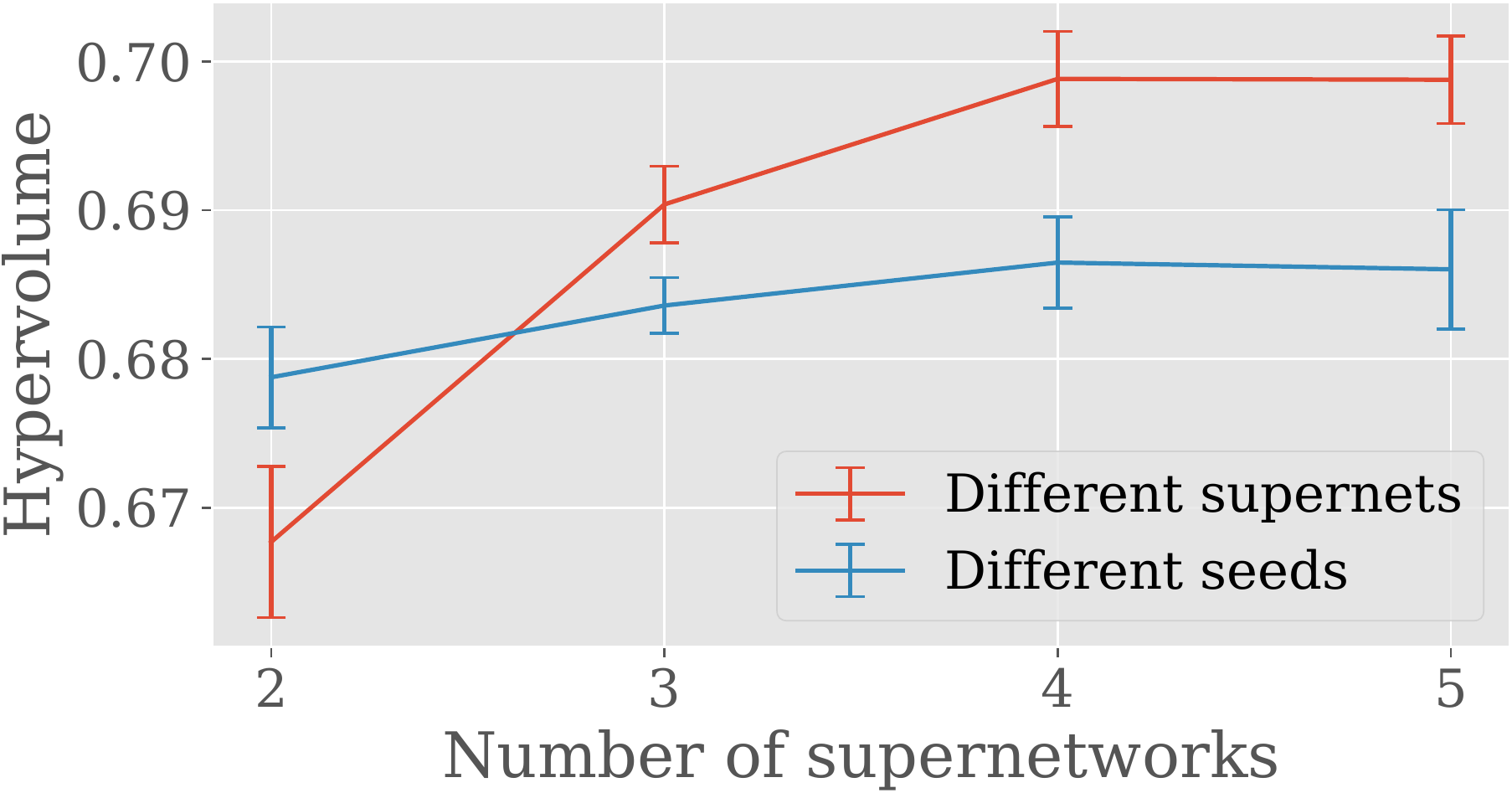}
        % \caption{CIFAR-100 (2 seeds)}
    \end{subfigure}
    \hfill
    \begin{subfigure}[c]{0.38\linewidth}
        \centering
        \includegraphics[width=\linewidth]{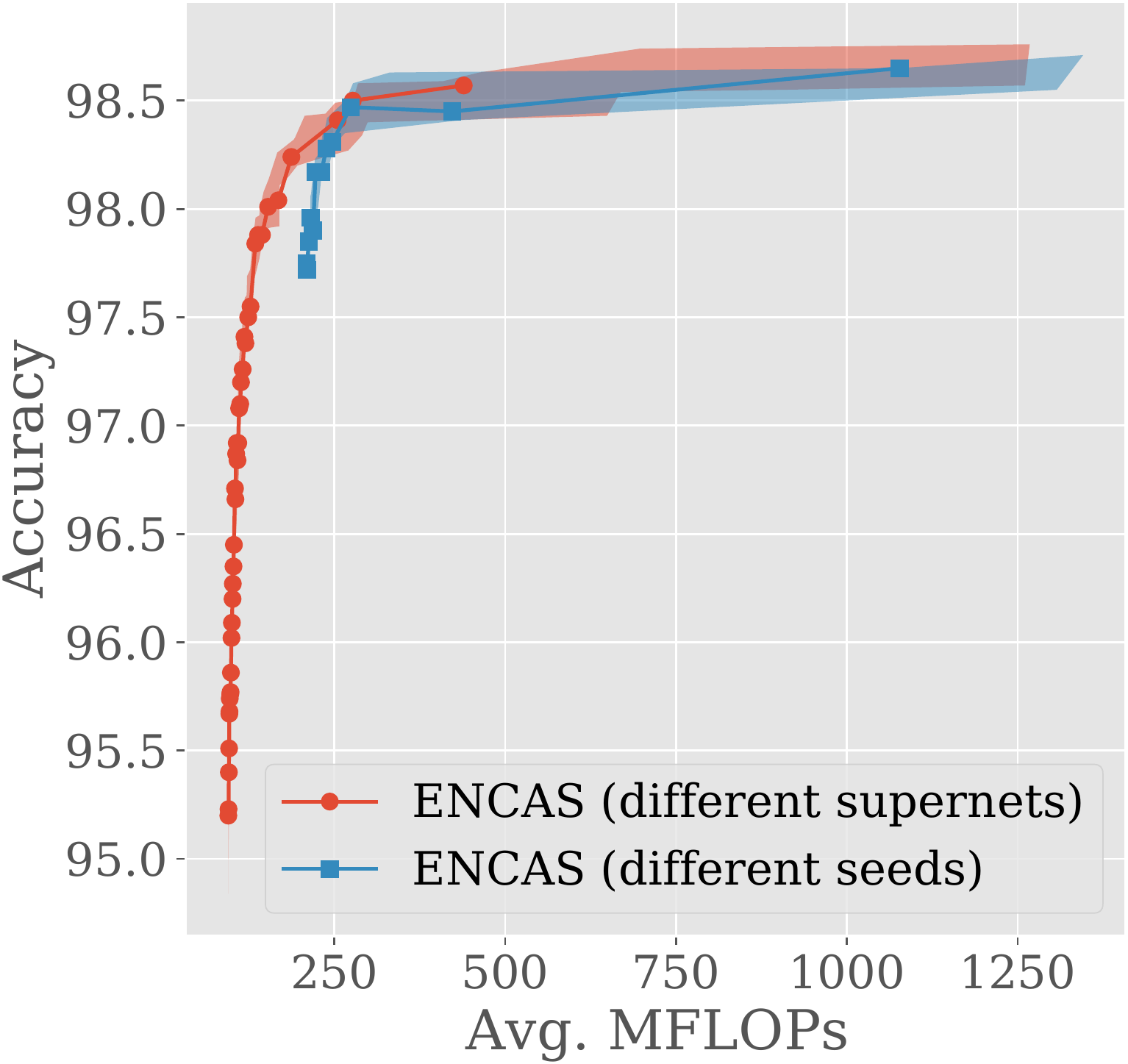}
        % \caption{CIFAR-100 (2 seeds)}
    \end{subfigure}
  \hfill\mbox{}
  \caption{Comparison between hypervolume when using different supernetworks, or the best supernetwork trained with different seeds on \textit{(top left)} CIFAR-10, \textit{(bottom left)} CIFAR-100. \textit{(right)} shows comparison between trade-off fronts of 5 supernetworks, or 5 seeds of the best supernetwork on CIFAR-10.}
  \Description{}
  \label{fig:abl:cloneshv}
\end{figure}

In Fig.~\ref{fig:abl:cloneshv} (left) we see that using different runs of the best supernetwork barely increases hypervolume, in contrast to using different supernetworks. If we inspect the trade-off fronts in Fig.~\ref{fig:abl:cloneshv} (right) for 5 supernetworks and for 5 seeds of the best supernetwork on CIFAR-10, we can see that the difference is in low-FLOPs models that are missing from the best supernetwork but are present in other supernetworks. Therefore, using diverse supernetworks is helpful for obtaining a larger trade-off front coverage. However, on the side of the most accurate networks, using multiple restarts of NAT with the best supernetwork is sufficient to get close to the best performance, unlike what could be expected, since the diversity in architectures and weights is arguably lower.

\section{Discussion} \label{disc}

While our approach achieves good results with limited resource usage, it still suffers from limitations. Notably, it relies on pretrained supernetworks, which are currently not very diverse, architecture-wise. Additionally, these supernetworks need to allow extraction and usage of subnetworks without retraining in order for our approach to work, which limits their selection even further.

In our experiments, we find that performing search after the supernetwork weights have been adapted is not much worse than joint training and search. This can mean that there is not a lot of benefit to be gained by fine-tuning architecture choices of different cascade components to each other; alternatively, perhaps ENCAS-joint was simply not able to realize these benefits, for instance because it may require more computational resources than we used in our experiments.

This paper has demonstrated the benefits brought by the usage of cascades. This reinforces the main thesis of~\cite{wang2020multiple}: researchers should pay more attention to cascades. However, the warnings of~\cite{wang2020multiple} should also be repeated, as they apply to any cascade approach, including ours: the decrease in FLOPs brought by cascades can be realized either when processing images one-by-one, or when processing a large amount of images offline. The benefits are not realized in online batch processing: once a batch has been created, due to the parallel nature of GPU accelerators, processing a part of a batch takes approximately the same resources as processing the whole batch.

\section{Conclusion} \label{conclusion}

In this paper, we considered the automatic creation of cascades of deep neural networks. We developed an effective algorithm called ENCAS that builds upon the literature on efficient NAS by searching for cascades across pretrained supernetworks either simultaneously with weight training or after weight training. ENCAS is the first NAS algorithm that searches for cascade architectures. It does so by solving the multi-objective optimization problem of finding well-performing small cascades with the help of an EA (MO-GOMEA). 

ENCAS was found to outperform SOTA efficient NAS approaches on several image classification datasets. Its search procedure can also be applied to an arbitrary model pool. By applying it to well-performing publicly available ImageNet models, we achieved a dominating trade-off front on ImageNet.

Finally, we find that searching for neural network architectures in more than one pretrained supernetwork is beneficial despite the limited diversity of the currently available supernetworks, which is expected to only increase with time.

\begin{acks}

We thank Arkadiy Dushatskiy for insightful discussions, Marco Virgolin for introducing us to~\cite{wang2020multiple}, Zhichao Lu for sharing his knowledge about NAT and parts of the NAT codebase.

This work is part of the research project DAEDALUS funded via the Open Technology Programme of the Netherlands Organization for Scientific Research (NWO), project number 18373; part of the funding is provided by Elekta and ORTEC LogiqCare. This work was carried out (in part) on the Dutch national e-infrastructure with the support of SURF Cooperative.

\end{acks}

\bibliographystyle{ACM-Reference-Format}
\bibliography{_refs}


\begin{thebibliography}{67}

%%% the Web address.

\ifx \showCODEN    \undefined \def \showCODEN     #1{\unskip}     \fi
\ifx \showDOI      \undefined \def \showDOI       #1{#1}\fi
\ifx \showISBNx    \undefined \def \showISBNx     #1{\unskip}     \fi
\ifx \showISBNxiii \undefined \def \showISBNxiii  #1{\unskip}     \fi
\ifx \showISSN     \undefined \def \showISSN      #1{\unskip}     \fi
\ifx \showLCCN     \undefined \def \showLCCN      #1{\unskip}     \fi
\ifx \shownote     \undefined \def \shownote      #1{#1}          \fi
\ifx \showarticletitle \undefined \def \showarticletitle #1{#1}   \fi
\ifx \showURL      \undefined \def \showURL       {\relax}        \fi
% invisible to TeX
\providecommand\bibfield[2]{#2}
\providecommand\bibinfo[2]{#2}
\providecommand\natexlab[1]{#1}
\providecommand\showeprint[2][]{arXiv:#2}

\bibitem[\protect\citeauthoryear{Angelova, Krizhevsky, Vanhoucke, Ogale, and
  Ferguson}{Angelova et~al\mbox{.}}{2015}]%
        {angelova2015real}
\bibfield{author}{\bibinfo{person}{Anelia Angelova}, \bibinfo{person}{Alex
  Krizhevsky}, \bibinfo{person}{Vincent Vanhoucke}, \bibinfo{person}{Abhijit
  Ogale}, {and} \bibinfo{person}{Dave Ferguson}.}
  \bibinfo{year}{2015}\natexlab{}.
\newblock \showarticletitle{Real-Time Pedestrian Detection With Deep Network
  Cascades}. In \bibinfo{booktitle}{\emph{Proceedings of BMVC 2015}}.
\newblock


\bibitem[\protect\citeauthoryear{Bahdanau, Cho, and Bengio}{Bahdanau
  et~al\mbox{.}}{2014}]%
        {bahdanau2014neural}
\bibfield{author}{\bibinfo{person}{Dzmitry Bahdanau},
  \bibinfo{person}{Kyunghyun Cho}, {and} \bibinfo{person}{Yoshua Bengio}.}
  \bibinfo{year}{2014}\natexlab{}.
\newblock \showarticletitle{Neural machine translation by jointly learning to
  align and translate}.
\newblock \bibinfo{journal}{\emph{arXiv preprint arXiv:1409.0473}}
  (\bibinfo{year}{2014}).
\newblock


\bibitem[\protect\citeauthoryear{Baker, Gupta, Raskar, and Naik}{Baker
  et~al\mbox{.}}{2017}]%
        {baker2017accelerating}
\bibfield{author}{\bibinfo{person}{Bowen Baker}, \bibinfo{person}{Otkrist
  Gupta}, \bibinfo{person}{Ramesh Raskar}, {and} \bibinfo{person}{Nikhil
  Naik}.} \bibinfo{year}{2017}\natexlab{}.
\newblock \showarticletitle{Accelerating neural architecture search using
  performance prediction}.
\newblock \bibinfo{journal}{\emph{arXiv preprint arXiv:1705.10823}}
  (\bibinfo{year}{2017}).
\newblock


\bibitem[\protect\citeauthoryear{{Blank}, {Deb}, {Dhebar}, {Bandaru}, and
  {Seada}}{{Blank} et~al\mbox{.}}{2020}]%
        {ref_dirs_energy}
\bibfield{author}{\bibinfo{person}{J. {Blank}}, \bibinfo{person}{K. {Deb}},
  \bibinfo{person}{Y. {Dhebar}}, \bibinfo{person}{S. {Bandaru}}, {and}
  \bibinfo{person}{H. {Seada}}.} \bibinfo{year}{2020}\natexlab{}.
\newblock \showarticletitle{Generating Well-Spaced Points on a Unit Simplex for
  Evolutionary Many-Objective Optimization}.
\newblock \bibinfo{journal}{\emph{IEEE Transactions on Evolutionary
  Computation}} \bibinfo{volume}{25}, \bibinfo{number}{1}
  (\bibinfo{year}{2020}), \bibinfo{pages}{48--60}.
\newblock


\bibitem[\protect\citeauthoryear{Bouter, Alderliesten, Pieters, Bel,
  Niatsetski, and Bosman}{Bouter et~al\mbox{.}}{2019}]%
        {bouter2019gpu}
\bibfield{author}{\bibinfo{person}{Anton Bouter}, \bibinfo{person}{Tanja
  Alderliesten}, \bibinfo{person}{Bradley~R Pieters}, \bibinfo{person}{Arjan
  Bel}, \bibinfo{person}{Yury Niatsetski}, {and} \bibinfo{person}{Peter A~N
  Bosman}.} \bibinfo{year}{2019}\natexlab{}.
\newblock \showarticletitle{GPU-accelerated bi-objective treatment planning for
  prostate high-dose-rate brachytherapy}.
\newblock \bibinfo{journal}{\emph{Medical Physics}} \bibinfo{volume}{46},
  \bibinfo{number}{9} (\bibinfo{year}{2019}), \bibinfo{pages}{3776--3787}.
\newblock


\bibitem[\protect\citeauthoryear{Branke and Schmeck}{Branke and
  Schmeck}{2003}]%
        {branke2003designing}
\bibfield{author}{\bibinfo{person}{J{\"u}rgen Branke} {and}
  \bibinfo{person}{Hartmut Schmeck}.} \bibinfo{year}{2003}\natexlab{}.
\newblock \showarticletitle{Designing evolutionary algorithms for dynamic
  optimization problems}.
\newblock In \bibinfo{booktitle}{\emph{Advances in Evolutionary Computing}}.
  \bibinfo{publisher}{Springer}, \bibinfo{pages}{239--262}.
\newblock


\bibitem[\protect\citeauthoryear{Broomhead and Lowe}{Broomhead and
  Lowe}{1988}]%
        {broomhead1988multivariable}
\bibfield{author}{\bibinfo{person}{David~S Broomhead} {and}
  \bibinfo{person}{David Lowe}.} \bibinfo{year}{1988}\natexlab{}.
\newblock \bibinfo{booktitle}{\emph{Radial basis functions, multi-variable
  functional interpolation and adaptive networks}}.
\newblock \bibinfo{type}{{T}echnical {R}eport}. \bibinfo{institution}{Royal
  Signals and Radar Establishment Malvern (United Kingdom)}.
\newblock


\bibitem[\protect\citeauthoryear{Brown, Mann, Ryder, Subbiah, Kaplan, Dhariwal,
  Neelakantan, Shyam, Sastry, Askell, et~al\mbox{.}}{Brown
  et~al\mbox{.}}{2020}]%
        {brown2020language}
\bibfield{author}{\bibinfo{person}{Tom Brown}, \bibinfo{person}{Benjamin Mann},
  \bibinfo{person}{Nick Ryder}, \bibinfo{person}{Melanie Subbiah},
  \bibinfo{person}{Jared~D Kaplan}, \bibinfo{person}{Prafulla Dhariwal},
  \bibinfo{person}{Arvind Neelakantan}, \bibinfo{person}{Pranav Shyam},
  \bibinfo{person}{Girish Sastry}, \bibinfo{person}{Amanda Askell},
  {et~al\mbox{.}}} \bibinfo{year}{2020}\natexlab{}.
\newblock \showarticletitle{Language models are few-shot learners}.
\newblock \bibinfo{journal}{\emph{Advances in Neural Information Processing
  Systems}}  \bibinfo{volume}{33} (\bibinfo{year}{2020}),
  \bibinfo{pages}{1877--1901}.
\newblock


\bibitem[\protect\citeauthoryear{Cai, Gan, Wang, Zhang, and Han}{Cai
  et~al\mbox{.}}{2020}]%
        {cai2019once}
\bibfield{author}{\bibinfo{person}{Han Cai}, \bibinfo{person}{Chuang Gan},
  \bibinfo{person}{Tianzhe Wang}, \bibinfo{person}{Zhekai Zhang}, {and}
  \bibinfo{person}{Song Han}.} \bibinfo{year}{2020}\natexlab{}.
\newblock \showarticletitle{Once for All: Train One Network and Specialize it
  for Efficient Deployment}. In \bibinfo{booktitle}{\emph{International
  Conference on Learning Representations}}.
\newblock


\bibitem[\protect\citeauthoryear{Cai, Zhu, and Han}{Cai et~al\mbox{.}}{2019}]%
        {cai2018proxylessnas}
\bibfield{author}{\bibinfo{person}{Han Cai}, \bibinfo{person}{Ligeng Zhu},
  {and} \bibinfo{person}{Song Han}.} \bibinfo{year}{2019}\natexlab{}.
\newblock \showarticletitle{Proxyless{NAS}: Direct Neural Architecture Search
  on Target Task and Hardware}. In \bibinfo{booktitle}{\emph{International
  Conference on Learning Representations}}.
\newblock
\urldef\tempurl%
\url{https://openreview.net/forum?id=HylVB3AqYm}
\showURL{%
\tempurl}


\bibitem[\protect\citeauthoryear{Cai, Saberian, and Vasconcelos}{Cai
  et~al\mbox{.}}{2015}]%
        {cai2015learning}
\bibfield{author}{\bibinfo{person}{Zhaowei Cai}, \bibinfo{person}{Mohammad
  Saberian}, {and} \bibinfo{person}{Nuno Vasconcelos}.}
  \bibinfo{year}{2015}\natexlab{}.
\newblock \showarticletitle{Learning complexity-aware cascades for deep
  pedestrian detection}. In \bibinfo{booktitle}{\emph{Proceedings of the IEEE
  International Conference on Computer Vision}}. \bibinfo{pages}{3361--3369}.
\newblock


\bibitem[\protect\citeauthoryear{Chen, Fu, and Ling}{Chen
  et~al\mbox{.}}{2021}]%
        {chen2021one}
\bibfield{author}{\bibinfo{person}{Minghao Chen}, \bibinfo{person}{Jianlong
  Fu}, {and} \bibinfo{person}{Haibin Ling}.} \bibinfo{year}{2021}\natexlab{}.
\newblock \showarticletitle{One-Shot Neural Ensemble Architecture Search by
  Diversity-Guided Search Space Shrinking}. In
  \bibinfo{booktitle}{\emph{Proceedings of the IEEE/CVF Conference on Computer
  Vision and Pattern Recognition}}. \bibinfo{pages}{16530--16539}.
\newblock


\bibitem[\protect\citeauthoryear{Chiong, Weise, and Michalewicz}{Chiong
  et~al\mbox{.}}{2012}]%
        {chiong2012variants}
\bibfield{author}{\bibinfo{person}{Raymond Chiong}, \bibinfo{person}{Thomas
  Weise}, {and} \bibinfo{person}{Zbigniew Michalewicz}.}
  \bibinfo{year}{2012}\natexlab{}.
\newblock \bibinfo{booktitle}{\emph{Variants of evolutionary algorithms for
  real-world applications}}.
\newblock \bibinfo{publisher}{Springer}.
\newblock


\bibitem[\protect\citeauthoryear{Cubuk, Zoph, Shlens, and Le}{Cubuk
  et~al\mbox{.}}{2020}]%
        {cubuk2020randaugment}
\bibfield{author}{\bibinfo{person}{Ekin~D Cubuk}, \bibinfo{person}{Barret
  Zoph}, \bibinfo{person}{Jonathon Shlens}, {and} \bibinfo{person}{Quoc~V Le}.}
  \bibinfo{year}{2020}\natexlab{}.
\newblock \showarticletitle{RandAugment: Practical automated data augmentation
  with a reduced search space}. In \bibinfo{booktitle}{\emph{Proceedings of the
  IEEE/CVF Conference on Computer Vision and Pattern Recognition Workshops}}.
  \bibinfo{pages}{702--703}.
\newblock


\bibitem[\protect\citeauthoryear{Das and Dennis}{Das and Dennis}{1998}]%
        {das1998normal}
\bibfield{author}{\bibinfo{person}{Indraneel Das} {and} \bibinfo{person}{John~E
  Dennis}.} \bibinfo{year}{1998}\natexlab{}.
\newblock \showarticletitle{Normal-boundary intersection: A new method for
  generating the Pareto surface in nonlinear multicriteria optimization
  problems}.
\newblock \bibinfo{journal}{\emph{SIAM Journal on Optimization}}
  \bibinfo{volume}{8}, \bibinfo{number}{3} (\bibinfo{year}{1998}),
  \bibinfo{pages}{631--657}.
\newblock


\bibitem[\protect\citeauthoryear{Deb and Jain}{Deb and Jain}{2013}]%
        {deb2013evolutionary}
\bibfield{author}{\bibinfo{person}{Kalyanmoy Deb} {and}
  \bibinfo{person}{Himanshu Jain}.} \bibinfo{year}{2013}\natexlab{}.
\newblock \showarticletitle{An evolutionary many-objective optimization
  algorithm using reference-point-based nondominated sorting approach, part I:
  solving problems with box constraints}.
\newblock \bibinfo{journal}{\emph{IEEE Transactions on Evolutionary
  Computation}} \bibinfo{volume}{18}, \bibinfo{number}{4}
  (\bibinfo{year}{2013}), \bibinfo{pages}{577--601}.
\newblock


\bibitem[\protect\citeauthoryear{DeVries and Taylor}{DeVries and
  Taylor}{2017}]%
        {devries2017improved}
\bibfield{author}{\bibinfo{person}{Terrance DeVries} {and}
  \bibinfo{person}{Graham~W Taylor}.} \bibinfo{year}{2017}\natexlab{}.
\newblock \showarticletitle{Improved regularization of convolutional neural
  networks with cutout}.
\newblock \bibinfo{journal}{\emph{arXiv preprint arXiv:1708.04552}}
  (\bibinfo{year}{2017}).
\newblock


\bibitem[\protect\citeauthoryear{Donahue, Jia, Vinyals, Hoffman, Zhang, Tzeng,
  and Darrell}{Donahue et~al\mbox{.}}{2014}]%
        {donahue2014decaf}
\bibfield{author}{\bibinfo{person}{Jeff Donahue}, \bibinfo{person}{Yangqing
  Jia}, \bibinfo{person}{Oriol Vinyals}, \bibinfo{person}{Judy Hoffman},
  \bibinfo{person}{Ning Zhang}, \bibinfo{person}{Eric Tzeng}, {and}
  \bibinfo{person}{Trevor Darrell}.} \bibinfo{year}{2014}\natexlab{}.
\newblock \showarticletitle{Decaf: A deep convolutional activation feature for
  generic visual recognition}. In \bibinfo{booktitle}{\emph{International
  Conference on Machine Learning}}. PMLR, \bibinfo{pages}{647--655}.
\newblock


\bibitem[\protect\citeauthoryear{Dong and Yang}{Dong and Yang}{2019a}]%
        {dong2019one}
\bibfield{author}{\bibinfo{person}{Xuanyi Dong} {and} \bibinfo{person}{Yi
  Yang}.} \bibinfo{year}{2019}\natexlab{a}.
\newblock \showarticletitle{One-shot neural architecture search via
  self-evaluated template network}. In \bibinfo{booktitle}{\emph{Proceedings of
  the IEEE/CVF International Conference on Computer Vision}}.
  \bibinfo{pages}{3681--3690}.
\newblock


\bibitem[\protect\citeauthoryear{Dong and Yang}{Dong and Yang}{2019b}]%
        {dong2019searching}
\bibfield{author}{\bibinfo{person}{Xuanyi Dong} {and} \bibinfo{person}{Yi
  Yang}.} \bibinfo{year}{2019}\natexlab{b}.
\newblock \showarticletitle{Searching for a robust neural architecture in four
  gpu hours}. In \bibinfo{booktitle}{\emph{Proceedings of the IEEE/CVF
  Conference on Computer Vision and Pattern Recognition}}.
  \bibinfo{pages}{1761--1770}.
\newblock


\bibitem[\protect\citeauthoryear{Dunn}{Dunn}{1961}]%
        {dunn1961multiple}
\bibfield{author}{\bibinfo{person}{Olive~Jean Dunn}.}
  \bibinfo{year}{1961}\natexlab{}.
\newblock \showarticletitle{Multiple comparisons among means}.
\newblock \bibinfo{journal}{\emph{J. Amer. Statist. Assoc.}}
  \bibinfo{volume}{56}, \bibinfo{number}{293} (\bibinfo{year}{1961}),
  \bibinfo{pages}{52--64}.
\newblock


\bibitem[\protect\citeauthoryear{Esposito and Saitta}{Esposito and
  Saitta}{2003}]%
        {esposito2003monte}
\bibfield{author}{\bibinfo{person}{Roberto Esposito} {and}
  \bibinfo{person}{Lorenza Saitta}.} \bibinfo{year}{2003}\natexlab{}.
\newblock \showarticletitle{Monte Carlo theory as an explanation of bagging and
  boosting}. In \bibinfo{booktitle}{\emph{IJCAI}}, Vol.~\bibinfo{volume}{3}.
  \bibinfo{pages}{499--504}.
\newblock


\bibitem[\protect\citeauthoryear{Howard, Sandler, Chu, Chen, Chen, Tan, Wang,
  Zhu, Pang, Vasudevan, et~al\mbox{.}}{Howard et~al\mbox{.}}{2019}]%
        {howard2019searching}
\bibfield{author}{\bibinfo{person}{Andrew Howard}, \bibinfo{person}{Mark
  Sandler}, \bibinfo{person}{Grace Chu}, \bibinfo{person}{Liang-Chieh Chen},
  \bibinfo{person}{Bo Chen}, \bibinfo{person}{Mingxing Tan},
  \bibinfo{person}{Weijun Wang}, \bibinfo{person}{Yukun Zhu},
  \bibinfo{person}{Ruoming Pang}, \bibinfo{person}{Vijay Vasudevan},
  {et~al\mbox{.}}} \bibinfo{year}{2019}\natexlab{}.
\newblock \showarticletitle{Searching for MobileNetv3}. In
  \bibinfo{booktitle}{\emph{Proceedings of the IEEE/CVF International
  Conference on Computer Vision}}. \bibinfo{pages}{1314--1324}.
\newblock


\bibitem[\protect\citeauthoryear{Isensee, Jaeger, Kohl, Petersen, and
  Maier-Hein}{Isensee et~al\mbox{.}}{2021}]%
        {isensee2018nnu}
\bibfield{author}{\bibinfo{person}{Fabian Isensee}, \bibinfo{person}{Paul~F
  Jaeger}, \bibinfo{person}{Simon~AA Kohl}, \bibinfo{person}{Jens Petersen},
  {and} \bibinfo{person}{Klaus~H Maier-Hein}.} \bibinfo{year}{2021}\natexlab{}.
\newblock \showarticletitle{nnU-Net: a self-configuring method for deep
  learning-based biomedical image segmentation}.
\newblock \bibinfo{journal}{\emph{Nature methods}} \bibinfo{volume}{18},
  \bibinfo{number}{2} (\bibinfo{year}{2021}), \bibinfo{pages}{203--211}.
\newblock


\bibitem[\protect\citeauthoryear{Izmailov, Podoprikhin, Garipov, Vetrov, and
  Wilson}{Izmailov et~al\mbox{.}}{2018}]%
        {izmailov2018averaging}
\bibfield{author}{\bibinfo{person}{Pavel Izmailov}, \bibinfo{person}{Dmitrii
  Podoprikhin}, \bibinfo{person}{Timur Garipov}, \bibinfo{person}{Dmitry
  Vetrov}, {and} \bibinfo{person}{Andrew~Gordon Wilson}.}
  \bibinfo{year}{2018}\natexlab{}.
\newblock \showarticletitle{Averaging weights leads to wider optima and better
  generalization}. In \bibinfo{booktitle}{\emph{34th Conference on Uncertainty
  in Artificial Intelligence 2018, UAI 2018}}. Association For Uncertainty in
  Artificial Intelligence, \bibinfo{pages}{876--885}.
\newblock


\bibitem[\protect\citeauthoryear{Kornblith, Shlens, and Le}{Kornblith
  et~al\mbox{.}}{2019}]%
        {kornblith2019better}
\bibfield{author}{\bibinfo{person}{Simon Kornblith}, \bibinfo{person}{Jonathon
  Shlens}, {and} \bibinfo{person}{Quoc~V Le}.} \bibinfo{year}{2019}\natexlab{}.
\newblock \showarticletitle{Do better ImageNet models transfer better?}. In
  \bibinfo{booktitle}{\emph{Proceedings of the IEEE/CVF Conference on Computer
  Vision and Pattern Recognition}}. \bibinfo{pages}{2661--2671}.
\newblock


\bibitem[\protect\citeauthoryear{Krizhevsky, Hinton, et~al\mbox{.}}{Krizhevsky
  et~al\mbox{.}}{2009}]%
        {krizhevsky2009learning}
\bibfield{author}{\bibinfo{person}{Alex Krizhevsky}, \bibinfo{person}{Geoffrey
  Hinton}, {et~al\mbox{.}}} \bibinfo{year}{2009}\natexlab{}.
\newblock \bibinfo{booktitle}{\emph{Learning multiple layers of features from
  tiny images}}.
\newblock \bibinfo{type}{{T}echnical {R}eport}.
\newblock


\bibitem[\protect\citeauthoryear{Kukenys and McCane}{Kukenys and
  McCane}{2008}]%
        {kukenys2008classifier}
\bibfield{author}{\bibinfo{person}{Ignas Kukenys} {and}
  \bibinfo{person}{Brendan McCane}.} \bibinfo{year}{2008}\natexlab{}.
\newblock \showarticletitle{Classifier cascades for support vector machines}.
  In \bibinfo{booktitle}{\emph{2008 23rd International Conference Image and
  Vision Computing New Zealand}}. IEEE, \bibinfo{pages}{1--6}.
\newblock


\bibitem[\protect\citeauthoryear{LeCun, Bengio, et~al\mbox{.}}{LeCun
  et~al\mbox{.}}{1995}]%
        {lecun1995convolutional}
\bibfield{author}{\bibinfo{person}{Yann LeCun}, \bibinfo{person}{Yoshua
  Bengio}, {et~al\mbox{.}}} \bibinfo{year}{1995}\natexlab{}.
\newblock \showarticletitle{Convolutional networks for images, speech, and time
  series}.
\newblock \bibinfo{journal}{\emph{The handbook of brain theory and neural
  networks}} \bibinfo{volume}{3361}, \bibinfo{number}{10}
  (\bibinfo{year}{1995}), \bibinfo{pages}{1995}.
\newblock


\bibitem[\protect\citeauthoryear{Li and Talwalkar}{Li and Talwalkar}{2020}]%
        {li2020random}
\bibfield{author}{\bibinfo{person}{Liam Li} {and} \bibinfo{person}{Ameet
  Talwalkar}.} \bibinfo{year}{2020}\natexlab{}.
\newblock \showarticletitle{Random Search and Reproducibility for Neural
  Architecture Search}. In \bibinfo{booktitle}{\emph{Proceedings of The 35th
  Uncertainty in Artificial Intelligence Conference}}
  \emph{(\bibinfo{series}{Proceedings of Machine Learning Research},
  Vol.~\bibinfo{volume}{115})}, \bibfield{editor}{\bibinfo{person}{Ryan~P.
  Adams} {and} \bibinfo{person}{Vibhav Gogate}} (Eds.).
  \bibinfo{publisher}{PMLR}, \bibinfo{pages}{367--377}.
\newblock
\urldef\tempurl%
\url{https://proceedings.mlr.press/v115/li20c.html}
\showURL{%
\tempurl}


\bibitem[\protect\citeauthoryear{Liu, Simonyan, and Yang}{Liu
  et~al\mbox{.}}{2018}]%
        {liu2018darts}
\bibfield{author}{\bibinfo{person}{Hanxiao Liu}, \bibinfo{person}{Karen
  Simonyan}, {and} \bibinfo{person}{Yiming Yang}.}
  \bibinfo{year}{2018}\natexlab{}.
\newblock \showarticletitle{DARTS: Differentiable architecture search}.
\newblock \bibinfo{journal}{\emph{arXiv preprint arXiv:1806.09055}}
  (\bibinfo{year}{2018}).
\newblock


\bibitem[\protect\citeauthoryear{Loshchilov and Hutter}{Loshchilov and
  Hutter}{2016}]%
        {loshchilov2016sgdr}
\bibfield{author}{\bibinfo{person}{Ilya Loshchilov} {and}
  \bibinfo{person}{Frank Hutter}.} \bibinfo{year}{2016}\natexlab{}.
\newblock \showarticletitle{SGDR: Stochastic gradient descent with warm
  restarts}.
\newblock \bibinfo{journal}{\emph{arXiv preprint arXiv:1608.03983}}
  (\bibinfo{year}{2016}).
\newblock


\bibitem[\protect\citeauthoryear{Lu, Deb, Goodman, Banzhaf, and Boddeti}{Lu
  et~al\mbox{.}}{2020}]%
        {lu2020nsganetv2}
\bibfield{author}{\bibinfo{person}{Zhichao Lu}, \bibinfo{person}{Kalyanmoy
  Deb}, \bibinfo{person}{Erik Goodman}, \bibinfo{person}{Wolfgang Banzhaf},
  {and} \bibinfo{person}{Vishnu~Naresh Boddeti}.}
  \bibinfo{year}{2020}\natexlab{}.
\newblock \showarticletitle{NSGANetv2: Evolutionary multi-objective
  surrogate-assisted neural architecture search}. In
  \bibinfo{booktitle}{\emph{European Conference on Computer Vision}}. Springer,
  \bibinfo{pages}{35--51}.
\newblock


\bibitem[\protect\citeauthoryear{Lu, Sreekumar, Goodman, Banzhaf, Deb, and
  Boddeti}{Lu et~al\mbox{.}}{2021}]%
        {lu2021neural}
\bibfield{author}{\bibinfo{person}{Zhichao Lu}, \bibinfo{person}{Gautam
  Sreekumar}, \bibinfo{person}{Erik Goodman}, \bibinfo{person}{Wolfgang
  Banzhaf}, \bibinfo{person}{Kalyanmoy Deb}, {and}
  \bibinfo{person}{Vishnu~Naresh Boddeti}.} \bibinfo{year}{2021}\natexlab{}.
\newblock \showarticletitle{Neural architecture transfer}.
\newblock \bibinfo{journal}{\emph{IEEE Transactions on Pattern Analysis and
  Machine Intelligence}} \bibinfo{volume}{43}, \bibinfo{number}{9}
  (\bibinfo{year}{2021}), \bibinfo{pages}{2971--2989}.
\newblock


\bibitem[\protect\citeauthoryear{Luong, Grond, La~Poutr{\'e}, and Bosman}{Luong
  et~al\mbox{.}}{2015}]%
        {luong2015scalable}
\bibfield{author}{\bibinfo{person}{Ngoc~Hoang Luong},
  \bibinfo{person}{Marinus~OW Grond}, \bibinfo{person}{Han La~Poutr{\'e}},
  {and} \bibinfo{person}{Peter A~N Bosman}.} \bibinfo{year}{2015}\natexlab{}.
\newblock \showarticletitle{Scalable and practical multi-objective distribution
  network expansion planning}. In \bibinfo{booktitle}{\emph{2015 IEEE Power \&
  Energy Society General Meeting}}. IEEE, \bibinfo{pages}{1--5}.
\newblock


\bibitem[\protect\citeauthoryear{Luong, La~Poutr{\'e}, and Bosman}{Luong
  et~al\mbox{.}}{2014}]%
        {luong2014multi}
\bibfield{author}{\bibinfo{person}{Ngoc~Hoang Luong}, \bibinfo{person}{Han
  La~Poutr{\'e}}, {and} \bibinfo{person}{Peter A~N Bosman}.}
  \bibinfo{year}{2014}\natexlab{}.
\newblock \showarticletitle{Multi-objective gene-pool optimal mixing
  evolutionary algorithms}. In \bibinfo{booktitle}{\emph{Proceedings of the
  2014 Annual Conference on Genetic and Evolutionary Computation}}.
  \bibinfo{pages}{357--364}.
\newblock


\bibitem[\protect\citeauthoryear{Narayanan, Zela, Saikia, Brox, and
  Hutter}{Narayanan et~al\mbox{.}}{2021}]%
        {narayanan2021multi}
\bibfield{author}{\bibinfo{person}{Ashwin~Raaghav Narayanan},
  \bibinfo{person}{Arber Zela}, \bibinfo{person}{Tonmoy Saikia},
  \bibinfo{person}{Thomas Brox}, {and} \bibinfo{person}{Frank Hutter}.}
  \bibinfo{year}{2021}\natexlab{}.
\newblock \showarticletitle{Multi-headed Neural Ensemble Search}. In
  \bibinfo{booktitle}{\emph{Workshop on Uncertainty and Robustness in Deep
  Learning (UDL@ICML`21)}}.
\newblock


\bibitem[\protect\citeauthoryear{Nesterov}{Nesterov}{1983}]%
        {nesterov1983method}
\bibfield{author}{\bibinfo{person}{Yurii~E Nesterov}.}
  \bibinfo{year}{1983}\natexlab{}.
\newblock \showarticletitle{A method for solving the convex programming problem
  with convergence rate O (1/$k^2$)}. In \bibinfo{booktitle}{\emph{Dokl. akad.
  nauk SSSR}}, Vol.~\bibinfo{volume}{269}. \bibinfo{pages}{543--547}.
\newblock


\bibitem[\protect\citeauthoryear{Petchrompo, Wannakrairot, and
  Parlikad}{Petchrompo et~al\mbox{.}}{2022}]%
        {petchrompo2022pruning}
\bibfield{author}{\bibinfo{person}{Sanyapong Petchrompo},
  \bibinfo{person}{Anupong Wannakrairot}, {and} \bibinfo{person}{Ajith~Kumar
  Parlikad}.} \bibinfo{year}{2022}\natexlab{}.
\newblock \showarticletitle{Pruning pareto optimal solutions for
  multi-objective portfolio asset management}.
\newblock \bibinfo{journal}{\emph{European Journal of Operational Research}}
  \bibinfo{volume}{297}, \bibinfo{number}{1} (\bibinfo{year}{2022}),
  \bibinfo{pages}{203--220}.
\newblock


\bibitem[\protect\citeauthoryear{Pham, Guan, Zoph, Le, and Dean}{Pham
  et~al\mbox{.}}{2018}]%
        {pham2018efficient}
\bibfield{author}{\bibinfo{person}{Hieu Pham}, \bibinfo{person}{Melody Guan},
  \bibinfo{person}{Barret Zoph}, \bibinfo{person}{Quoc Le}, {and}
  \bibinfo{person}{Jeff Dean}.} \bibinfo{year}{2018}\natexlab{}.
\newblock \showarticletitle{Efficient neural architecture search via parameters
  sharing}. In \bibinfo{booktitle}{\emph{International Conference on Machine
  Learning}}. PMLR, \bibinfo{pages}{4095--4104}.
\newblock


\bibitem[\protect\citeauthoryear{Real, Aggarwal, Huang, and Le}{Real
  et~al\mbox{.}}{2019}]%
        {real2019regularized}
\bibfield{author}{\bibinfo{person}{Esteban Real}, \bibinfo{person}{Alok
  Aggarwal}, \bibinfo{person}{Yanping Huang}, {and} \bibinfo{person}{Quoc~V
  Le}.} \bibinfo{year}{2019}\natexlab{}.
\newblock \showarticletitle{Regularized evolution for image classifier
  architecture search}. In \bibinfo{booktitle}{\emph{Proceedings of the AAAI
  Conference on Artificial Intelligence}}, Vol.~\bibinfo{volume}{33}.
  \bibinfo{pages}{4780--4789}.
\newblock


\bibitem[\protect\citeauthoryear{Recht, Roelofs, Schmidt, and Shankar}{Recht
  et~al\mbox{.}}{2019}]%
        {recht2019imagenet}
\bibfield{author}{\bibinfo{person}{Benjamin Recht}, \bibinfo{person}{Rebecca
  Roelofs}, \bibinfo{person}{Ludwig Schmidt}, {and} \bibinfo{person}{Vaishaal
  Shankar}.} \bibinfo{year}{2019}\natexlab{}.
\newblock \showarticletitle{Do ImageNet classifiers generalize to ImageNet?}.
  In \bibinfo{booktitle}{\emph{International Conference on Machine Learning}}.
  PMLR, \bibinfo{pages}{5389--5400}.
\newblock


\bibitem[\protect\citeauthoryear{Rodrigues, Bauer, and Bosman}{Rodrigues
  et~al\mbox{.}}{2016}]%
        {rodrigues2016multi}
\bibfield{author}{\bibinfo{person}{S Rodrigues}, \bibinfo{person}{Pavol Bauer},
  {and} \bibinfo{person}{Peter A~N Bosman}.} \bibinfo{year}{2016}\natexlab{}.
\newblock \showarticletitle{Multi-objective optimization of wind farm
  layouts--Complexity, constraint handling and scalability}.
\newblock \bibinfo{journal}{\emph{Renewable and Sustainable Energy Reviews}}
  \bibinfo{volume}{65} (\bibinfo{year}{2016}), \bibinfo{pages}{587--609}.
\newblock


\bibitem[\protect\citeauthoryear{Rokach}{Rokach}{2010}]%
        {rokach2010ensemble}
\bibfield{author}{\bibinfo{person}{Lior Rokach}.}
  \bibinfo{year}{2010}\natexlab{}.
\newblock \showarticletitle{Ensemble-based classifiers}.
\newblock \bibinfo{journal}{\emph{Artificial Intelligence Review}}
  \bibinfo{volume}{33}, \bibinfo{number}{1} (\bibinfo{year}{2010}),
  \bibinfo{pages}{1--39}.
\newblock


\bibitem[\protect\citeauthoryear{Russakovsky, Deng, Su, Krause, Satheesh, Ma,
  Huang, Karpathy, Khosla, Bernstein, et~al\mbox{.}}{Russakovsky
  et~al\mbox{.}}{2015}]%
        {russakovsky2015imagenet}
\bibfield{author}{\bibinfo{person}{Olga Russakovsky}, \bibinfo{person}{Jia
  Deng}, \bibinfo{person}{Hao Su}, \bibinfo{person}{Jonathan Krause},
  \bibinfo{person}{Sanjeev Satheesh}, \bibinfo{person}{Sean Ma},
  \bibinfo{person}{Zhiheng Huang}, \bibinfo{person}{Andrej Karpathy},
  \bibinfo{person}{Aditya Khosla}, \bibinfo{person}{Michael Bernstein},
  {et~al\mbox{.}}} \bibinfo{year}{2015}\natexlab{}.
\newblock \showarticletitle{ImageNet large scale visual recognition challenge}.
\newblock \bibinfo{journal}{\emph{International Journal of Computer Vision}}
  \bibinfo{volume}{115}, \bibinfo{number}{3} (\bibinfo{year}{2015}),
  \bibinfo{pages}{211--252}.
\newblock


\bibitem[\protect\citeauthoryear{Sandler, Howard, Zhu, Zhmoginov, and
  Chen}{Sandler et~al\mbox{.}}{2018}]%
        {sandler2018mobilenetv2}
\bibfield{author}{\bibinfo{person}{Mark Sandler}, \bibinfo{person}{Andrew
  Howard}, \bibinfo{person}{Menglong Zhu}, \bibinfo{person}{Andrey Zhmoginov},
  {and} \bibinfo{person}{Liang-Chieh Chen}.} \bibinfo{year}{2018}\natexlab{}.
\newblock \showarticletitle{MobileNetv2: Inverted residuals and linear
  bottlenecks}. In \bibinfo{booktitle}{\emph{Proceedings of the IEEE Conference
  on Computer Vision and Pattern Recognition}}. \bibinfo{pages}{4510--4520}.
\newblock


\bibitem[\protect\citeauthoryear{Sharma, Deora, and Vishvakarma}{Sharma
  et~al\mbox{.}}{2020}]%
        {sharma2020alphanet}
\bibfield{author}{\bibinfo{person}{Rishab Sharma}, \bibinfo{person}{Rahul
  Deora}, {and} \bibinfo{person}{Anirudha Vishvakarma}.}
  \bibinfo{year}{2020}\natexlab{}.
\newblock \showarticletitle{AlphaNet: An Attention Guided Deep Network for
  Automatic Image Matting}. In \bibinfo{booktitle}{\emph{2020 International
  Conference on Omni-layer Intelligent Systems}}. IEEE, \bibinfo{pages}{1--8}.
\newblock


\bibitem[\protect\citeauthoryear{Shu, Chen, Dai, and Low}{Shu
  et~al\mbox{.}}{2021}]%
        {shu2021going}
\bibfield{author}{\bibinfo{person}{Yao Shu}, \bibinfo{person}{Yizhou Chen},
  \bibinfo{person}{Zhongxiang Dai}, {and} \bibinfo{person}{Bryan Kian~Hsiang
  Low}.} \bibinfo{year}{2021}\natexlab{}.
\newblock \showarticletitle{Going Beyond Neural Architecture Search with
  Sampling-based Neural Ensemble Search}.
\newblock \bibinfo{journal}{\emph{arXiv preprint arXiv:2109.02533}}
  (\bibinfo{year}{2021}).
\newblock


\bibitem[\protect\citeauthoryear{Singh and Deb}{Singh and Deb}{2006}]%
        {singh2006comparison}
\bibfield{author}{\bibinfo{person}{Gulshan Singh} {and}
  \bibinfo{person}{Kalyanmoy Deb}.} \bibinfo{year}{2006}\natexlab{}.
\newblock \showarticletitle{Comparison of multi-modal optimization algorithms
  based on evolutionary algorithms}. In \bibinfo{booktitle}{\emph{Proceedings
  of the 8th Annual Conference on Genetic and Evolutionary Computation}}.
  \bibinfo{pages}{1305--1312}.
\newblock


\bibitem[\protect\citeauthoryear{Stamoulis, Ding, Wang, Lymberopoulos,
  Priyantha, Liu, and Marculescu}{Stamoulis et~al\mbox{.}}{2019}]%
        {stamoulis2019single}
\bibfield{author}{\bibinfo{person}{Dimitrios Stamoulis},
  \bibinfo{person}{Ruizhou Ding}, \bibinfo{person}{Di Wang},
  \bibinfo{person}{Dimitrios Lymberopoulos}, \bibinfo{person}{Bodhi Priyantha},
  \bibinfo{person}{Jie Liu}, {and} \bibinfo{person}{Diana Marculescu}.}
  \bibinfo{year}{2019}\natexlab{}.
\newblock \showarticletitle{Single-path NAS: Designing hardware-efficient
  convnets in less than 4 hours}.
\newblock \bibinfo{journal}{\emph{arXiv preprint arXiv:1904.02877}}
  (\bibinfo{year}{2019}).
\newblock


\bibitem[\protect\citeauthoryear{Streeter}{Streeter}{2018}]%
        {streeter2018approximation}
\bibfield{author}{\bibinfo{person}{Matthew Streeter}.}
  \bibinfo{year}{2018}\natexlab{}.
\newblock \showarticletitle{Approximation algorithms for cascading prediction
  models}. In \bibinfo{booktitle}{\emph{International Conference on Machine
  Learning}}. PMLR, \bibinfo{pages}{4752--4760}.
\newblock


\bibitem[\protect\citeauthoryear{Tan and Le}{Tan and Le}{2019}]%
        {tan2019efficientnet}
\bibfield{author}{\bibinfo{person}{Mingxing Tan} {and} \bibinfo{person}{Quoc
  Le}.} \bibinfo{year}{2019}\natexlab{}.
\newblock \showarticletitle{{E}fficient{N}et: Rethinking Model Scaling for
  Convolutional Neural Networks}. In \bibinfo{booktitle}{\emph{Proceedings of
  the 36th International Conference on Machine Learning}}
  \emph{(\bibinfo{series}{Proceedings of Machine Learning Research},
  Vol.~\bibinfo{volume}{97})}, \bibfield{editor}{\bibinfo{person}{Kamalika
  Chaudhuri} {and} \bibinfo{person}{Ruslan Salakhutdinov}} (Eds.).
  \bibinfo{publisher}{PMLR}, \bibinfo{pages}{6105--6114}.
\newblock
\urldef\tempurl%
\url{https://proceedings.mlr.press/v97/tan19a.html}
\showURL{%
\tempurl}


\bibitem[\protect\citeauthoryear{Thierens and Bosman}{Thierens and
  Bosman}{2011}]%
        {thierens2011optimal}
\bibfield{author}{\bibinfo{person}{Dirk Thierens} {and} \bibinfo{person}{Peter
  A~N Bosman}.} \bibinfo{year}{2011}\natexlab{}.
\newblock \showarticletitle{Optimal mixing evolutionary algorithms}. In
  \bibinfo{booktitle}{\emph{Proceedings of the 13th Annual Conference on
  Genetic and Evolutionary Computation}}. \bibinfo{pages}{617--624}.
\newblock


\bibitem[\protect\citeauthoryear{Van~Veldhuizen and Lamont}{Van~Veldhuizen and
  Lamont}{1998}]%
        {van1998multiobjective}
\bibfield{author}{\bibinfo{person}{David~A Van~Veldhuizen} {and}
  \bibinfo{person}{Gary~B Lamont}.} \bibinfo{year}{1998}\natexlab{}.
\newblock \bibinfo{booktitle}{\emph{Multiobjective evolutionary algorithm
  research: A history and analysis}}.
\newblock \bibinfo{type}{{T}echnical {R}eport}.
  \bibinfo{institution}{Citeseer}.
\newblock


\bibitem[\protect\citeauthoryear{Viola and Jones}{Viola and Jones}{2001}]%
        {viola2001rapid}
\bibfield{author}{\bibinfo{person}{Paul Viola} {and} \bibinfo{person}{Michael
  Jones}.} \bibinfo{year}{2001}\natexlab{}.
\newblock \showarticletitle{Rapid object detection using a boosted cascade of
  simple features}. In \bibinfo{booktitle}{\emph{Proceedings of the 2001 IEEE
  Computer Society Conference on Computer Vision and Pattern Recognition}},
  Vol.~\bibinfo{volume}{1}. IEEE, \bibinfo{pages}{I--I}.
\newblock


\bibitem[\protect\citeauthoryear{Wang, Li, Gong, and Chandra}{Wang
  et~al\mbox{.}}{2021}]%
        {wang2021attentivenas}
\bibfield{author}{\bibinfo{person}{Dilin Wang}, \bibinfo{person}{Meng Li},
  \bibinfo{person}{Chengyue Gong}, {and} \bibinfo{person}{Vikas Chandra}.}
  \bibinfo{year}{2021}\natexlab{}.
\newblock \showarticletitle{AttentiveNAS: Improving neural architecture search
  via attentive sampling}. In \bibinfo{booktitle}{\emph{Proceedings of the
  IEEE/CVF Conference on Computer Vision and Pattern Recognition}}.
  \bibinfo{pages}{6418--6427}.
\newblock


\bibitem[\protect\citeauthoryear{Wang, Kondratyuk, Kitani, Movshovitz-Attias,
  and Eban}{Wang et~al\mbox{.}}{2020}]%
        {wang2020multiple}
\bibfield{author}{\bibinfo{person}{Xiaofang Wang}, \bibinfo{person}{Dan
  Kondratyuk}, \bibinfo{person}{Kris~M Kitani}, \bibinfo{person}{Yair
  Movshovitz-Attias}, {and} \bibinfo{person}{Elad Eban}.}
  \bibinfo{year}{2020}\natexlab{}.
\newblock \showarticletitle{Multiple networks are more efficient than one: Fast
  and accurate models via ensembles and cascades}.
\newblock \bibinfo{journal}{\emph{arXiv preprint arXiv:2012.01988}}
  (\bibinfo{year}{2020}).
\newblock


\bibitem[\protect\citeauthoryear{White, Zela, Ru, Liu, and Hutter}{White
  et~al\mbox{.}}{2021}]%
        {white2021powerful}
\bibfield{author}{\bibinfo{person}{Colin White}, \bibinfo{person}{Arber Zela},
  \bibinfo{person}{Binxin Ru}, \bibinfo{person}{Yang Liu}, {and}
  \bibinfo{person}{Frank Hutter}.} \bibinfo{year}{2021}\natexlab{}.
\newblock \showarticletitle{How Powerful are Performance Predictors in Neural
  Architecture Search?}. In \bibinfo{booktitle}{\emph{Advances in Neural
  Information Processing Systems}}, Vol.~\bibinfo{volume}{34}.
  \bibinfo{pages}{28454--28469}.
\newblock


\bibitem[\protect\citeauthoryear{Wichard, Merkwirth, and Ogorzalek}{Wichard
  et~al\mbox{.}}{2003}]%
        {wichard2003building}
\bibfield{author}{\bibinfo{person}{J{\"o}rg Wichard},
  \bibinfo{person}{Christian Merkwirth}, {and} \bibinfo{person}{Maciej
  Ogorzalek}.} \bibinfo{year}{2003}\natexlab{}.
\newblock \showarticletitle{Building ensembles with heterogeneous models}.
\newblock  (\bibinfo{year}{2003}).
\newblock


\bibitem[\protect\citeauthoryear{Wightman}{Wightman}{2019}]%
        {rw2019timm}
\bibfield{author}{\bibinfo{person}{Ross Wightman}.}
  \bibinfo{year}{2019}\natexlab{}.
\newblock \bibinfo{title}{PyTorch Image Models}.
\newblock
  \bibinfo{howpublished}{\url{https://github.com/rwightman/pytorch-image-models}}.
\newblock
\urldef\tempurl%
\url{https://doi.org/10.5281/zenodo.4414861}
\showDOI{\tempurl}


\bibitem[\protect\citeauthoryear{Wilcoxon}{Wilcoxon}{1992}]%
        {wilcoxon1992individual}
\bibfield{author}{\bibinfo{person}{Frank Wilcoxon}.}
  \bibinfo{year}{1992}\natexlab{}.
\newblock \showarticletitle{Individual comparisons by ranking methods}.
\newblock In \bibinfo{booktitle}{\emph{Breakthroughs in statistics}}.
  \bibinfo{publisher}{Springer}, \bibinfo{pages}{196--202}.
\newblock


\bibitem[\protect\citeauthoryear{Xu, Kusner, Weinberger, Chen, and Chapelle}{Xu
  et~al\mbox{.}}{2014}]%
        {xu2014classifier}
\bibfield{author}{\bibinfo{person}{Zhixiang Xu}, \bibinfo{person}{Matt~J
  Kusner}, \bibinfo{person}{Kilian~Q Weinberger}, \bibinfo{person}{Minmin
  Chen}, {and} \bibinfo{person}{Olivier Chapelle}.}
  \bibinfo{year}{2014}\natexlab{}.
\newblock \showarticletitle{Classifier cascades and trees for minimizing
  feature evaluation cost}.
\newblock \bibinfo{journal}{\emph{The Journal of Machine Learning Research}}
  \bibinfo{volume}{15}, \bibinfo{number}{1} (\bibinfo{year}{2014}),
  \bibinfo{pages}{2113--2144}.
\newblock


\bibitem[\protect\citeauthoryear{Yu, Jin, Liu, Bender, Kindermans, Tan, Huang,
  Song, Pang, and Le}{Yu et~al\mbox{.}}{2020}]%
        {yu2020bignas}
\bibfield{author}{\bibinfo{person}{Jiahui Yu}, \bibinfo{person}{Pengchong Jin},
  \bibinfo{person}{Hanxiao Liu}, \bibinfo{person}{Gabriel Bender},
  \bibinfo{person}{Pieter-Jan Kindermans}, \bibinfo{person}{Mingxing Tan},
  \bibinfo{person}{Thomas Huang}, \bibinfo{person}{Xiaodan Song},
  \bibinfo{person}{Ruoming Pang}, {and} \bibinfo{person}{Quoc Le}.}
  \bibinfo{year}{2020}\natexlab{}.
\newblock \showarticletitle{BigNAS: Scaling up neural architecture search with
  big single-stage models}. In \bibinfo{booktitle}{\emph{European Conference on
  Computer Vision}}. Springer, \bibinfo{pages}{702--717}.
\newblock


\bibitem[\protect\citeauthoryear{Yun, Han, Oh, Chun, Choe, and Yoo}{Yun
  et~al\mbox{.}}{2019}]%
        {yun2019cutmix}
\bibfield{author}{\bibinfo{person}{Sangdoo Yun}, \bibinfo{person}{Dongyoon
  Han}, \bibinfo{person}{Seong~Joon Oh}, \bibinfo{person}{Sanghyuk Chun},
  \bibinfo{person}{Junsuk Choe}, {and} \bibinfo{person}{Youngjoon Yoo}.}
  \bibinfo{year}{2019}\natexlab{}.
\newblock \showarticletitle{CutMix: Regularization strategy to train strong
  classifiers with localizable features}. In
  \bibinfo{booktitle}{\emph{Proceedings of the IEEE/CVF International
  Conference on Computer Vision}}. \bibinfo{pages}{6023--6032}.
\newblock


\bibitem[\protect\citeauthoryear{Zaidi, Zela, Elsken, Holmes, Hutter, and
  Teh}{Zaidi et~al\mbox{.}}{2021}]%
        {zaidi2020neural}
\bibfield{author}{\bibinfo{person}{Sheheryar Zaidi}, \bibinfo{person}{Arber
  Zela}, \bibinfo{person}{Thomas Elsken}, \bibinfo{person}{Christopher~C.
  Holmes}, \bibinfo{person}{Frank Hutter}, {and} \bibinfo{person}{Yee~Whye
  Teh}.} \bibinfo{year}{2021}\natexlab{}.
\newblock \showarticletitle{Neural Ensemble Search for Uncertainty Estimation
  and Dataset Shift}. In \bibinfo{booktitle}{\emph{Thirty-Fifth Conference on
  Neural Information Processing Systems}}.
\newblock


\bibitem[\protect\citeauthoryear{Zitzler, Thiele, Laumanns, Fonseca, and
  Da~Fonseca}{Zitzler et~al\mbox{.}}{2003}]%
        {zitzler2003performance}
\bibfield{author}{\bibinfo{person}{Eckart Zitzler}, \bibinfo{person}{Lothar
  Thiele}, \bibinfo{person}{Marco Laumanns}, \bibinfo{person}{Carlos~M
  Fonseca}, {and} \bibinfo{person}{Viviane~Grunert Da~Fonseca}.}
  \bibinfo{year}{2003}\natexlab{}.
\newblock \showarticletitle{Performance assessment of multiobjective
  optimizers: An analysis and review}.
\newblock \bibinfo{journal}{\emph{IEEE Transactions on Evolutionary
  Computation}} \bibinfo{volume}{7}, \bibinfo{number}{2}
  (\bibinfo{year}{2003}), \bibinfo{pages}{117--132}.
\newblock


\bibitem[\protect\citeauthoryear{Zoph and Le}{Zoph and Le}{2016}]%
        {zoph2016neural}
\bibfield{author}{\bibinfo{person}{Barret Zoph} {and} \bibinfo{person}{Quoc~V
  Le}.} \bibinfo{year}{2016}\natexlab{}.
\newblock \showarticletitle{Neural architecture search with reinforcement
  learning}.
\newblock \bibinfo{journal}{\emph{arXiv preprint arXiv:1611.01578}}
  (\bibinfo{year}{2016}).
\newblock


\end{thebibliography}

\newpage
\clearpage
\appendix
\section*{APPENDIX}
\section{Hyperparameters} \label{appendix:hyperparameters}

In this section, hyperparameters for our experiments are listed.

Reproducing NAT: initial learning rate 0.01, optimizer Stochastic Gradient Descent with Nesterov momentum 0.9~\cite{nesterov1983method}, training for 30 meta-iteration of 5 epochs each, cosine learning rate schedule~\cite{loshchilov2016sgdr}, train batch size 96, validation batch size 150, CutOut~\cite{devries2017improved} with size 56, RandAugment~\cite{cubuk2020randaugment} with the policy "rand-m9-mstd0.5", 10,000 surrogate evaluations for each meta-iteration, the surrogate is an ensemble of 500 RBFs~\cite{broomhead1988multivariable}, archive size 300, sampling a network to train by constructing a distribution of possible architecture choices from archive and sampling each choice from this distribution.

Better NAT results (only differences are listed): cosine cyclic learning rate schedule with period of 5 epochs, CutOut is replaced with CutMix~\cite{yun2019cutmix}, Stochastic Weight Averaging~\cite{izmailov2018averaging} of models from the last 20 meta-iterations (i.e. epochs 150, 145,.. 55), sampling a network to train by directly sampling an architecture from the archive.

ENCAS-joint: 5 epochs per supernetwork per iteration, the rest follows "Better NAT results".

Techniques suggested to us privately by the NAT authors to improve NAT performance are: usage of CutOut, RandAugment, and Riesz s-energy~\cite{ref_dirs_energy} for reference point generation instead of the Das-Dennis method~\cite{das1998normal}.

ENCAS was run for 600,000 evaluations. For experiments with \texttt{timm} an Nvidia A100 card was used for its superior VRAM size.

% \section{Robustness to hyperparameters}

% Increasing the maximum cascade size does not influence the results much, ENCAS is still able to find a good trade-off front.

\section{Evolutionary Neural Ensemble Search} \label{appendix:ensembles}

In this section we briefly discuss a restriction of ENCAS to ensemble search, an algorithm we name Evolutionary Neural ENsemble Search (ENENS). Whereas ENCAS searches for a sequence of models and confidence thresholds, ENENS searches only for the models, as confidence thresholds are not needed in an ensemble. The overall procedure is a simplified version of the one in ENCAS (Section~\ref{methods:ens}): a solution representation consists of model indices (thresholds are no longer a part of the representation); a solution is evaluated by averaging the predictions of all the models in it; the FLOPs count is the sum of FLOPs counts of all the models in the ensemble. The rest of the algorithm is unchanged.

\subsection{Comparison with existing Neural Ensemble Search algorithms}

The existing algorithms for ensemble architecture search do not use pretrained supernetworks. Being able to use pretrained supernetworks is an important advantage in deep learning, an advantage that is realized by ENENS, which can be seen to exhibit superior performance in terms of top-1 accuracy, as reported in Table~\ref{tab:ensembles}. 

\begin{table}[h]
  \caption{Top-1 accuracy of different algorithms for search of network architectures for an ensemble. For our experiment, mean and standard deviation across 10 seeds are reported.}
  \label{tab:ensembles}
  \begin{tabular}{cccc}
    \toprule
    Method&CIFAR-10 & CIFAR-100 & ImageNet\\
    \midrule
    NES~\cite{zaidi2020neural} &  92.4 & 76.5 & --- \\
    MH-NES~\cite{narayanan2021multi} & --- & 80.35 & --- \\
    NESS~\cite{shu2021going} & 97.64 & 85.45 & 77.7 \\
    NEAS~\cite{chen2021one} & --- & --- & 80.0 \\
    \midrule
    \makecell{ENENS} & 98.62$_{\pm0.06}$ & 88.99$_{\pm0.09}$ & 82.80$_{\pm0.07}$ \\
  \bottomrule
\end{tabular}
\end{table}

\subsection{Cascade vs ensemble} 

\begin{figure}[h]
  \centering
  \begin{subfigure}[t]{0.32\linewidth}
        \centering
        \includegraphics[width=\textwidth]{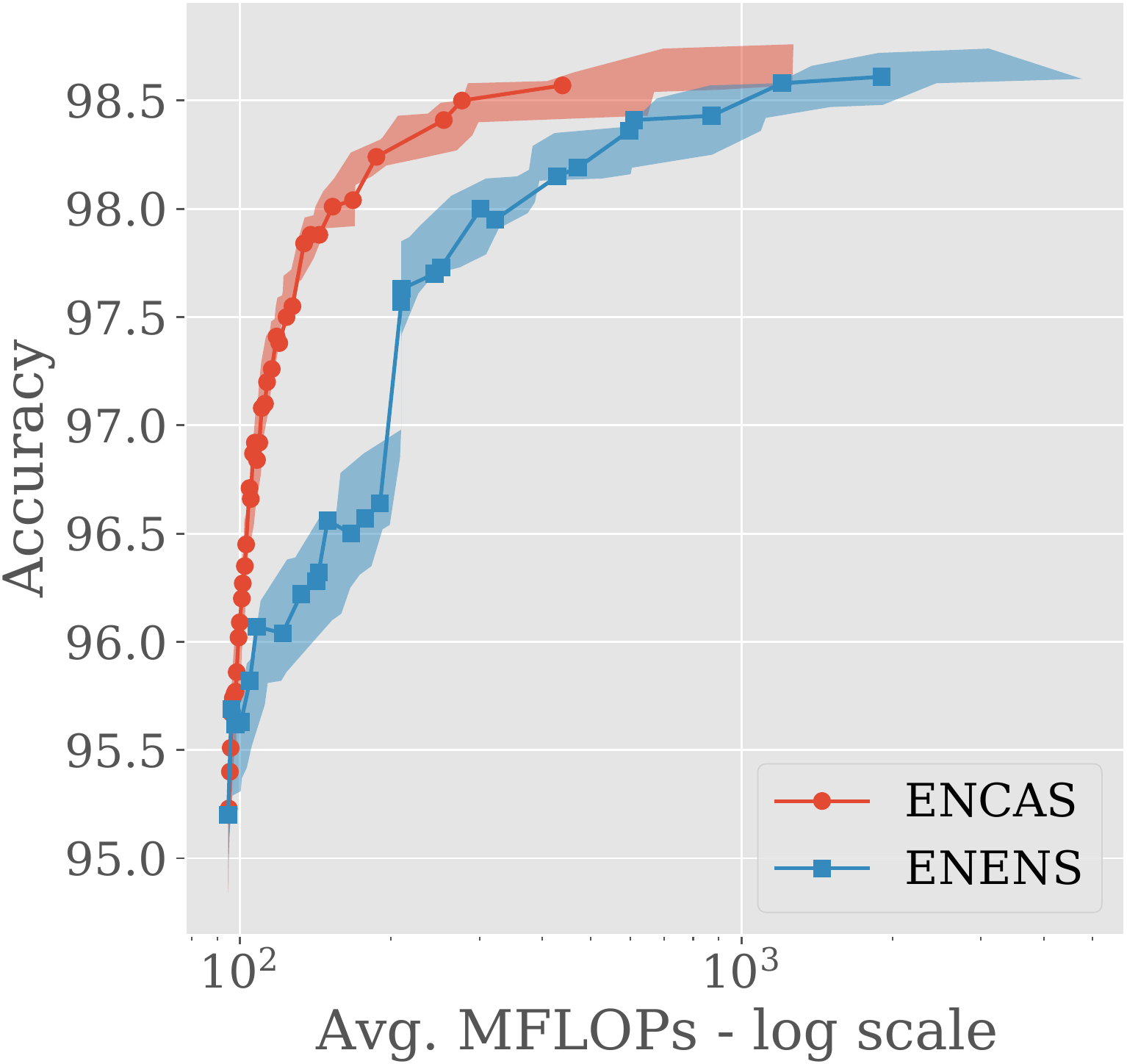}
        \caption{CIFAR-10}
    \end{subfigure}%
    ~ 
    \begin{subfigure}[t]{0.32\linewidth}
        \centering
        \includegraphics[width=\linewidth]{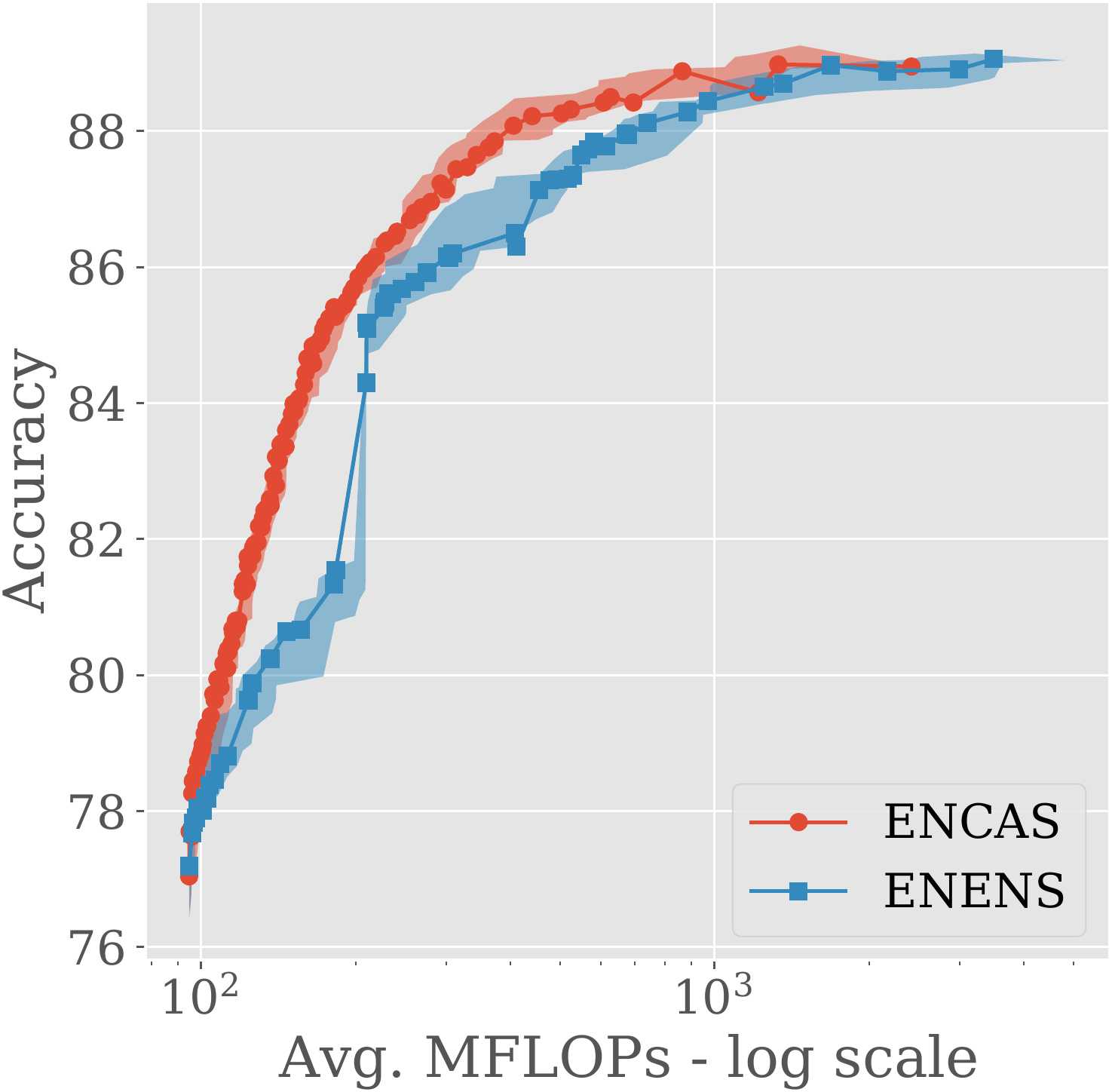}
        \caption{CIFAR-100}
    \end{subfigure}
    ~ 
    \begin{subfigure}[t]{0.32\linewidth}
        \centering
        \includegraphics[width=\linewidth]{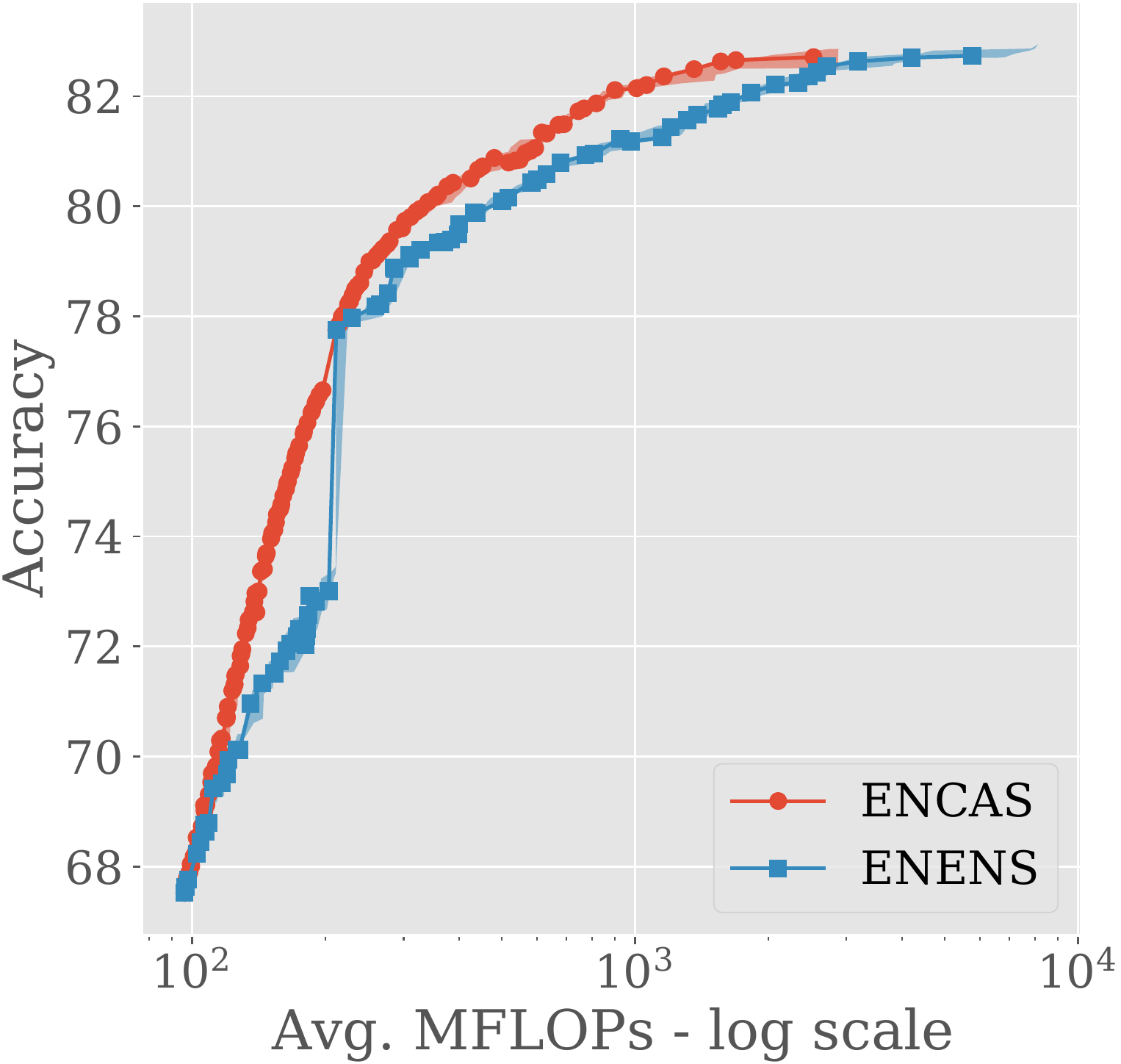}
        \caption{ImageNet}
    \end{subfigure}
  \caption{The trade-off front of cascades is better than the trade-off front of ensembles on the test set.}
  \Description{}
  \label{fig:abl:cascadevsensemble}
\end{figure}

The comparison of ENCAS and ENENS in Figure~\ref{fig:abl:cascadevsensemble} shows that the front of cascades dominates for most FLOPs values. The largest ensembles perform very similarly, or, in case of ImageNet, perhaps perform slightly better, even though technically they could be found by ENCAS as special cases of cascades. Best we can tell, this does not happen because in the largest-FLOPs regime ENCAS finds smaller cascades and ensembles (than ENENS) that perform better on the validation set, but lose a bit of performance on the test set. The larger ensembles found by ENENS do not face this issue. The small increase in accuracy comes at the cost of thousands more MFLOPs, however, which means that ensembles should be considered only when efficiency is of no concern.

% The comparison of ENCAS and ENCAS-ensemble in Figure~\ref{fig:abl:cascadevsensemble} shows that the front of cascades dominates for most FLOPS values up to a point: the largest ensembles seem to be better, although technically they could be found by ENCAS as special cases of cascades. As best we can tell, this does not happen because in the largest-FLOPS regime ENCAS finds smaller cascades and ensembles (than ENCAS-ensemble) that perform better on the validation set, but lose a bit of performance on the test set. The larger ensembles found by ENCAS-ensemble do not face this issue, thus in terms of maximum accuracy ENCAS-ensemble outperforms ENCAS. We would like to note that the difference is minimal (not statistically significant (?)) and unlikely to be relevant in practice, as trading off hundreds of FLOPS for a potential increase of 0.02 percentage points of accuracy is usually not desired.

% \vspace{-5pt}
\section{Comparison to random search} \label{abl:random}

\begin{figure}
  \centering
  \begin{subfigure}[t]{0.32\linewidth}
        \centering
        \includegraphics[width=\textwidth]{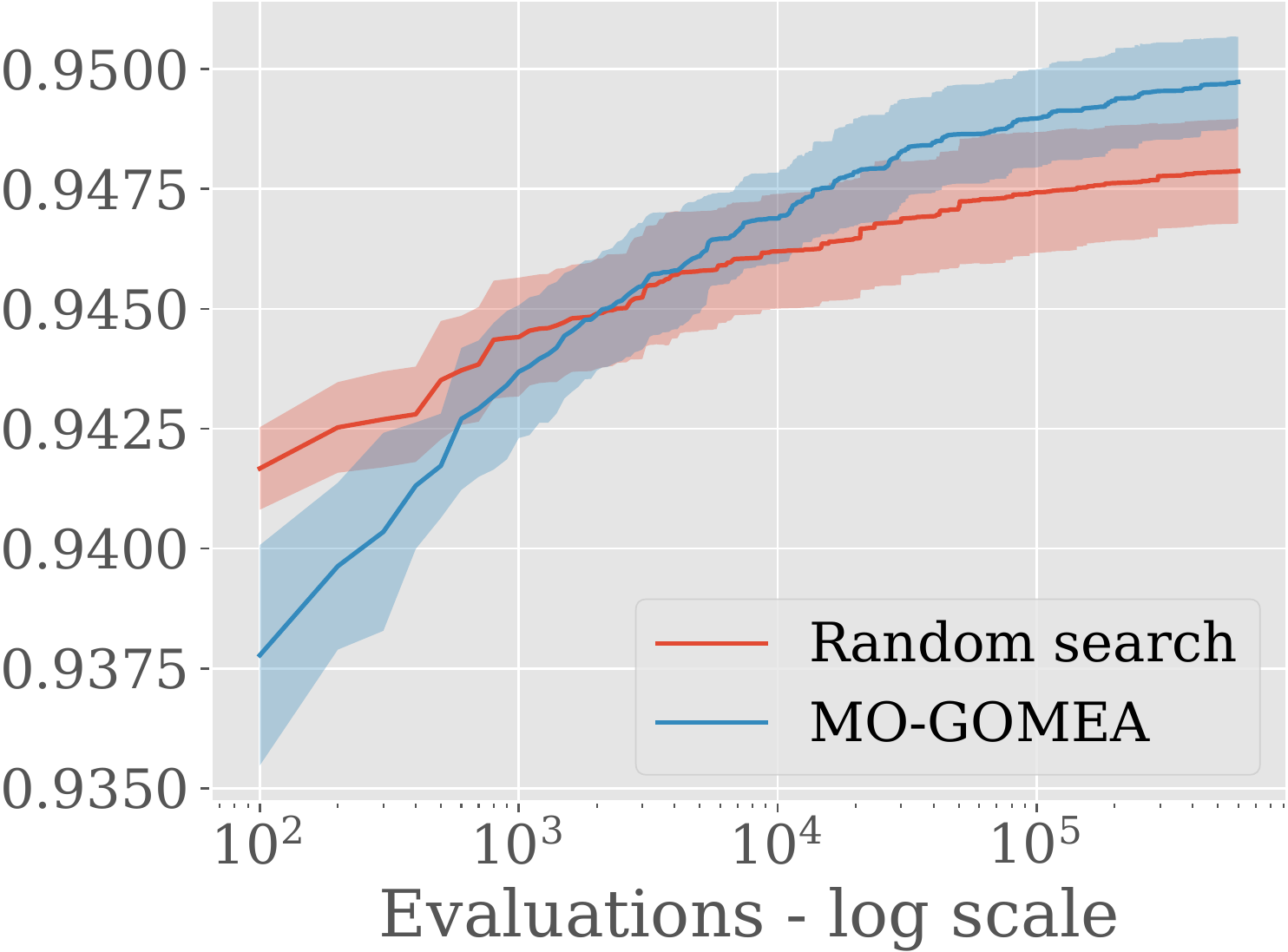}
        \caption{CIFAR-10}
    \end{subfigure}%
    ~ 
    \begin{subfigure}[t]{0.32\linewidth}
        \centering
        \includegraphics[width=\linewidth]{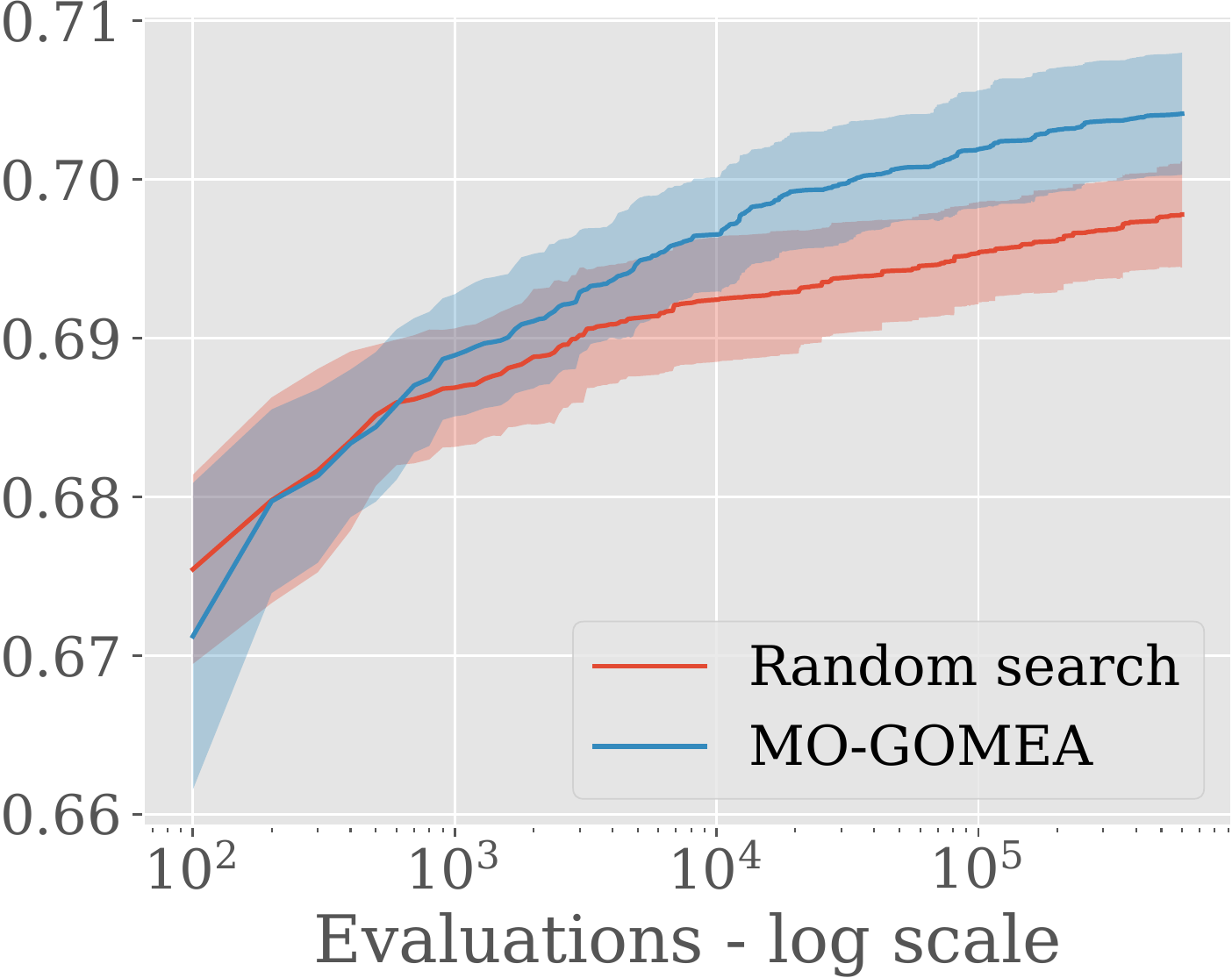}
        \caption{CIFAR-100}
    \end{subfigure}
    ~ 
    \begin{subfigure}[t]{0.32\linewidth}
        \centering
        \includegraphics[width=\linewidth]{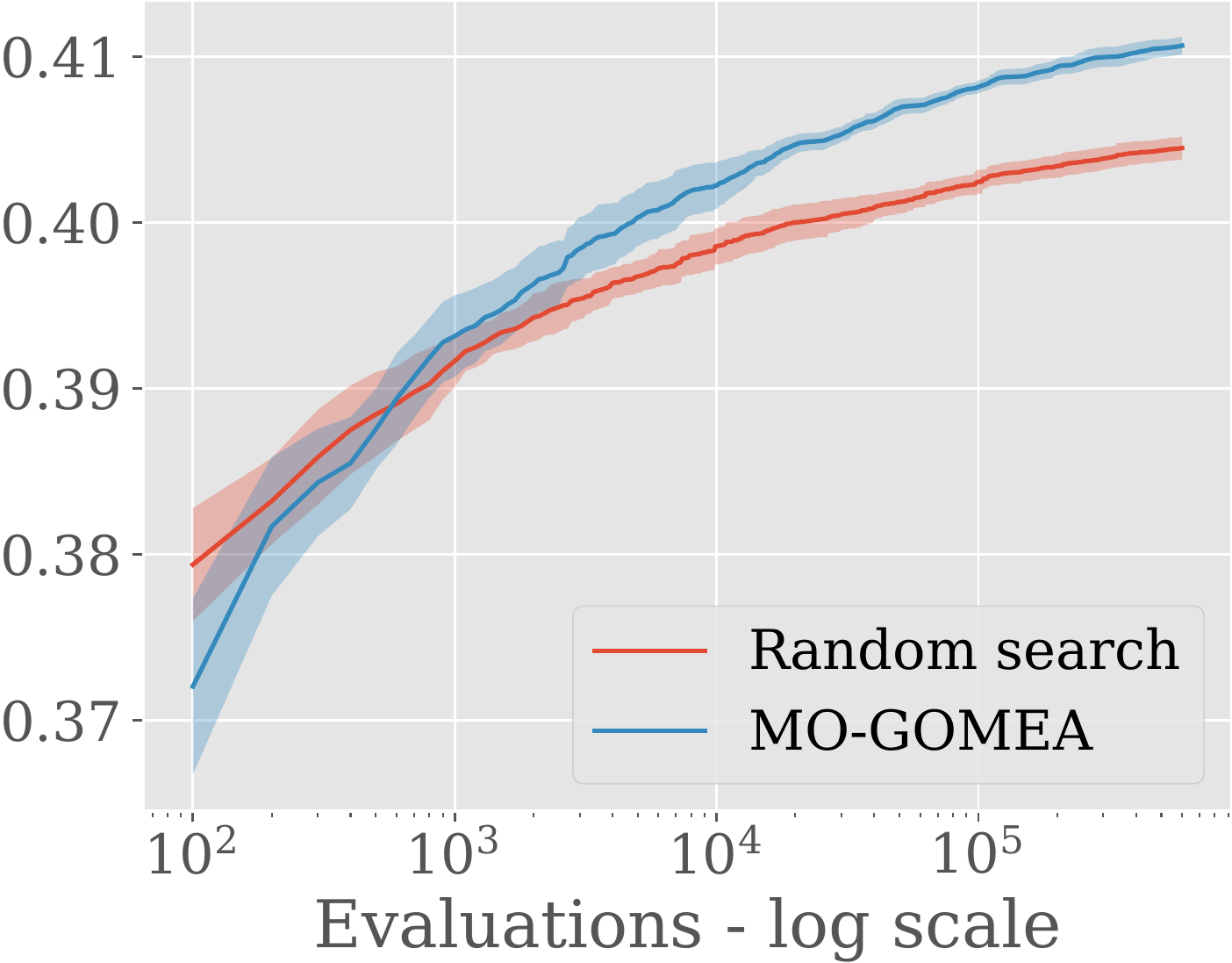}
        \caption{ImageNet}
    \end{subfigure}
  \caption{MO-GOMEA outperforms random search in terms of hypervolume of the validation trade-off fronts.}
  \Description{}
  \label{fig:abl:hypervolume}
\end{figure}

Random search is a strong baseline for NAS~\cite{li2020random}. Consequently, we replace MO-GOMEA in ENCAS with random search to investigate whether a simpler search algorithm will suffice. In Fig.~\ref{fig:abl:hypervolume} we can see that while random search achieves better results in the first several hundred evaluations of search, it is overtaken by MO-GOMEA after that, demonstrating the benefit of the more sophisticated search approach in terms of the optimization problem it is tasked to solve.

\begin{table}[h]
  \caption{Comparison of using MO-GOMEA and random search in ENCAS (test set performance, mean and standard deviation are computed across 10 seeds)}
  \label{tab:vs_random}
  \begin{tabular}{cccc}
    \toprule
    Method&\makecell{Hypervolume} & \makecell{Max accuracy} & \makecell{Max MFLOPs}\\
    \midrule
    \multicolumn{4}{c}{CIFAR-10} \\
    Random & 0.941$_{\pm0.002}$ & 98.61$_{\pm0.08}$ & 814$_{\pm307}$ \\
    MO-GOMEA  & 0.941$_{\pm0.002}$ & 98.60$_{\pm0.09}$ & 749$_{\pm298}$ \\
    \multicolumn{4}{c}{CIFAR-100} \\
    Random  & 0.698$_{\pm0.003}$ & 88.91$_{\pm0.15}$ & 1298$_{\pm389}$ \\
    MO-GOMEA & 0.699$_{\pm0.003}$ & 88.96$_{\pm0.17}$ & 1401$_{\pm420}$ \\
    \multicolumn{4}{c}{ImageNet} \\
    Random  & 0.535$_{\pm0.001}$ & 82.66$_{\pm0.07}$ & 2708$_{\pm588}$ \\
    MO-GOMEA & 0.537$_{\pm0.001}$ & 82.72$_{\pm0.10}$ & 2614$_{\pm221}$ \\
  \bottomrule
\end{tabular}
\end{table}

However, Table~\ref{tab:vs_random} shows that the difference in test performance over all the datasets is small (not statistically significant). 
% This implies that despite containing hundreds of models, search space of ENCAS is quite simple, possibly because only good models from the NAT-discovered trade-off fronts are used. 
Better performance of MO-GOMEA on the validation set but not the test set means that overfitting to the validation set occurs. Combatting overfitting is an avenue for future work.

% \begin{figure}[h]
%   \centering
%   \includegraphics[width=\linewidth]{_pics/5_ablations/vs_random_fronts.png}
%   \caption{MO-GOMEA achieves very minor improvements over random search in terms of test accuracy.}
%   \Description{}
%   \label{fig:abl:fronts}
% \end{figure}

% \subsubsection{Impact of the number of evaluations} \label{abl:n_evals}

% Thanks to the precomputation of network outputs in ENCAS, its run time is low, around 1 GPU-hour for 600.000 evaluations. Can this number be decreased for even faster search, or, in contrast, increased for better results? Figure~\ref{fig:abl:hypervolume} shows diminishing performance after (?) evaluations for all datasets. \textit{DEBUG: have not run for more than 600,000 evaluations yet. But considering that the graphs are in log-scale, mo-gomea seems mostly converged already.}

\section{ENCAS models for reference} \label{appendix:named_models}

In the main text we avoid naming specific models on the trade-off front, as we believe that it is helpful to have NAS produce tens or hundreds of models that may differ only slightly but that allow users more flexibility in choosing which one to use. Nonetheless, to make referencing our models easier, here we specify FLOPs and top-1 accuracy of selected models. A model is included if its MFLOPs value is closest to a value divisible by 100~---~this way we select a representative and spread-out sample of the trade-off front while avoiding cherry-picking.  The models are taken from the median run (by hypervolume as evaluated on the test set).

\begin{table}[h]
  \caption{CIFAR-10, selected models}
  \begin{tabular}{lcc}
    \toprule
    Name& MFLOPs & Accuracy\\
    \midrule
    ENCAS@100 & 100 & 96.19 \\
    ENCAS@186 & 186 & 98.23 \\
    ENCAS@276 & 276 & 98.49 \\
    ENCAS@439 & 439 & 98.57 \\
  \bottomrule
\end{tabular}
\end{table}
% \vspace{-5pt}

\begin{table}[h]
  \caption{CIFAR-100, selected models}
  \begin{tabular}{lcc}
    \toprule
    Name& MFLOPs & Accuracy\\
    \midrule
    ENCAS@100 & 100 & 78.98 \\
    ENCAS@202 & 202 & 85.85 \\
    ENCAS@299 & 299 & 87.14 \\
    ENCAS@405 & 405 & 88.08 \\
    ENCAS@503 & 503 & 88.26 \\
    ENCAS@607 & 607 & 88.42 \\
    ENCAS@694 & 694 & 88.42 \\
    ENCAS@865 & 865 & 88.88 \\
    ENCAS@1217 & 1217 & 88.57 \\
    ENCAS@1330 & 1330 & 88.98 \\
    ENCAS@2419 & 2419 & 88.95 \\
  \bottomrule
\end{tabular}
\end{table}

% \vspace{-5pt}
\begin{table}[h]
  \caption{ImageNet, selected models}
  \begin{tabular}{lcc}
    \toprule
    Name& MFLOPs & Accuracy\\
    \midrule
    ENCAS@100 & 100 & 68.19 \\
    ENCAS@211 & 211 & 77.75 \\
    ENCAS@301 & 301 & 79.74 \\
    ENCAS@387 & 387 & 80.43 \\
    ENCAS@517 & 517 & 80.79 \\
    ENCAS@615 & 615 & 81.34 \\
    ENCAS@690 & 690 & 81.49 \\
    ENCAS@817 & 817 & 81.87 \\
    ENCAS@900 & 900 & 82.11 \\
    ENCAS@1006 & 1006 & 82.14 \\
    ENCAS@1060 & 1060 & 82.20 \\
    ENCAS@1160 & 1160 & 82.36 \\
    ENCAS@1357 & 1357 & 82.49 \\
    ENCAS@1559 & 1559 & 82.63 \\
    ENCAS@1687 & 1687 & 82.66 \\
    ENCAS@2526 & 2526 & 82.71 \\
  \bottomrule
\end{tabular}
\end{table}
% \vspace{-5pt}

\section{Statistical testing} \label{appendix:stats}

Table~\ref{tab:pvalues} lists p-values for one-sided Wilcoxon pairwise rank tests with Bonferroni correction (null hypothesis is that Algorithm 1 is worse than Algorithm 2). 

All benchmark datasets (CIFAR-10, CIFAR-100, ImageNet) are tested together to increase sample size. P-values below a significance threshold of 0.0005 are highlighted (target $p$-value=0.01, 20 tests, corrected $p$=0.0005). P-values are rounded for visualization purposes. Unless specified otherwise, 5 supernetworks are used. Experiments are separated into two groups: those in the main text and those in the appendix.

\begin{table*}[!htbp]
  \caption{P-values of conducted experiments}
  \label{tab:pvalues}
  \begin{tabular}{llccc}
    \toprule
    Algorithm 1 & Algorithm 2 & Split &Hypervolume&Max accuracy\\
    \midrule
    ENCAS (1 supernetwork) & NAT (best) & test & \cellcolor{mycolor} 5.6e-6 & 0.0008 \\
    ENCAS & NAT (best) & test & \cellcolor{mycolor} 8.7e-7 & \cellcolor{mycolor} 8.7e-7 \\
    ENCAS & ENCAS (1 supernetwork) & test & \cellcolor{mycolor} 8.7e-7 & \cellcolor{mycolor} 8.7e-7 \\
    ENCAS & GreedyCascade & test & \cellcolor{mycolor} 8.7e-7 & \cellcolor{mycolor} 8.7e-7 \\
    ENCAS-joint+ & ENCAS-joint & test & \cellcolor{mycolor} 1.9e-6 & 0.0008 \\
    ENCAS-joint+ & ENCAS & test & 0.0133 & 0.0093 \\
    ENCAS & ENCAS (5 seeds of the best supernet) & test & \cellcolor{mycolor} 9.5e-7 & 0.9385 \\
    \midrule
    ENCAS & ENENS & test & \cellcolor{mycolor} 8.7e-7 & 0.9829 \\
    ENCAS with MO-GOMEA & ENCAS with Random search & val & \cellcolor{mycolor} 8.7e-7 & \cellcolor{mycolor} 1.8e-6 \\
    ENCAS with MO-GOMEA & ENCAS with Random search & test & 0.0247 & 0.0512 \\
  \bottomrule
\end{tabular}
\end{table*}

\section{Hypervolume} \label{appendix:hypervolume}

\begin{figure}[h]
  \centering
  \includegraphics[width=0.5\linewidth]{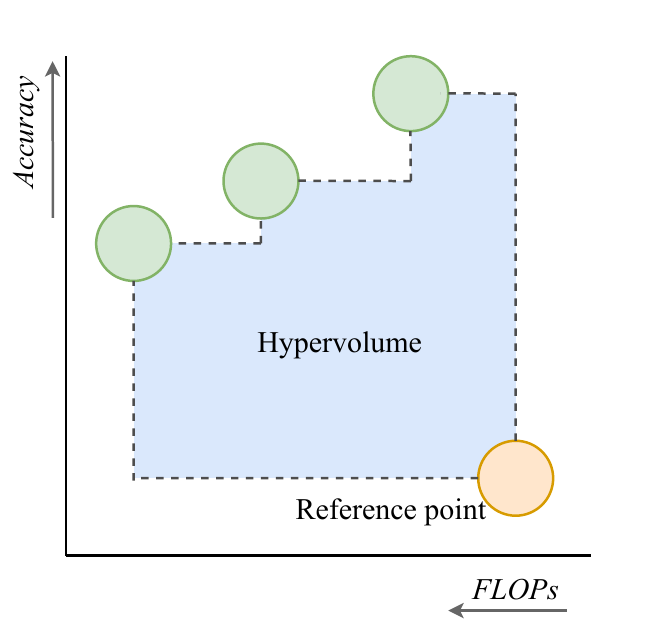}
  \caption{Hypervolume indicator for the task of maximizing accuracy and minimizing FLOPs}
  \Description{}
  \label{fig:abl:hvdef}
\end{figure}

The hypervolume indicator measures the volume encapsulated between the trade-off front and a reference point in the objective space~\cite{zitzler2003performance}. Hypervolume is a useful metric of the quality of a trade-off front.

Since hypervolume is measured relative to a reference point, its absolute value has no meaning, as changing the reference point changes it. However, if the reference point is fixed, relative differences between the hypervolumes of different algorithms can indicate which of them finds a better trade-off front. Note however that hypervolume, while useful, does not perfectly catch every aspect that is important when comparing fronts. Especially when values are close, other measures or visual front inspection are advised.

A visual representation of hypervolume can be seen in Fig.~\ref{fig:abl:hvdef}. Normalized hypervolume means hypervolume divided by its largest possible value. In our experiments the reference point is fixed at (MFLOPs=$4000$, accuracy=$60$).

\end{document}